\documentclass[10pt,twocolumn,letterpaper]{article}

\usepackage{sty_files/iccv}
\usepackage{times}
\usepackage{epsfig}
\usepackage{graphicx}
\usepackage{amsmath}
\usepackage{amssymb}
\usepackage{bm}
\usepackage{booktabs}
\usepackage{nicefrac}
\usepackage{xcolor}
\usepackage{multirow}
\usepackage{overpic}
\usepackage{transparent}
\usepackage{wrapfig}
\usepackage{setspace}
\usepackage{algorithmic}
\usepackage{nicefrac}
\usepackage{mathtools}
\usepackage{multicol}
\usepackage{pifont}  
\usepackage{colortbl}
\usepackage{url}
\usepackage{relsize}

\usepackage[font=small]{caption}
\usepackage[font=small]{subcaption}
\usepackage{algorithm}
\usepackage[group-separator={,}]{siunitx}
\usepackage{bm}
\DeclareMathOperator*{\argmax}{argmax}
\DeclareMathOperator*{\argmin}{argmin}
\definecolor{ourmethod}{gray}{0.93}
\newcommand\FBox[1]{{\setlength{\fboxsep}{-0.5pt}\setlength{\fboxrule}{0.5pt}\frame{#1}}}

\usepackage[pagebackref=true,breaklinks=true,letterpaper=true,colorlinks,bookmarks=false]{hyperref}

\iccvfinalcopy

\definecolor{grey}{rgb}{0.8, 0.8, 0.8}

\definecolor{mplblue}{HTML}{377eb8}
\definecolor{mplgreen}{HTML}{4daf4a}
\definecolor{goalred}{RGB}{181,23,0}
\definecolor{mplorange}{HTML}{ff7f00}
\definecolor{scrossgrey}{HTML}{363636}

\newcommand{\algname}[0]{PRECOG}
\newcommand{\expectation}[2]{\mathbb{E}_{#1}[#2]}

\newcommand{\latent}[0]{\mathbf{z}} 
\newcommand{\Latent}[0]{\mathbf{Z}} 
\newcommand{\State}[0]{\mathbf{S}}
\newcommand{\state}[0]{\mathbf{s}}

\newcommand{\bmu}{\boldsymbol{\mu}}
\newcommand{\bsigma}{\boldsymbol{\sigma}}
\newcommand{\bSigma}{\boldsymbol{\Sigma}}
\newcommand{\bchi}{\boldsymbol{\chi}}
\newcommand{\bGamma}{\boldsymbol{\Gamma}}
\newcommand{\netm}{\mathbf{m}}
\newcommand{\Eye}{\mathbf{I}}
\newcommand{\Zero}{\mathbf{0}}
\newcommand{\waypoint}{\mathbf{w}}
\newcommand{\data}[0]{\mathcal{D}}
\newcommand\iidsim{\stackrel{\mathclap{\mathrm{iid}}}{\sim}}
\newcommand\alltime{}

\newcommand{\app}[1]{Appendix~\ref{app:#1}}
\newcommand{\eqn}[1]{\eqref{eqn:#1}}
\newcommand{\fig}[1]{Fig.~\ref{fig:#1}}

\newcommand{\sct}[1]{Sec.~\ref{sec:#1}}
\newcommand{\tab}[1]{Tab.~\ref{tab:#1}}
\definecolor{goalyellow}{RGB}{253,238,0}

\makeatletter
  \let\ps@plain\ps@empty
\makeatother

\begin{document}

\title{{PRECOG}: PREdiction Conditioned On Goals in Visual Multi-Agent Settings}

\author{Nicholas Rhinehart$^1$\\
\and
Rowan McAllister$^2$\\
\and 
Kris Kitani$^1$\\
\and 
Sergey Levine$^2$\\
\and
{${}^{1}$Carnegie Mellon University}\\
{\tt\small\{nrhineha,kkitani\}@cs.cmu.edu}
\and
{${}^{2}$University of California, Berkeley}\\
{\tt\small\{rmcallister,svlevine\}@berkeley.edu} 
}
\pagestyle{empty}
\maketitle

\ificcvfinal\pagestyle{empty}\fi  

\begin{abstract}
\vspace{-1mm}
For autonomous vehicles (AVs) to behave appropriately on roads populated by human-driven vehicles, they must be able to reason about the uncertain intentions and decisions of other drivers from rich perceptual information. Towards these capabilities, we present a probabilistic forecasting model of future interactions between a variable number of agents. We perform both standard forecasting and the novel task of conditional forecasting, which reasons about how all agents will likely respond to the goal of a controlled agent (here, the AV). We train models on real and simulated data to forecast vehicle trajectories given past positions and LIDAR. Our evaluation shows that our model is substantially more accurate in multi-agent driving scenarios compared to existing state-of-the-art. Beyond its general ability to perform conditional forecasting queries, we show that our model's predictions of all agents improve when conditioned on knowledge of the AV's goal, further illustrating its capability to model agent interactions.
\end{abstract}

\vspace{-2mm}  
\section{Introduction}

\begin{figure}[!th]
  \centering
        \begin{overpic}[width=0.945\columnwidth,trim={1mm 0 0mm 12mm},clip]{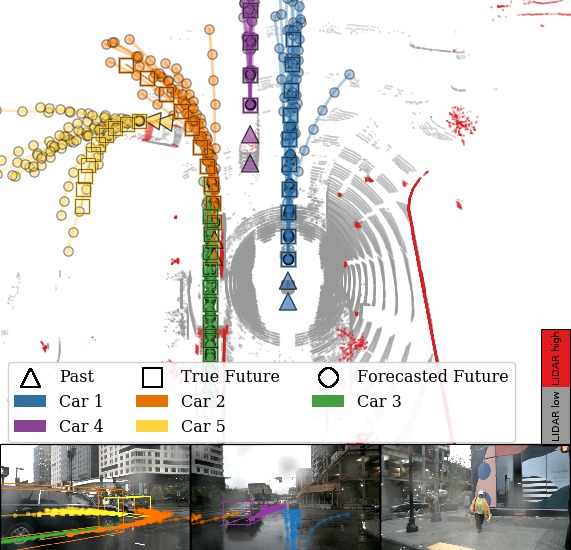}
    \put(1,15.75){{\setlength{\fboxsep}{1pt}\fontsize{8pt}{0pt}\selectfont\transparent{0.7}\colorbox{white}{\textcolor{black}{\texttransparent{1.0}{Left\vphantom{g}}}}}}
    \put(33,15.75){{\setlength{\fboxsep}{1pt}\fontsize{8pt}{0pt}\selectfont\transparent{0.7}\colorbox{white}{\textcolor{black}{\texttransparent{1.0}{Front\vphantom{g}}}}}}
    \put(67,15.75){{\setlength{\fboxsep}{1pt}\fontsize{8pt}{0pt}\selectfont\transparent{0.7}\colorbox{white}{\textcolor{black}{\texttransparent{1.0}{Right}}}}}
    \end{overpic}
    \caption{\small Forecasting on nuScenes \cite{nuscenes}. The input to our model is a high-dimensional LIDAR observation, which informs a distribution over all agents' future trajectories.} \label{fig:unconditioned_teaser}
\end{figure}

\begin{figure}[!th]
  \centering
        \begin{overpic}[width=0.71\columnwidth]{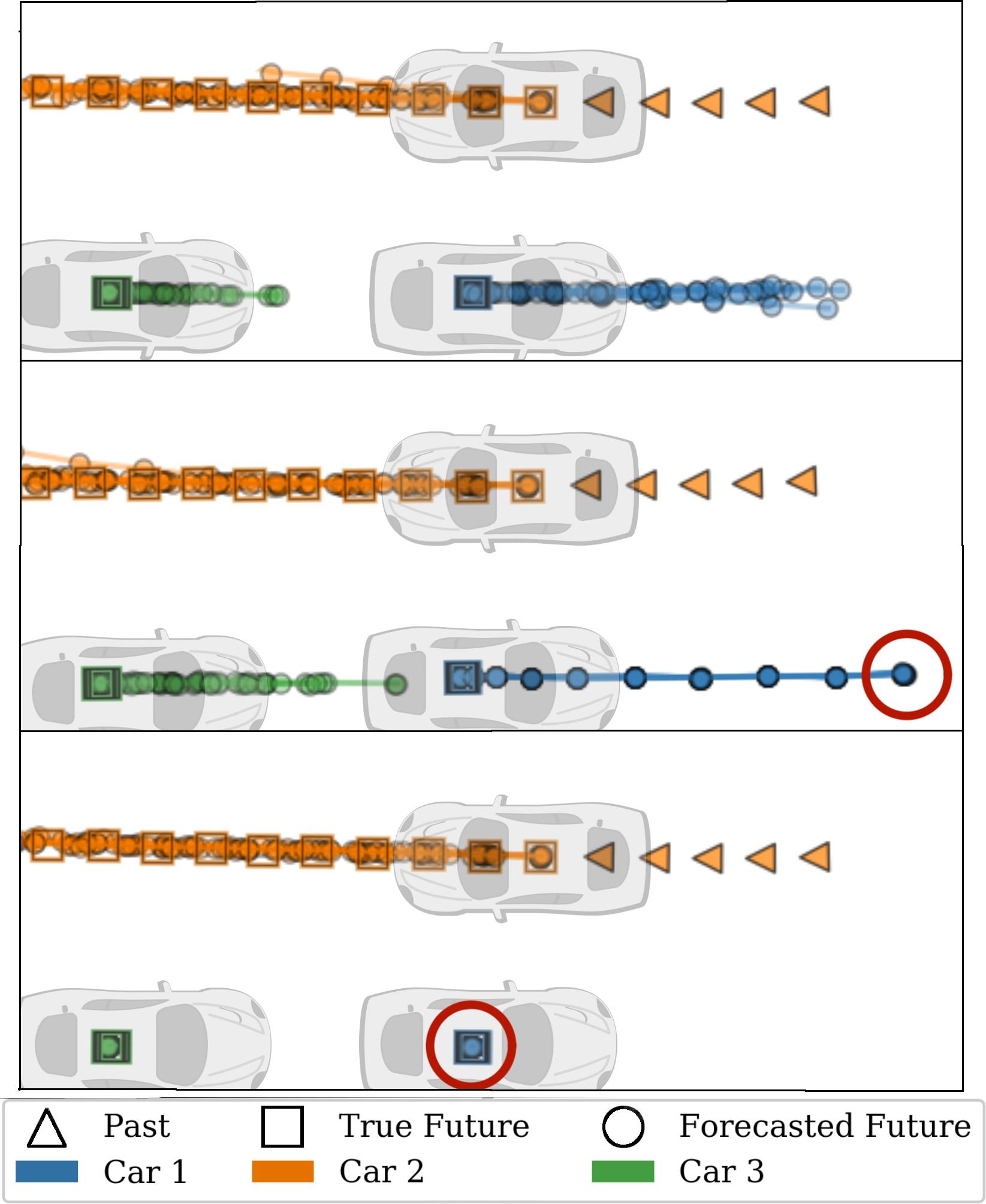} 
        \put(2,96.5){{\setlength{\fboxsep}{1pt}\fontsize{8pt}{0pt}\selectfont\transparent{0.0}\colorbox{white}{\textcolor{black}{\texttransparent{1.0}{\bf Forecasting \vphantom{g}}}}}}
        \put(2,66.5){{\setlength{\fboxsep}{1pt}\fontsize{8pt}{0pt}\selectfont\transparent{0.0}\colorbox{white}{\textcolor{black}{\texttransparent{1.0}{\bf Conditional Forecast:\, Set Car 1 Goal=Ahead\vphantom{g}}}}}}
        \put(2,36){{\setlength{\fboxsep}{1pt}\fontsize{8pt}{0pt}\selectfont\transparent{0.0}\colorbox{white}{\textcolor{black}{\texttransparent{1.0}{\bf Conditional Forecast:\, Set Car 1 Goal=Stop\vphantom{g}}}}}}
        
        \put(60,49){{\setlength{\fboxsep}{1pt}\fontsize{7pt}{0pt}\selectfont\transparent{0.0}\colorbox{white}{\textcolor{goalred}{\texttransparent{1.0}{Goal=Ahead\vphantom{g}}}}}}  
        \put(43,19){{\setlength{\fboxsep}{1pt}\fontsize{7pt}{0pt}\selectfont\transparent{0.0}\colorbox{white}{\textcolor{goalred}{\texttransparent{1.0}{Goal=Stop \vphantom{g}}}}}}
        
        \put(34,81){{\setlength{\fboxsep}{1pt}\fontsize{7pt}{0pt}\selectfont\transparent{0.0}\colorbox{white}{\textcolor{mplblue}{\texttransparent{1.0}{Car 1\vphantom{g}}}}}} 
        \put(34,49){{\setlength{\fboxsep}{1pt}\fontsize{7pt}{0pt}\selectfont\transparent{0.0}\colorbox{white}{\textcolor{mplblue}{\texttransparent{1.0}{Car 1\vphantom{g}}}}}} 
        \put(34,19){{\setlength{\fboxsep}{1pt}\fontsize{7pt}{0pt}\selectfont\transparent{0.0}\colorbox{white}{\textcolor{mplblue}{\texttransparent{1.0}{Car 1\vphantom{g}}}}}} 
        
        \put(34,97){{\setlength{\fboxsep}{1pt}\fontsize{7pt}{0pt}\selectfont\transparent{0.0}\colorbox{white}{\textcolor{mplorange}{\texttransparent{1.0}{Car 2\vphantom{g}}}}}} 
        
        \put(5,81.5){{\setlength{\fboxsep}{1pt}\fontsize{7pt}{0pt}\selectfont\transparent{0.0}\colorbox{white}{\textcolor{mplgreen}{\texttransparent{1.0}{Car 3\vphantom{g}}}}}} 
        \put(5,48.5){{\setlength{\fboxsep}{1pt}\fontsize{7pt}{0pt}\selectfont\transparent{0.0}\colorbox{white}{\textcolor{mplgreen}{\texttransparent{1.0}{Car 3\vphantom{g}}}}}} 
        \put(5,19){{\setlength{\fboxsep}{1pt}\fontsize{7pt}{0pt}\selectfont\transparent{0.0}\colorbox{white}{\textcolor{mplgreen}{\texttransparent{1.0}{Car 3\vphantom{g}}}}}} 

      \end{overpic}
  \caption{\small Conditioning the model on different Car 1 goals produces different predictions: here it forecasts Car 3 to move if Car 1 yields space, or stay stopped if Car 1 stays stopped.} \label{fig:conditional_teaser}
  \vspace{-12mm} 
\end{figure}
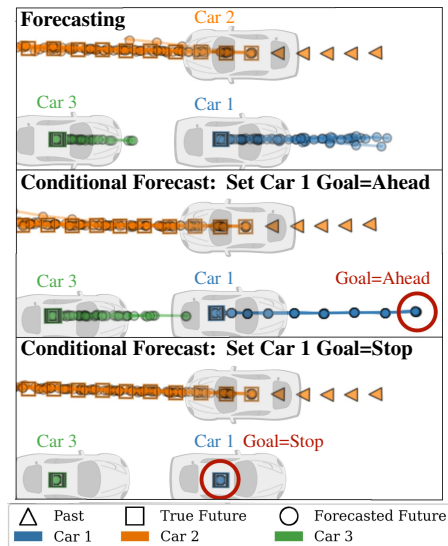

Autonomous driving requires reasoning about the future behaviors of agents in a variety of situations: at stop signs, roundabouts, crosswalks, when parking, when merging etc. In multi-agent settings, each agent's behavior affects the behavior of others.
Motivated by people's ability to reason in these settings, we present a method to forecast multi-agent interactions from perceptual data, such as images and LIDAR. Beyond forecasting the behavior of all agents, we want our model to \emph{conditionally forecast} how other agents are likely to respond to different decisions each agent could make. We want to forecast what other agents would likely do in response to a robot's intention to achieve a goal. This reasoning is essential for agents to make good decisions in multi-agent environments: 
they must reason how their future decisions could affect the multi-agent system around them. 
Examples of forecasting and conditioning forecasts on robot goals are shown in 
\fig{unconditioned_teaser} and \fig{conditional_teaser}.
Videos of the outputs of our approach are available at \url{https://sites.google.com/view/precog}.


Throughout the paper, we use \emph{goal} to mean a future states that an agent desires. \emph{Planning} means the algorithmic process of producing a sequence of 
future decisions (in our model, choices of latent values) likely to satisfy a goal. \emph{Forecasting} means the prediction of a sequence of likely future states; forecasts can either be single-agent or multi-agent. Finally, \emph{conditional forecasting} means \emph{forecasting} by conditioning on one or more agent goals. By \emph{planning} an agent's decisions to a \emph{goal} and sampling from the other agents's stochastic decisions, we perform \emph{multi-agent conditional forecasting}. Although we plan future decisions in order to perform conditional forecasting, executing these plans on the robot is outside the scope of this work.

Towards conditional forecasting, we propose a factorized flow-based generative model that forecasts the joint state of all agents. 
Our model reasons probabilistically about plausible future interactions between agents given rich observations of their environment.
It uses latent variables to capture the uncertainty in other agents' decisions. Our key idea is the use of \textit{factorized} latent variables to model decoupled agent decisions even though agent dynamics are coupled. Factorization across agents and time enable us to \emph{query} the effects of changing an arbitrary agent's decision at an arbitrary time step. Our contributions are:
\begin{enumerate} \itemsep0em
    \item \textbf{State-of-the-art multi-agent vehicle forecasting:}
    We develop a multi-agent forecasting model called Estimating Social-forecast Probabilities (ESP) that uses \textit{exact} likelihood inference (unlike VAEs or GANs) to outperform three state-of-the-art methods on real and simulated vehicle datasets \cite{nuscenes,dosovitskiy17carla}.
    \item \textbf{Goal-conditioned multi-agent forecasting:} 
    We present the \emph{first} generative multi-agent forecasting method able to condition on agent goals, called PREdiction Conditioned on Goals (PRECOG). After modelling agent interactions, conditioning on one agent's goal alters the predictions of other agents. 
    \item \textbf{Multi-agent imitative planning objective:}
    We derive a data-driven objective for motion planning in multi-agent environments. It balances the likelihood of reaching a goal with the probability that expert demonstrators would execute the same plan. We use this objective for offline planning to known goals, which improves forecasting performance.
\end{enumerate}

\section{Related Work}\label{sct:relatedwork}

Multi-agent modeling and forecasting is a challenging problem for control applications in which agents react to each other concurrently. Safe control requires faithful models of reality to anticipate dangerous situations before they occur. Modeling dependencies between agents is especially critical in tightly-coupled scenarios such as intersections.

\noindent {\bf Game-theoretic planning:} Traditionally, multi-agent planning and game theory approaches  explicitly model multiple agents' policies or internal states, usually by generalizing Markov decision processes (MDPs) to multiple decisions makers \cite{claus1998dynamics,tan1993multi}. These frameworks facilitate reasoning about collaboration strategies, but suffer from ``state space explosion'' intractability except when interactions are known to be sparse \cite{melo2011decentralized} or hierarchically decomposable \cite{fisac2018hierarchical}.

\noindent {\bf Multi-agent forecasting:} Data-driven approaches have been applied to forecast complex interactions between multiple pedestrians \cite{alahi2016social,bartoli2017context,fernando2018soft,gupta2018social,ma2017forecasting}, vehicles \cite{deo2018multi,lee2017desire,park2018sequence}, and athletes \cite{felsen2018will,le2017coordinated,lee2016predicting,sun2019stochastic,zhan2018generative,zhao2019multi}. These methods attempt to generalize from previously observed interactions to predict multi-agent behavior in new situations. Forecasting is related to  Imitation Learning \cite{osa2018algorithmic}, which learns a model to mimic demonstrated behavior. In contrast to some Imitation Learning methods, e.g. behavior cloning \cite{pomerleau1989alvinn}, behavior forecasting models are not executed in the environment of the observed agent -- they are instead predictive models of the agent. In this sense, forecasting can be considered non-interactive Imitation Learning without execution.


\noindent {\bf Forecasting for control and planning:}
Generative models for multi-agent forecasting and control have been proposed.
In terms of multi-agent forecasting, our work is related to \cite{schmerling2018multimodal} which uses a conditional VAE \cite{kingma2013auto} encoding of the joint states of multiple agents together with recurrent cells to predict future human actions. However, our work differs in three crucial ways.
First, we model continual co-influence between agents, versus ``robot-only influence'' where a robot's responses to the human are not modeled.
Second, our method uses contextual visual information useful for generalization to many new scenes. 
Third, we model interactions between more than two vehicles \textit{jointly}. While \cite{ivanovic2018generative} assumes conditional independencies for computational reasons, we do not, as they impose minimal overhead.

We consider scenarios in which the model may control one of the agents (a ``robot''). In terms of planned control, our method generalizes imitative models \cite{rhinehart2018deep}. In \cite{rhinehart2018deep}, single-agent forecasting models are used for deterministic single-agent planning. Our work instead considers multi-agent forecasting, and therefore must plan over a distribution of possible paths: from our robot's perspective, the future actions of other human drivers are uncertain. By modeling co-influence, our robot's
trajectory are conditional on the (uncertain) future human trajectories, and therefore future robots states are necessarily uncertain. Thus, our work proposes a nontrivial extension for imitative models: we consider future path planning uncertainty induced by the uncertain actions of other agents in a multi-agent setting. While \cite{rhinehart2018deep} could implicitly model other agents through its visual conditioning, we show explicit modeling of other agents yields better forecasting results, in addition to giving us the tools to predict responses to agent's plans.

\section{Deep Multi-Agent Forecasting}

In this section, we will describe our likelihood-based model for multi-agent forecasting, and then describe how we use it to perform planning and multi-agent conditional forecasting. First, we define our notation and terminology. We treat our multi-agent system as a continuous-space, discrete-time, partially-observed Markov process, composed of $A$ agents (vehicles) that interact over $T$ time steps. We model all agent positions at time $t$ as $\State_t \in \mathbb{R}^{A\times D}$, where $D\!=\!2$. $\State_t^a$ represents agent $a$'s $(x,y)$ coordinates on the ground plane. We assume there is one ``robot agent'' (e.g. the AV) and $A\!-\!1$ ``human agents'' (e.g. human drivers that our model cannot control). We define $\State_t^r \doteq \State_t^1\in\mathbb{R}^{D}$ to index the robot state, and $\State_t^h \doteq  \State_t^{2:A}\in\mathbb{R}^{(A-1)\times D}$ to index the human states.
Bold font distinguishes variables from functions. Capital English letters denote random variables. We define $t\!=\!0$ to be the current time. Subscript absence denotes \textit{all} future time steps, and superscript absence denotes \emph{all} agents, e.g.\ $\State \doteq \State_{ 1:T }^{1:A} \in \mathbb{R}^{T\times A\times D}$. 

Each agent has access to environment perception $\phi \doteq \{\state_{-\tau:0}, \bchi\}$, where $\tau$ is the number of past multi-agent positions we condition on and $\bchi$ is a high-dimensional observation of the scene. $\bchi$ might represent LIDAR or camera images, and is the robot's observation of the world. In our setting, LIDAR is provided as $\bchi=\mathbb R^{200 \times 200 \times 2}$, with $\bchi_{ij}$ representing a 2-bin histogram of points above and at ground level in $0.5\mathrm{m}^2$ cells. Although our perception is robot-centric, each agent is modeled to have access to $\bchi$. 

\subsection{Estimating Social-forecast Probability (ESP)}\label{sct:model}
We propose a data-driven likelihood-based generative model of multi-agent interaction to probabilistically predict $T$-step dynamics of a multi-agent system: $ \State\alltime\!\sim\!q(\State\alltime|\phi;\!\data)$,
 where $\data$ is training data of observed multi-agent state trajectories. 
Our model learns to map latent variables $\Latent\alltime$ via an invertible function $f$ to multi-agent trajectories $\State$ conditioned on $\phi$. $f$'s invertibility induces $q(\State\alltime|\phi)$, a \emph{pushforward distribution} \cite{mccann1995existence}, also known as an \emph{invertible generative model} \cite{dinh2016realnvp,grathwohl2018ffjord,guan2019generative,kingma2018glow,rhinehart2018r2p2}. Invertible generative models can efficiently and exactly compute probabilities of samples. Here, it means we can compute the probability of joint multi-agent trajectories, critical to our goal of \emph{planning} with the model. We name the model ``Estimating Social-forecast Probabilities'' (ESP). $\State\alltime$ is sampled from $q$ as follows:
\begin{equation}
\textstyle \Latent\alltime \sim \mathcal{N}(\Zero, \Eye); \quad \State\alltime = f(\Latent\alltime; \phi); \quad \State,\Latent\alltime\in\mathbb{R}^{T\times A\times D}. \label{eqn:flowfull}
\end{equation}
Our latent variables $\Latent\doteq\Latent_{ 1:T }^{1:A}$ 
factorize across agents and time, which allows us to \emph{decide} agent $a$'s reaction at time $t$ by setting $\Latent_t^a\leftarrow\latent_t^a$, discussed later.  Our model is related to the  R2P2 single-agent generative model \cite{rhinehart2018r2p2}, which constructs a deep likelihood-based generative model for single-agent vehicle forecasting. For multi-step prediction, we generalize R2P2's autoregressive one-step single-agent prediction for the multi-agent setting, 
and assume a one-step time delay for agents to react to each other:
\begin{equation}
\textstyle \State_{t}^a \;=\; \mu_\theta^a(\State_{1:t-1}, \phi) + \sigma_\theta^a(\State_{1:t-1}, \phi)\cdot\Latent_{t}^a \; \in \mathbb{R}^{D}, \label{eqn:flowperagent}
\end{equation}
where $\mu_\theta^a(\cdot)$ and $\sigma_\theta^a(\cdot)$ are neural network functions (with trainable weights $\theta$)
outputting a one-step mean prediction $\bmu_t^a \in \mathbb R^D$ and standard-deviation matrix $\bsigma_t^a \in \mathbb R^{D \times D}$ of agent $a$, 
defining the system's transition function $q$ as
\begin{equation}
\textstyle q(\State_{t} | \State_{1:t-1}, \phi) = \prod_{a=1}^{A} \mathcal{N} (\State_{t}^a; \bmu_{t}^a, \bSigma_{t}^a),
\end{equation}
where $\bSigma_{t}^a=\bsigma_{t}^a \bsigma_{t}^{a\top}$.
Note that \eqn{flowperagent} predicts the $a^{\text{th}}$ agent's state $\State_{t}^a$ given the previous \textit{multi-agent} states $\State_{1:t-1}$.
We can see that given $\State_{1:t-1}$, the one-step prediction in \eqn{flowperagent} is unimodal Gaussian.
However, multi-step predictions are generally multimodal given the recursive nonlinear conditioning of neural network outputs $\bmu_{t}^a$ and $\bsigma_{t}^a$ on previous predictions. The final joint of this model can be written as
\begin{align}
 \textstyle   q(\State\alltime|\phi) &= \textstyle \prod_{t=1}^{T}\!q(\State_{t}|\State_{1:t-1},\phi). \label{eqn:jointpdf} 
\end{align}

\begin{figure*}[ht]
    \centering
    \begin{subfigure}[b]{0.23\textwidth}
        \includegraphics[width=\textwidth]{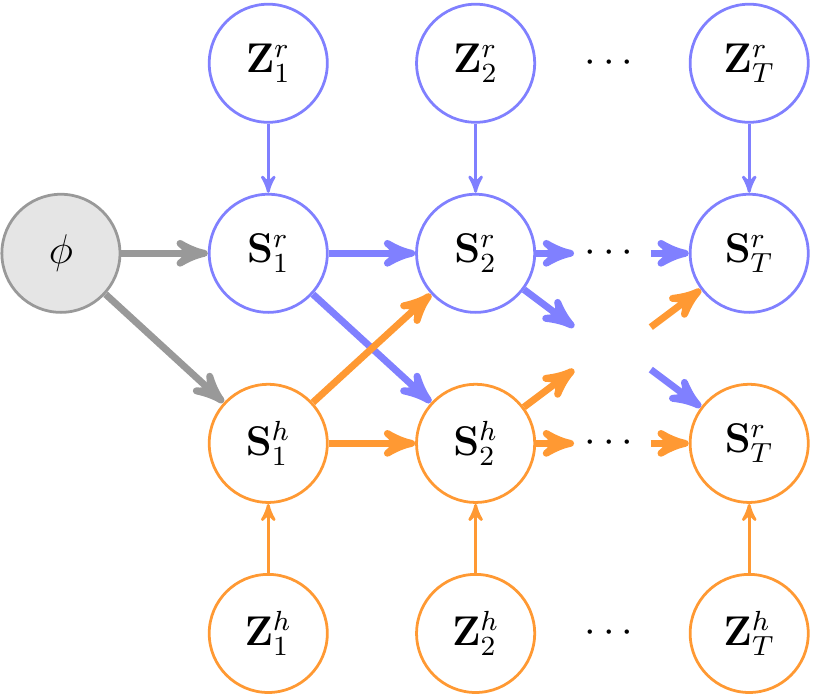}
        \caption{ESP forecasting}
        \label{fig:forecast-coinfluence}
    \end{subfigure}
    \hfill
    \begin{subfigure}[b]{0.23\textwidth}
        \includegraphics[width=\textwidth]{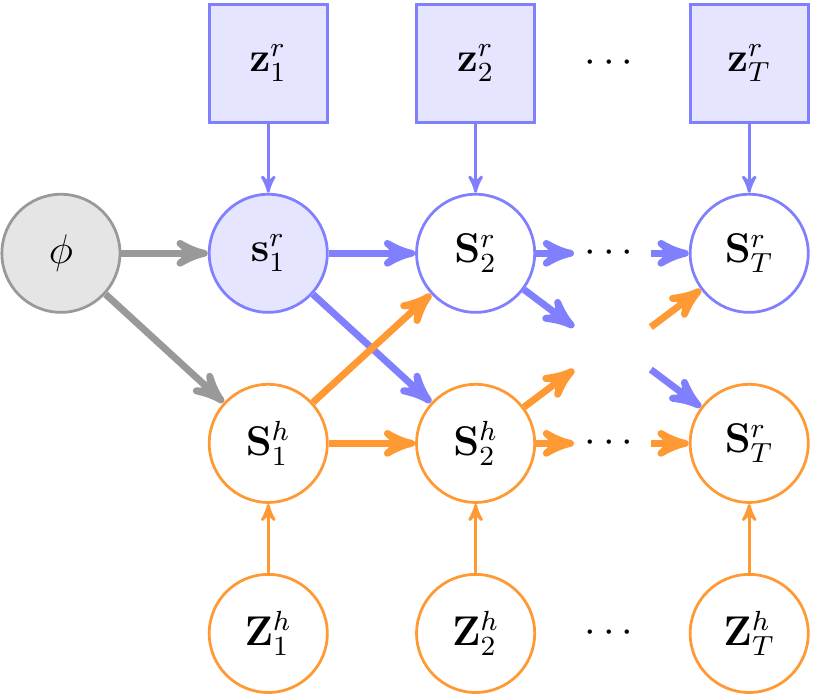}
        \caption{PRECOG planning}
        \label{fig:plan-coinfluence}
    \end{subfigure}
    \hfill
    \begin{subfigure}[b]{.5\textwidth}
      \includegraphics[width=.98\textwidth]{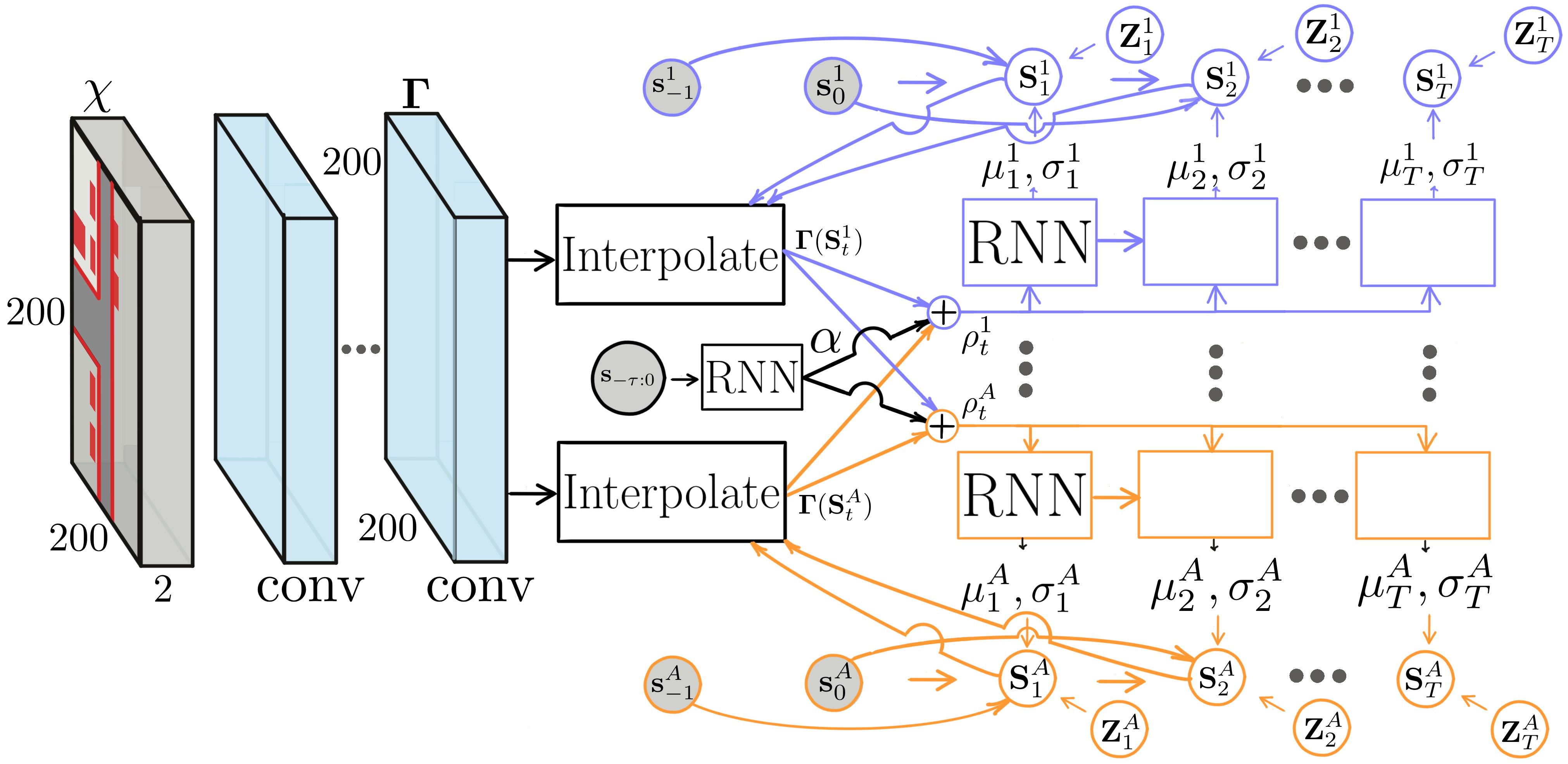}
    \caption{ESP model implementation} \label{fig:arch}
    \end{subfigure}
    \caption{
    Our factorized latent variable model of forecasting and planning shown for 2 agents. In \fig{forecast-coinfluence} our model uses latent variable $\Latent_{t+1}^a$ to represent variation in agent $a$'s \textit{plausible} scene-conditioned reactions to all agents $\State_t$, causing uncertainty in every agents' future states $\State\alltime$. Variation exists because of unknown driver goals and different driving styles observed in the training data. Beyond forecasting, our model admits \textit{planning} robot decisions by \textit{deciding} $\Latent^r\!=\!\latent^r$ (\fig{plan-coinfluence}). Shaded nodes represent observed or determined variables, and square nodes represent robot decisions \cite{barber2012bayesian}. 
    Thick arrows represent grouped dependencies of \textit{non-Makovian} $\State_t$ ``carried forward'' 
    (a regular edge exists between any pair of nodes linked by a chain of thick edges).
    Note $\Latent$  \textit{factorizes} across agents, isolating the robot's reaction variable $\latent^r$. Human reactions remain uncertain ($\Latent^h$ is unobserved) and uncontrollable (the robot cannot decide $\Latent^h$), and yet the robot's decisions $\latent^r$ will still \textit{influence} human drivers $\State^h_{2:T}$ (and vice-versa). \fig{arch} shows our implementation. See \app{implementation} for details.
    }
    \label{fig:forecast-and-planning-graph} 
\end{figure*} 
\vspace{-10pt}
\subsection{Model Implementation}\label{sct:implementation}  
To implement our model $q(\State\alltime|\phi)$, we design neural networks that output $\bmu_t^a$ and $\bsigma_t^a$. Similar to \cite{rhinehart2018r2p2}, we expand $\mu_\theta^a(\cdot)$ to represent a ``Verlet'' step, which predicts a constant-velocity mean when $\netm_t^a=m^a_\theta(\State_{1:t\!-\!1}, \!\phi)=0$:
\begin{equation}
 \textstyle \State^a_{t}= \underbrace{2\State^a_{t\!-\!1}\!\!-\!\State^a_{t\!-\!2}\!\!+\!m^a_\theta(\State_{1:t\!-\!1}, \!\phi)}_{\bmu_{t}^a}+\underbrace{\sigma^a_\theta(\State_{1:t\!-\!1},\!\phi)}_{\bsigma_{t}^a}\cdot\Latent^a_{t}. \label{eqn:verlet}
\end{equation}
A high-level diagram of our implementation shown in \fig{arch}.
Recall $\phi\!=\!\{\state_{-\tau:0}, \bchi\}$: the context contains the past positions of all agents, $\state_{-\tau:0}$, and a feature map $\bchi$, implemented as LIDAR observed by the robot.
We encode $\state_{-\tau:0}$ with a GRU. A CNN processes $\bchi$ to $\bGamma$ at the same spatial resolution as $\bchi$. Features for each agent's predicted position $\State_t^a$ are computed by interpolating into $\bGamma$ as $\bGamma(\State_t^a)$. Positional ``social features'' for agent $a$ are computed: $\State_t^{a}\!-\!\State_t^{b}~\forall~b\!\in\!A\!\setminus\!\{a\}$, as well as visual ``social features'' $\gamma_t^a=\bGamma(\state_t^1)\oplus\dots\oplus\bGamma(\state_t^A)$. The social features, past encoding, and CNN features are fed to a per-agent GRU, which produces $\netm^a_t$ and $\bsigma_t^a$ in \eqn{verlet}. We train with observations of expert multi-agent interaction $\State^{*} \sim p(\State^*|\phi)$  by maximizing likelihood with respect to our model parameters $\theta$. We use shared parameters to produce $\bGamma$ and the past encoding. See \app{implementation} for an architecture table and other details.

\noindent{\bf Flexible-count implementation}: While the implementation described so far is limited to predict for a \emph{fixed-count} of agents in a scene, we also implemented a \emph{flexible-count} version. There are two flavors of a model that is flexible in practice. (1) A fully-flexible model applicable to any scene with agent count $A_{\text{test}} \in \mathbb N$. (2) A partially-flexible model applicable to any scene with agent count $A_{\text{test}} \in \{1..A_{\text{train}}\}$, controlled by a hyperparameter upper-bound $A_{\text{train}}$ set at training time. To implement (1), the count of model parameters must be \emph{independent} of $A_{\text{test}}$ in order for the same architecture to apply to scenes with different counts of agents. To implement (2), ``missing agents'' must not affect the \emph{joint distribution over the existing agents}, equivalent to ensuring $\nicefrac{\partial \State^{\text{existing}}}{\partial \Latent^{\text{missing}}}\!=\!0$ in our framework. We implemented (2) by using a mask $M\!\in\!\{0,1\}^{A_{\text{train}}}$ to mask features of missing agents. In this model, we shared parameters across agents, and trained it on data with varying counts of agents.

\subsection{Conditional Forecasting }

A distinguishing feature of our generative model for multi-step, multi-agent prediction is its latent variables $\Latent\alltime\doteq\Latent_{ 1:T }^{1:A}$ that factorizes over agents and time. Factorization makes it possible to use the model for highly flexible conditional forecasts. 
Conditional forecasts predict how other agents would likely respond to different robot decisions at different moments in time.
Since robots are not merely passive observers, but one of potentially many agents, the ability to anticipate how they affect others is critical to their ability to plan useful, safe, and effective actions, critical to their utility within a planning and control framework \cite{mcallister2017concrete}.

Human drivers can appear to take highly stochastic actions in part because we cannot observe their goals. In our model, the source of this uncertainty comes from the latent variables $\Latent\alltime\!\sim\!\mathcal N(\Zero,\Eye)$. In practical scenarios, the robot knows its own goals, can choose its own actions, and can plan a course of action to achieve a desired goal.
Recall from \eqn{flowperagent} that \textit{one-step} agent predictions are conditionally independent from each other give the previous multi-agent states. 
Therefore, certainty in the latent state $\Latent_t^a$ corresponds to certainty of the $a^{\text{th}}$ agent's \textit{reaction} at time $t$ to the multi-agent system history $\State_{1:t-1}$.
Different values of $\Latent_t^a$ correspond to different ways of reacting to the same information.
Deciding values of $\Latent_t^a$ corresponds to controlling the agent $a$.
We can therefore implement control of the robot via assigning values to its latent variables $\Latent^r\alltime \leftarrow \latent^r\alltime$.
In contrast, human reactions $\Latent_t^h$ cannot be decided by the robot, 
but remain uncertain from the robot's perspective.
Thus, humans can only be influenced by their conditioning on the robot's previous states in $\State_{1:t-1}$, as seen \fig{plan-coinfluence}.
Therefore, to generate conditional-forecasts,
we decide $\latent^r\alltime$, sample $\Latent\alltime^h$, concatenate $\Latent\alltime\!=\!\latent\alltime^r\oplus\Latent\alltime^h$, and warp $\State\alltime\!=\!f(\Latent\alltime, \phi)$. This factorization of latent variables easily facilitates conditional forecasting. 
To forecast $\State\alltime$, we can fix $\latent^r\alltime$ while sampling the human agents' reactions from their distribution $p(\Latent^h\alltime)\!=\!\mathcal{N}(\Zero,\Eye)$, which are warped via \eqn{flowfull}.

\subsection{\hbox{PREdiction Conditioned On Goals (PRECOG)}}

We discussed how forecasting can condition on a value of $\latent^r\alltime$, but not yet how to find \emph{desirable} values of $\latent^r\alltime$, e.g. values that would safely direct the robot towards its goal location. We perform multi-agent planning by optimizing an objective $\mathcal{L}$ \wrt the control variables $\latent^r\alltime$, which allows us to produce the ``best'' forecasts under $\mathcal{L}$. 

While many objectives are valid, we use imitative models (IM), which estimate the likeliest state trajectory an expert ``would have taken'' to satisfy a goal, based on prior expert demonstrations \cite{rhinehart2018deep}. IM modeled single-agent environments where robot trajectories are planned without consideration of other agents. Multi-agent planning is different, because future robot states are uncertain (states $\State^r_{t>1}$ in \fig{plan-coinfluence}), even when conditioned on control variables $\latent^r\alltime$, because of the uncertainty in surrounding human drivers $\Latent^h\alltime$. 

We generalize IM to multi-agent environments, and plan \wrt the uncertainty of human drivers close by. 
First, we chose a ``goal likelihood'' function that represents the likelihood that a robot reaches its goal $\mathcal{G}$ given state trajectory $\State\alltime$. For instance, the likelihood could be a waypoint $\waypoint\!\in\!\mathbb{R}^D$ the robot should approach: $p(\mathcal{G} | \State\alltime,\phi)\!=\!\mathcal{N}(\waypoint;\State^r_T, \epsilon \Eye)$.
Second, we combine the goal likelihood with a ``prior probability'' model of safe multi-agent state trajectories $q(\State\alltime| \phi)$, learned from expert demonstrations. Note that unlike many other generative multi-agent models, we can compute the probability of generating $\State$ from $q(\State\alltime| \phi)$ exactly, which is critical to our planning approach. This results in a ``posterior'' $p(\State\alltime | \mathcal{G},\phi)$.
Finally, we plan a goal-seeking path in the learned distribution of demonstrated multi-agent behavior under the log-posterior probability derived as:
\begin{align}
\textstyle \log &\textstyle\,\expectation{\Latent^h\alltime}{p(\State\alltime | \mathcal{G},\phi)} 
\textstyle \;\geq\;\textstyle\expectation{\Latent^h\alltime}{\log p(\State\alltime | \mathcal{G},\phi)} \label{eqn:jensens} \\
    &= \expectation{\Latent^h\alltime}{\log \big(q(\State\alltime | \phi)p(\mathcal{G} | \State\alltime,\phi)\big)}\!-\!\log p(\mathcal{G} | \phi) \label{eqn:bayes_rule} \\ 
\mathcal{L}(\latent^r,\mathcal{G})
       &\doteq \expectation{\Latent^h\alltime}{\log q(\State\alltime| \phi) + \log p(\mathcal G| \State\alltime, \phi)} \label{eqn:objectiveS} \\
&= \expectation{\Latent^h\alltime}{\log\!\!\underbrace{q(f(\Latent\alltime)| \phi)}_{\text{multi-agent prior}}\!\!+ \log \underbrace{p(\mathcal G| f(\Latent\alltime), \phi)}_{\text{goal likelihood}}}, \label{eqn:objectiveL}
\end{align}
where \eqn{jensens} follows by Jensen's inequality, 
which we use to avoid the numerical issue of a single sampled $\Latent^h$
dominating the batch.
\eqn{bayes_rule} follows from Bayes' rule and uses our learned model $q$ as the prior. In \eqn{objectiveS}, we drop $p(\mathcal{G}|\phi)$ because it is constant \wrt $\latent^r$. 
Recall $\Latent\alltime=\latent\alltime^r\oplus\Latent\alltime^h$ is the concatenation of robot and human control variables.
The robot can plan using our ESP model by optimizing \eqn{objectiveL}:
\begin{equation}
    \textstyle \latent^{r*} = \argmax_{\latent^r} \mathcal{L}(\latent^r, \mathcal{G}). \label{eqn:objectiveZ}
\end{equation}

Other objectives might be used instead, e.g. maximizing the posterior probability of the robot trajectories only. This may place human agents in unusual, precarious driving situations, outside the prior distribution of ``usual driving interaction''.~\eqn{objectiveZ} encourages the robot to avoid actions likely to put the joint system in an unexpected situation.

\section{Experiments}

We first compare our forecasting model against existing state-of-the-art multi-agent forecasting methods, including 
SocialGAN \cite{gupta2018social}, DESIRE \cite{lee2017desire}. We also include a baseline model: R2P2-MA (adapted from R2P2 \cite{rhinehart2018r2p2} to instead handle multiple agent inputs), which does not model how agents will react to each others' future decisions.
Second, we investigate the novel problem of \textit{conditional} forecasting. To quantify forecasting performance, we study scenarios where we have pairs of the robot's true goal and the sequence of joint states. Knowledge of goals should enable our model to better predict what the robot and each agent could do.
Third, we ablate the high-dimensional contextual input $\bchi$ from our model to determine its relevance to forecasting. Appendix~\ref{app:additional_evaluation} and~\ref{app:visualizations} provide: (1) more conditional forecasting results, (2) localization sensitivity analysis and mitigation (3) evaluations on more datasets, and (4) several pages of qualitative results. 

\noindent{{\bf nuScenes  dataset:}} We used the recently-released full nuScenes dataset \cite{nuscenes}, a real-world dataset for multi-agent trajectory forecasting, in which 850 episodes of 20 seconds of driving were recorded and labelled at 2Hz with the positions of all agents, and synced with many sensors, including LIDAR. We processed each of the examples to train, val, and test splits. Each example has 2 seconds of past and 4 seconds of future positions at 5Hz and is accompanied by a LIDAR map composited from 1 second of previous scans. We also experimented concatenating a binary road mask to $\chi$, indicated as ``Road'' in our evaluation.

\noindent{{\bf CARLA dataset:}} We generated a realistic dataset for multi-agent trajectory forecasting and planning with the CARLA simulator \cite{dosovitskiy17carla}. We ran the autopilot in {\tt Town01} for over 900 episodes of 100 seconds each in the presence of 100 other vehicles, and recorded the trajectory of every vehicle and the autopilot's LIDAR observation. We randomized episodes to either train, validation, or test sets. We created sets of $\num{60701}$ train, $\num{7586}$ validation, and $\num{7567}$ test examples, each with 2 seconds of past and 2 seconds of future positions at 10Hz.  See \app{carla_dataset} for details and \url{https://sites.google.com/view/precog} for data.


\subsection{Metrics}
\noindent{\bf Log-likelihood:} As our models can perform exact likelihood inference (unlike GANs or VAEs), we can precisely evaluate how likely held-out samples are under each model. Test log-likelihood is given by the forward cross-entropy $H(p,q)\!=\!-\mathbb E_{\State^* \sim p(\State^* | \phi)} \log q(\State^* | \phi)$, which is unbounded for general $p$ and $q$. However, by perturbing samples from $p(\State^*|\phi)$ with noise drawn from a known distribution $\eta$ (e.g. a Gaussian) to produce a perturbed distribution $p'$, we can enforce a lower bound \cite{rhinehart2018r2p2}. The lower bound is given by $H(p',q) \geq H(p') \geq H(\eta)$. We use $\eta\!=\!\mathcal N(\Zero, 0.01\cdot \Eye)$ (n.b. $H(\eta)$ is known analytically). Our likelihood statistic is: 
\begin{equation}
\displaystyle \hat{e} \doteq \big[H(p',q) - H(\eta)\big] \nicefrac{}{(TAD)} \geq 0,  \label{eqn:ehat}
\end{equation}
which has $\nicefrac{\text{nats}}{\text{dim.}}$ units. We call $\hat e$ ``extra nats'' because it represents the (normalized) extra nats above the lower bound of $0$. Normalization enables comparison across models of different dimensionalities. 

\noindent{\bf Sample quality:} For sample metrics, we must take care not to penalize the distribution when it generates plausible samples different than the expert trajectory.
We extend the ``minMSD'' metric \cite{lee2017desire,park2018sequence,rhinehart2018r2p2} to measure quality of \emph{joint trajectory samples}. The ``minMSD'' metric samples a model and computes the error of the best sample in terms of MSD. In contrast to the commonly-used average displacement error (ADE) and final displacement error (FDE) metrics that computes the mean Euclidean error from a batch of samples to a \emph{single} ground-truth sample \cite{alahi2016social,deo2018multi,fernando2018soft,gupta2018social,pellegrini2009you}, minMSD has the desirable property of not penalizing plausible samples that correspond to decisions the agents could have made, but did not. \emph{This prevents erroneously penalizing models that make diverse behavior predictions}. We hope other multimodal prediction methods will also measure the quality of joint samples with minMSD, given by:
\vspace{-2mm}
\begin{equation}
\begin{gathered}
\textstyle \hat{m}_{K} \doteq \mathbb{E}_{\State^*} \underset{k\in\{1..K\}}{\min}||\State^*\alltime-\State\alltime^{(k)}||^2 \nicefrac{}{(TA)}, \label{eqn:mhat}
\end{gathered}
\end{equation}
\vspace{-3mm}

\noindent where $\State^*\!\sim\!p(\State^* | \phi), \State\alltime^{(k)}\,\iidsim\,q(\State|\phi)$. We denote the per-agent error of the best \emph{joint} trajectory with
\vspace{-2mm}
\begin{equation}
\begin{gathered}
\textstyle \hat{m}_K^a \doteq \mathbb{E}_{\State^* \sim p(\State^* | \phi)} ||\State\alltime^{*a} - \State\alltime^{a,(k^\dagger)}||^2 \nicefrac{}{T},\\
\textstyle k^\dagger \doteq \argmin_{k\in\{1..K\}} ||\State^*\alltime -  \State\alltime^{(k)}||^2.
\end{gathered}
\end{equation}
\vspace{-6mm}

\newcommand\hfilll{\hspace{0pt plus 1filll}}

\begin{table*}[h!]
\centering
\caption{CARLA and nuScenes multi-agent forecasting evaluation.
All CARLA-trained models use {\tt Town01 Train} only, and are tested on {\tt Town02 Test}. No training data is collected from {\tt Town02}. Means and their standard errors are reported. The en-dash (--) indicates an approach unable to compute $\hat e$. The R2P2-MA model generalizes \cite{rhinehart2018r2p2} to multi-agent. Variants of our ESP method (gray) outperform prior work. For additional evaluations on {\tt Town01 Test} and single agent settings, see \app{additional_evaluation}.}
\label{tab:carla_forecast} 
\resizebox{1.0\textwidth}{!}{
\begin{tabular}{lccccccccccc}
\toprule
Approach  & Test $\hat{m}_{K\!=\!12}$ & Test $\hat{e}$    &  & Test $\hat{m}_{K\!=\!12}$ & Test $\hat{e}$     &  & Test $\hat{m}_{K\!=\!12}$ & Test $\hat{e}$ &  & Test $\hat{m}_{K\!=\!12}$ & Test $\hat{e}$   \\
\midrule
\textbf{CARLA Town02 Test}                 & \multicolumn{2}{c}{2 agents} &  & \multicolumn{2}{c}{3 agents} & & \multicolumn{2}{c}{4 agents} &  & \multicolumn{2}{c}{5 agents} \\
                     \cmidrule{2-3}                                     \cmidrule{5-6}                                      \cmidrule{8-9}    \cmidrule{11-12}
                     KDE & $\num{4.488118} \pm \num{0.145325}$& $\num{8.17915} \pm \num{1.52268}$ & & $\num{5.963687} \pm \num{0.099377}$ & $\num{6.02873} \pm \num{0.39432}$ && $\num{7.845975} \pm \num{0.086886}$ & $\num{5.18057} \pm \num{0.17181} $ && $\num{9.610110} \pm \num{0.078066}$ & $\num{5.11585} \pm \num{0.09696}$ \\
DESIRE \hfilll \cite{lee2017desire} &  $\num{1.158645} \pm \num{0.027208}$ & --  & & $\num{1.099186} \pm \num{0.018021}$ & -- & & $\num{1.409834} \pm \num{0.018145}$ & -- & & $\num{1.696926} \pm \num{0.016550}$ & -- \\
SocialGAN \hfilll \cite{gupta2018social} & $\num{0.902} \pm \num{0.022}$ & --  & & $\num{0.756} \pm \num{0.015}$ & -- & & $\num{0.932} \pm \num{0.014}$ & -- & & $\num{0.979} \pm \num{0.015}$ & --\\
R2P2-MA \hfilll \cite{rhinehart2018r2p2} & $\num{0.454215} \pm \num{0.013709}$ & $\num{0.57730} \pm \num{0.00413}$ & & $\num{0.516012} \pm \num{0.012212}$ & $\num{0.63969} \pm \num{0.02200}$ & & $\num{0.574973} \pm \num{0.011463}$ &  $\num{0.59776} \pm \num{0.01009}$ & &  $\num{0.631876} \pm \num{0.010865}$ & $\num{0.61969} \pm \num{0.00987}$\\
\rowcolor{ourmethod}
Ours: ESP, no LIDAR                            & $\num{0.632597} \pm \num{0.016747}$ &  $\num{0.57947} \pm \num{0.00630}$ & & $\num{0.581810} \pm \num{0.013953}$ & $\num{0.62002} \pm \num{0.01284}$ &  & $\num{0.654508} \pm \num{0.012981}$ & $\num{0.59111} \pm \num{0.00640}$ & & $\num{0.783829} \pm \num{0.012790}$ & $\num{0.58369} \pm \num{0.00441}$ \\
\rowcolor{ourmethod}
Ours: ESP                                   & $\boldmath \num{0.3925} \pm \num{0.013545}$ & $ \num{0.54986} \pm \num{0.00440}$ & & $\boldmath \num{0.377020} \pm \num{0.010520}$ & $ \num{0.52881} \pm \num{0.00411}$ & & $ \num{0.438481} \pm \num{0.009752}$ & $ \num{0.54032} \pm \num{0.00425}$& & $ \num{0.564709} \pm \num{0.009450}$ & $ \num{0.59161} \pm \num{0.00360}$  \\
\rowcolor{ourmethod}
Ours: ESP, flex. count & $\num{0.488494} \pm \num{0.016999}$ & $\boldmath \num{0.536565} \pm \num{0.001593}$  &&  $\num{0.412276} \pm \num{0.012256}$ & $\boldmath \num{0.507822} \pm \num{0.001378}$ && $\boldmath \num{0.398196} \pm \num{0.010492}$ & $\boldmath \num{0.499244} \pm \num{0.001208}$ && $\boldmath \num{0.435110} \pm \num{0.010517}$ & $\boldmath \num{0.495612} \pm \num{0.001090}$\\
\midrule
\textbf{nuScenes Test}                   & \multicolumn{2}{c}{2 agents} &  & \multicolumn{2}{c}{3 agents} & & \multicolumn{2}{c}{4 agents} & & \multicolumn{2}{c}{5 agents} \\
                     \cmidrule{2-3}                                     \cmidrule{5-6}                                      \cmidrule{8-9}    \cmidrule{11-12}
                             KDE   & $\num{19.375} \pm \num{0.797753}$ &  $\num{3.7595} \pm \num{0.015212}$ &    & $\num{31.663} \pm \num{0.894048}$ & $\num{4.1017} \pm \num{0.023157}$ &   & $\num{41.289} \pm \num{1.169585}$& $\num{4.3686} \pm \num{0.026382}$ &  & $\num{52.071} \pm \num{1.449489}$ & $\num{4.615} \pm \num{0.027548}$\\
DESIRE \hfilll \cite{lee2017desire} 
& $\num{3.4729} \pm \num{0.101523}$& --
& &  $\num{4.421} \pm \num{0.130379}$ & -- & & $\num{5.9571} \pm \num{0.161714}$ & -- & & $\num{6.5752} \pm \num{0.198214}$ & -- \\
SocialGAN \hfilll \cite{gupta2018social}      
& $2.119 \pm 0.087$ & --
& &  $3.033 \pm 0.110$ & -- & & $3.484 \pm 0.129$ & -- & & $3.871 \pm 0.148$ & -- \\ 
R2P2-MA \hfilll \cite{rhinehart2018r2p2}    
&  $\num{1.3363} \pm \num{0.061696}$ & $\num{0.95051} \pm \num{0.007443}$ 
& & $\num{2.055} \pm \num{0.092943}$&	$\num{0.98917} \pm \num{0.008412}$ & & $\num{2.6948} \pm \num{0.100032}$ &	$\num{1.0201} \pm \num{0.01139}$ & & $\num{3.3109} \pm \num{0.165694}$	& $\num{1.0495} \pm \num{0.012277}$\\
\rowcolor{ourmethod}
Ours: ESP, no LIDAR                          
&  $\num{1.4964} \pm \num{0.068677}$ & $\boldmath \num{0.91982} \pm \num{0.007977}$
& & $\num{2.2401} \pm \num{0.083811}$ &	$\boldmath \num{0.95521} \pm \num{0.008388}$ & & $\num{3.2006} \pm \num{0.113273}$	 & $\num{1.0328} \pm \num{0.011836}$ & & $\num{3.4419} \pm \num{0.138938}$&	$\num{1.1065} \pm \num{0.018179}$ \\
\rowcolor{ourmethod}
Ours: ESP  
& $\num{1.3248} \pm \num{0.064942}$ & $\num{0.93349} \pm \num{0.007746}$
& & $\num{1.7048} \pm \num{0.089497}$ &	$\num{1.0183} \pm \num{0.010974}$ & & $\num{2.5466} \pm \num{0.095264}$ &	$\num{1.0533} \pm \num{0.015243}$&  &$\num{3.266} \pm \num{0.155313}$	& $\num{1.0819} \pm \num{0.01331}$ \\
\rowcolor{ourmethod}
Ours: ESP, Road & $\boldmath \num{1.0809} \pm \num{0.052563}$ & $\num{0.92875} \pm \num{0.008385}$& & $\boldmath \num{1.5053} \pm \num{0.070451}$	 & $\num{1.016} \pm \num{0.010782}$ & & $\boldmath \num{2.3599} \pm \num{0.093448}$&	$\boldmath \num{1.0134} \pm \num{0.011897}$ & & $\boldmath \num{2.8924} \pm \num{0.162437}$ &	$\num{1.1141} \pm \num{0.023551}$ \\
\rowcolor{ourmethod}
    Ours: ESP, Road, flex. & $\num{1.464281} \pm \num{0.067176}$ &  $\num{0.979780} \pm \num{0.002711}$ &&$\num{2.029230} \pm \num{0.079214}$ & $\num{1.001285} \pm \num{0.002858}$ && $\num{2.525027} \pm \num{0.099480}$ & $\num{1.014644} \pm \num{0.002467}$ && $\num{2.933192} \pm \num{0.128841}$ &  $\boldmath \num{1.028862} \pm \num{0.002387}$ \\
\bottomrule
\end{tabular}
}
\end{table*}
\subsection{Baselines}
\noindent{\bf KDE} \cite{parzen1962estimation,rosenblatt1956remarks} serves as a useful performance bound on all methods; it can compute both $\hat m$ and $\hat e$. We selected a bandwidth using the validation data. Note KDE ignores $\phi$. 

\noindent{\bf DESIRE} \cite{lee2017desire} proposed a conditional VAE model that observes past trajectories and visual context. We followed the implementation as described. Whereas DESIRE is trained with a single-agent evidence lower bound (ELBO), our model jointly models multiple agents with an exact likelihood.  DESIRE cannot compute joint likelihood or $\hat e$.

\noindent{\bf SocialGAN} \cite{gupta2018social} proposed a conditional GAN multi-agent forecasting model that observes the past trajectories of all modeled agents, but not $\bchi$. We used the authors' public implementation. In contrast to SocialGAN, we model joint trajectories and can compute likelihoods (and therefore $\hat e$). 

\noindent{\bf R2P2} \cite{rhinehart2018r2p2} proposed a likelihood-based conditional generative forecasting model for single-agents. We extend R2P2 to the multi-agent setting and use it as our R2P2-MA model; R2P2 does not jointly model agents. We otherwise followed the implementation as described. We trained it and our model with the forward-cross entropy loss. R2P2-MA's likelihood is given by $q(\State|\phi)=\prod_{a=1}^Aq^a(\State^a|\phi)$.

\subsection{Multi-Agent Forecasting Experiments}
\noindent{\bf Didactic Example:} In the didactic example, a robot (blue) and a human (orange) both navigate in an intersection, the human has a stochastic goal: with $0.5$ probability they will turn left, and otherwise they will drive straight. The human always travels straight for $4$ time steps, and then reveals its intention by either going straight or left. The robot attempts to drive straight, but will acquiesce to the human if the human turns in front of the robot. We trained our models and evaluate them in \fig{social_cross}.
Each trajectory has length $T\!=\!20$. While both models closely match the training distribution in terms of likelihood, their sample qualities are significantly different. The R2P2-MA model generates samples that crash $50\%$ of the time, because it does not condition future positions for the robot on future positions of the human, and vice-versa. In the ESP model, the robot is able to react to the human's decision during the generation process by choosing to turn when the human turns.

\noindent{\bf CARLA and nuScenes:} We build 10 datasets from CARLA and nuScenes data, corresponding to modeling different numbers of agents $\{2..5\}$. Agents are sorted by their distances to the autopilot, at $t\!=\!0$. When 1 agent is included, only the autopilot is modeled; for $A$ agents, the autopilot and the $A\!-\!1$ closest vehicles are modeled.

For each method, we report its best test-set score at the best val-set score.
In R2P2 and our method, the val-set score is $\hat e$. In DESIRE and SocialGAN, the val-set score is $\hat m$, as they cannot compute $\hat e$. \tab{carla_forecast} shows the multi-agent forecasting results. \textbf{Across all $10$ settings, our model achieves the best $\hat m$ and $\hat e$ scores}. We also ablated our model's access to $\bchi$ (``ESP, no LIDAR''), which puts it on equal footing with SocialGAN, in terms of model inputs. Visual context provides a uniform improvement in every case.

Qualitative examples of our forecasts are shown in \fig{example_forecast}. We observe three important types of multimodality: 1) multimodality in speed along a common specific direction, 2) the model properly predicts diverse plausible paths at intersections, and 3) when the agents are stopped, the model predicts sometimes the agents will stay still, and sometimes they will accelerate forward. The model also captures qualitative social behaviors, such as predicting that one car will wait for another before accelerating. See \app{visualizations} for additional visualizations.

\newcommand{\scwidth}[0]{.24\columnwidth}
\begin{figure}[hb]
    \centering
    \begin{subfigure}{\scwidth}
     \begin{overpic}[width=\textwidth]{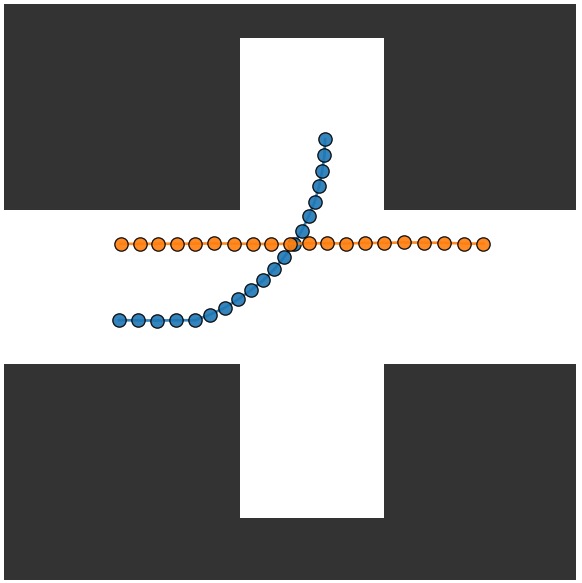}
    \put(1,88){{\setlength{\fboxsep}{1pt}\fontsize{8pt}{0pt}\selectfont\transparent{1.0}\colorbox{scrossgrey}{\textcolor{white}{\texttransparent{1.0}{R2P2-MA\vphantom{g}}}}}}
    \end{overpic}   
    \end{subfigure}
    \begin{subfigure}{\scwidth}
    \begin{overpic}[width=\textwidth]{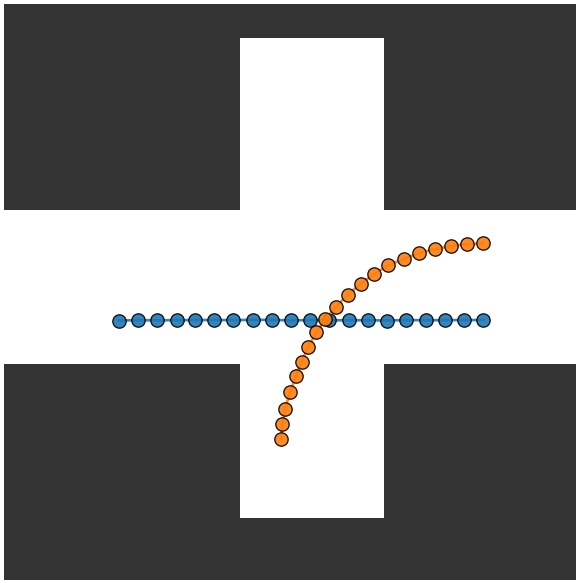}
    \put(1,88){{\setlength{\fboxsep}{1pt}\fontsize{8pt}{0pt}\selectfont\transparent{1.0}\colorbox{scrossgrey}{\textcolor{white}{\texttransparent{1.0}{R2P2-MA\vphantom{g}}}}}}
    \end{overpic}
    \end{subfigure}
    \begin{subfigure}{\scwidth}
    \begin{overpic}[width=\textwidth]{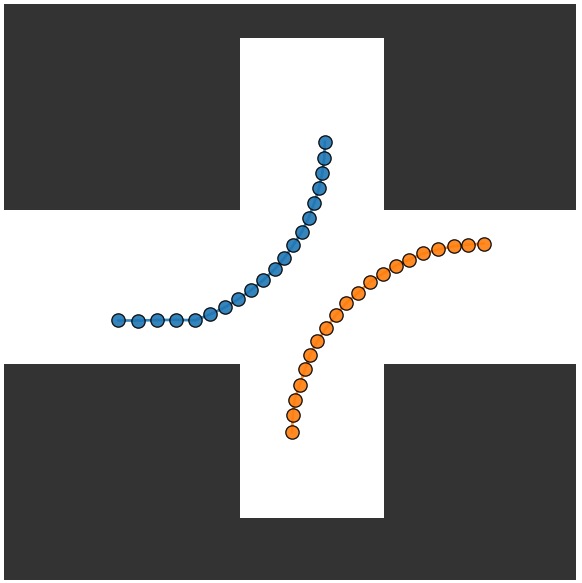}
        \put(1,88){{\setlength{\fboxsep}{1pt}\fontsize{8pt}{0pt}\selectfont\transparent{1.0}\colorbox{scrossgrey}{\textcolor{white}{\texttransparent{1.0}{ESP\vphantom{g}}}}}}
    \end{overpic}
    \end{subfigure}
    \begin{subfigure}{\scwidth}
    \begin{overpic}[width=\textwidth]{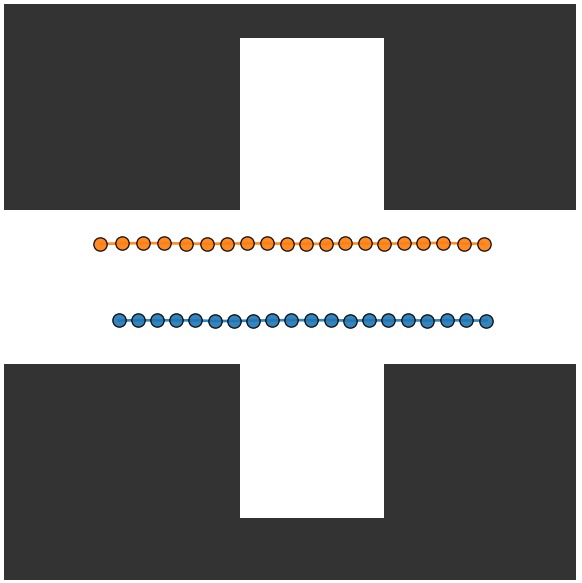}
            \put(1,88){{\setlength{\fboxsep}{1pt}\fontsize{8pt}{0pt}\selectfont\transparent{1.0}\colorbox{scrossgrey}{\textcolor{white}{\texttransparent{1.0}{ESP\vphantom{g}}}}}}
    \end{overpic}
    \end{subfigure}
    \resizebox{\columnwidth}{!}{
\begin{tabular}{lcccc}
\toprule
Model        &  Test $\hat{m}_{K=12}$ &  Test $\hat{e}$ & Forecasting crashes & Planning crashes  \\
\midrule
R2P2-MA &  $\num{0.3313}$ & $\num{0.085}$  & $50.8\%$ & $49.5\%$\\
ESP & ${\boldmath \num{0.0001}}$ & ${\boldmath \num{0.031}}$ & ${\bf{1.17}}\%$ & ${\bf {0.00}}\%$\\
\bottomrule
\end{tabular}
}
    \caption{Didactic evaluation. \emph{Left plots:} R2P2-MA cannot model agent interaction, and generates joint behaviors not present in the data. \emph{Right plots:} ESP allows agents to influence each other, and does not generate undesirable joint behaviors.}
    \label{fig:social_cross}
\end{figure}

\newcommand{\forcwidth}[0]{.24\textwidth}
\begin{figure*}[htp]
    \centering
    \FBox{ 
    \begin{subfigure}[t]{\forcwidth}
       \includegraphics[width=\textwidth,clip]{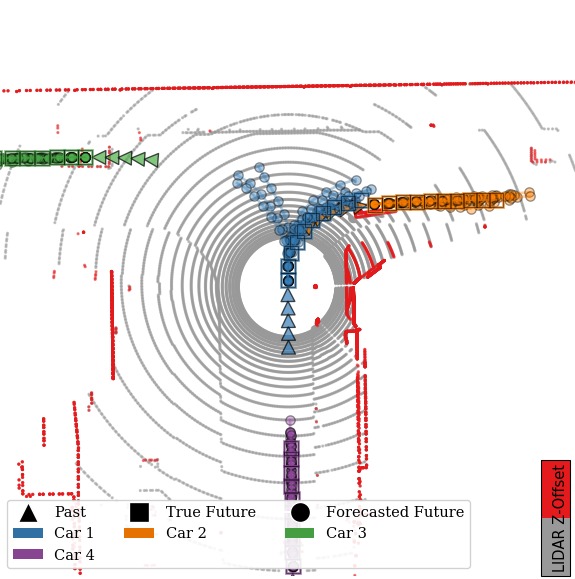}
       \end{subfigure}
    }
    \FBox{     \begin{subfigure}[t]{\forcwidth}
        \includegraphics[width=\textwidth]{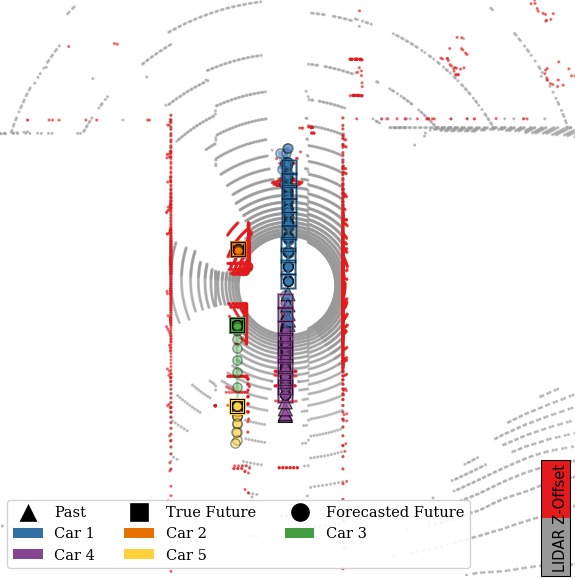}
    \end{subfigure} }
        \FBox{     \begin{subfigure}[t]{\forcwidth}
           \begin{overpic}[width=\textwidth]{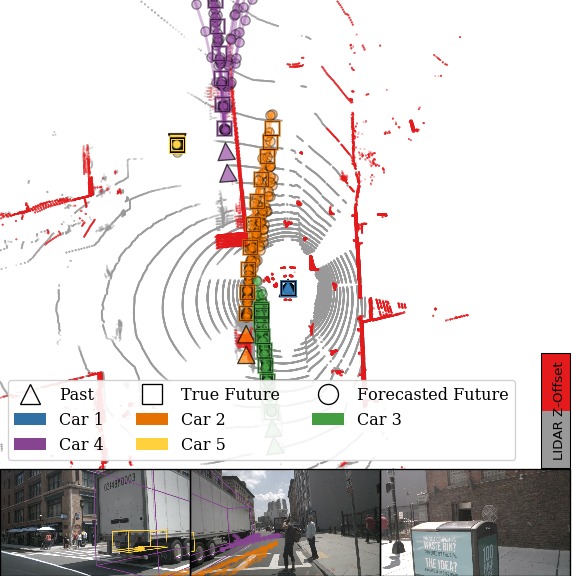}
      \put(0,15.5){{\setlength{\fboxsep}{1pt}\fontsize{5pt}{0pt}\selectfont\transparent{0.7}\colorbox{white}{\textcolor{black}{\texttransparent{1.0}{Left\vphantom{g}}}}}}
    \put(33,15.5){{\setlength{\fboxsep}{1pt}\fontsize{5pt}{0pt}\selectfont\transparent{0.7}\colorbox{white}{\textcolor{black}{\texttransparent{1.0}{Front\vphantom{g}}}}}}
    \put(66,15.5){{\setlength{\fboxsep}{1pt}\fontsize{5pt}{0pt}\selectfont\transparent{0.7}\colorbox{white}{\textcolor{black}{\texttransparent{1.0}{Right}}}}}
      \end{overpic} 
    \end{subfigure}     }
    \FBox{     \begin{subfigure}[t]{\forcwidth}
           \begin{overpic}[width=\textwidth]{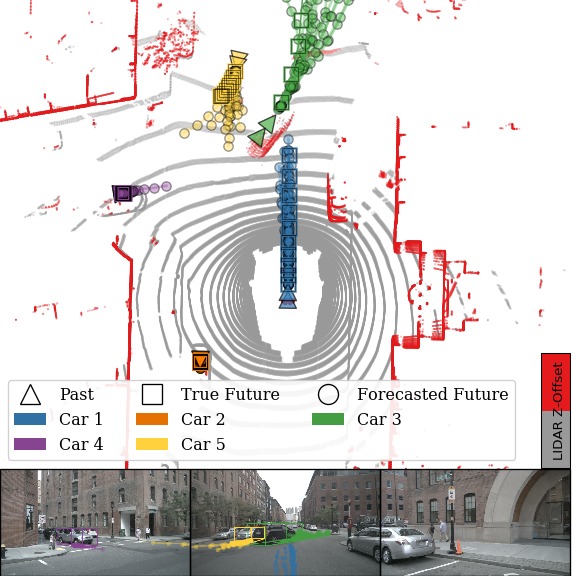}
      \put(0,15.5){{\setlength{\fboxsep}{1pt}\fontsize{5pt}{0pt}\selectfont\transparent{0.7}\colorbox{white}{\textcolor{black}{\texttransparent{1.0}{Left\vphantom{g}}}}}}
    \put(33,15.5){{\setlength{\fboxsep}{1pt}\fontsize{5pt}{0pt}\selectfont\transparent{0.7}\colorbox{white}{\textcolor{black}{\texttransparent{1.0}{Front\vphantom{g}}}}}}
    \put(66,15.5){{\setlength{\fboxsep}{1pt}\fontsize{5pt}{0pt}\selectfont\transparent{0.7}\colorbox{white}{\textcolor{black}{\texttransparent{1.0}{Right}}}}}
      \end{overpic} 
    \end{subfigure}     }
    \caption{
    Examples of multi-agent forecasting with our learned ESP model. In each scene, 12 joint samples are shown, and LIDAR colors are discretized to near-ground and above-ground.  \emph{Left:} (CARLA) the model predicts Car 1 could either turn left or right, while the other agents' future maintain multimodality in their speeds. \emph{Center-left:} The model predicts Car 2 will likely wait (it is blocked by Cars 3 and 5), and that Cars 3 and 5 sometimes move forward together, and sometimes stay stationary. \emph{Center-right:} Car 2 is predicted to overtake Car 1, which itself is forecasted to continue to wait for pedestrians and Car 2. \emph{Right:} Car 4 is predicted to wait for the other cars to clear the intersection, and Car 5 is predicted to either start turning or continue straight.
    } 
    \label{fig:example_forecast}
\end{figure*} 
\subsection{PRECOG Experiments} \label{sec:multiagentplanningexperiments}

Now we perform our second set of evaluations. We investigate if our planning approach enables us to sample more plausible joint futures of all agents. Unlike the previous unconditional forecasting scenario, when the robot is using the ESP model for planning, it knows its own goal. We can simulate planning offline by assuming the goal was the state that the robot actually reached at $t\!=\!T$, and then planning a path from the current time step to this goal position. We can then evaluate the quality of the agent's path and the stochastic paths of other agents under this plan. While this does not test our model in a full control scenario, it does allow us to evaluate whether conditioning on the goal provides more accurate and higher-confidence predictions.
We use our model's multi-agent prior \eqn{jointpdf} in the stochastic latent multi-agent planning objective \eqn{objectiveL}, and define the goal-likelihood $p(\mathcal G|\State\alltime,\phi)\!=\!\mathcal N(\State^r_{T} ; \State^{*r}_T, 0.1\!\cdot\!\Eye)$, \emph{i.e.} a normal distribution at the controlled agent's last true future position, $\State^{*r}_T$.
As discussed, this knowledge might be available in control scenarios where we are confident we can achieve this positional goal. Other goal-likelihoods could be applied to relax this assumption, but this setup allows us to easily measure the quality of the resulting joint samples. We use gradient-descent on \eqn{objectiveL} to approximate $\latent\alltime^{r*}$ (see supplement for details). The resulting latent plan yields highly likely joint trajectories under the uncertainty of other agents and approximately maximizes the goal-likelihood. Note that since we planned in latent space, the resulting robot trajectory is not fully determined -- it can evolve differently depending on the stochasticity of the other agents. We next illustrate a scenario where joint modeling is critical to accurate forecasting and planning. Then, we conduct planning experiments on the CARLA and nuScenes datasets.

\newcommand{\precogrow}[0]{\addlinespace[-0.1ex]}
\begin{figure*}[thbp]
\centering
\captionsetup{type=table} 
\caption{Forecasting evaluation of our model on CARLA {\tt Town01 Test} and nuScenes Test data. Planning the robot to a goal position (PRECOG) generates better predictions for all agents. Means and their standard errors are reported. See \tab{planning_forecast_app} for all $A=\{2..5\}$.}
\label{tab:carla_planning_forecast} 
\resizebox{.85\textwidth}{!}{
\begin{tabular}{clcccccc}
\toprule
Data & Approach & \multirow{1}{*}{Test $\hat{m}_{K=12}$} & \multirow{1}{*}{Test $\hat{m}_{K=12}^{a=1}$} & \multirow{1}{*}{Test $\hat{m}_{K=12}^{a=2}$} & \multirow{1}{*}{Test $\hat{m}_{K=12}^{a=3}$} & \multirow{1}{*}{Test $\hat{m}_{K=12}^{a=4}$} & \multirow{1}{*}{Test $\hat{m}_{K=12}^{a=5}$} \\ 
\midrule
\multirow{4}{*}{CARLA\, $A\!=\!2$} & DESIRE \cite{lee2017desire} & $\num{1.83703} \pm \num{0.04810}$ &	$\num{1.99149} \pm \num{0.06605}$ & $\num{1.68257} \pm \num{0.05013}$ & -- & -- & -- \\\precogrow
 & DESIRE-plan & $\num{1.85810} \pm \num{0.04614}$ & $\num{0.91834} \pm \num{0.04441}$ & $\num{2.79785} \pm \num{0.07331}$ & -- & -- & -- \\\precogrow
 
&  ESP & $\num{0.33705} \pm \num{0.0132}$ & $\num{	0.196} \pm \num{0.009}$ &$\num{0.478} \pm \num{0.024}$ & -- & -- & --\\ \precogrow
&  PRECOG & ${\boldmath \num{0.2406397} \pm \num{0.01207}}$	& ${\boldmath \num{0.055} \pm \num{0.003}}$ & ${\boldmath \num{0.426} \pm \num{0.024 }}$ & -- & -- & -- \\ \precogrow
\midrule
\multirow{4}{*}{CARLA\, $A\!=\!5$} & DESIRE \cite{lee2017desire} & $\num{2.62247} \pm \num{0.02953}$ & $\num{2.62066} \pm \num{0.04547}$ & $\num{2.42165} \pm \num{0.04842}$ & $\num{2.71018} \pm \num{0.06613}$ & $\num{2.96901} \pm \num{0.05700}$ & $\num{2.39086} \pm \num{0.04891}$ \\
& DESIRE-plan & $\num{2.32850} \pm \num{0.03837}$ & $\num{0.19446} \pm \num{0.00359}$ & $\num{2.23880} \pm \num{0.05707}$ & $\num{3.11885} \pm \num{0.09775}$ & $\num{3.33222} \pm \num{0.09021}$ & $\num{2.75818} \pm \num{0.08278}$ \\ 
& ESP & $\num{0.71759} \pm \num{0.01187}$ & $\num{0.34031} \pm \num{0.01146}$ & $\num{0.75908} \pm \num{0.02449}$ & $\num{0.80909} \pm \num{0.02504}$ & $\num{0.85122} \pm \num{0.02335}$ & $\num{0.82827} \pm \num{0.02411}$ \\ \precogrow
 & PRECOG & ${\boldmath \num{0.64039} \pm \num{0.01109}}$ & ${\boldmath \num{0.06557}\pm \num{0.00348}} $ & ${\boldmath \num{0.74088} \pm \num{0.02363}}$ & ${\boldmath \num{0.79019}\pm \num{0.02449}} $ & ${\boldmath \num{0.80444} \pm \num{0.02238}}$ & ${\boldmath \num{0.80085} \pm \num{0.02401}}$\\ \precogrow
\midrule
\multirow{4}{*}{nuScenes\, $A\!=\!2$} & DESIRE \cite{lee2017desire} & $\num{3.30741} \pm \num{0.09280}$ & $\num{3.00155} \pm \num{0.08819}$ & $\num{3.61328} \pm \num{0.13969}$ & -- & -- & -- \\ \precogrow
& DESIRE-plan & $\num{4.52804} \pm \num{0.15055}$ & $\num{0.45641} \pm \num{0.01544}$ & $\num{8.59966} \pm \num{0.29765}$ & -- &	-- &	-- \\\precogrow
& ESP & $\num{1.09373} \pm \num{0.05281}$ & $\num{0.95464} \pm \num{0.05651}$ & $\num{1.23282} \pm \num{0.07814}$ & -- &  -- & --\\ \precogrow
& PRECOG &  $\boldmath \num{0.51431} \pm \num{0.03667}$ & $\boldmath \num{0.15778} \pm \num{0.01621}$ & $\boldmath \num{0.87085} \pm \num{0.07002}$ & -- & -- & --\\ \precogrow
\midrule
\multirow{4}{*}{nuScenes\, $A\!=\!5$} & DESIRE \cite{lee2017desire} & $\num{6.82986} \pm \num{0.20435}$ & $\num{4.99886} \pm \num{0.21870}$ & $\num{6.41475} \pm \num{0.29442}$ & $\num{7.02696} \pm \num{0.36014}$ & 	$\num{7.41826} \pm \num{0.32365}$ & $\num{8.29048} \pm \num{0.53172}$ \\\precogrow
& DESIRE-plan & $\num{6.56219} \pm \num{0.20719}$ & $\num{2.26053} \pm \num{0.09984}$ & $\num{6.64410} \pm \num{0.31435}$ & $\num{6.18376} \pm \num{0.32513}$ & $\num{9.20305} \pm \num{0.44800}$ & 	$\num{8.51950} \pm \num{0.51390}$ \\\precogrow
& ESP & $\num{2.92126} \pm \num{0.17499}$ & $ \num{1.86066} \pm \num{0.10935}$ & $\num{2.36853} \pm \num{0.18780}$ & $ \num{2.81241} \pm \num{0.18794}$ & $\num{3.20137} \pm \num{0.25363}$ & $\num{4.36335} \pm \num{0.65235}$ \\ \precogrow
& PRECOG & $\boldmath \num{2.50763} \pm \num{0.15214}$ & $\boldmath \num{0.14913} \pm \num{0.02075}$ & $\boldmath \num{2.32361} \pm \num{0.18743}$ & $\boldmath \num{2.65441} \pm \num{0.19017}$ & $\boldmath \num{3.15719} \pm \num{0.27262}$ & $\boldmath \num{4.25379} \pm \num{0.58602}$ \\
 \bottomrule
\end{tabular}
}
\end{figure*}

 \newcommand{\plwidth}[0]{.24\textwidth}
\begin{figure*}[htp]
    \centering
    \begin{subfigure}{\plwidth}
    \includegraphics[width=\textwidth]{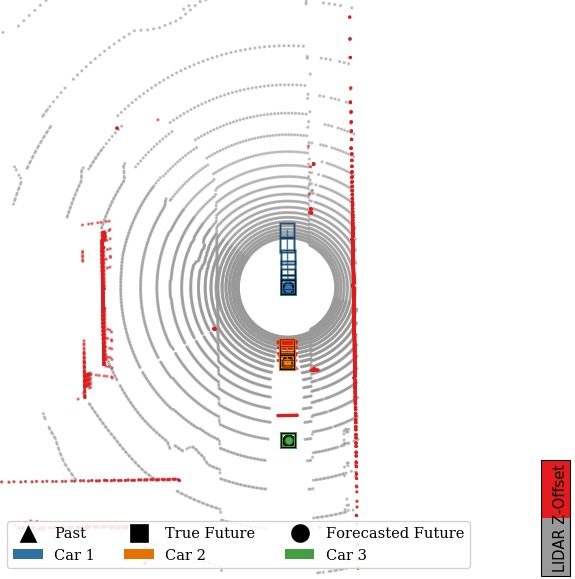}
    \caption{CARLA, ESP}
    \end{subfigure}
    \begin{subfigure}{\plwidth}
    \includegraphics[width=\textwidth]{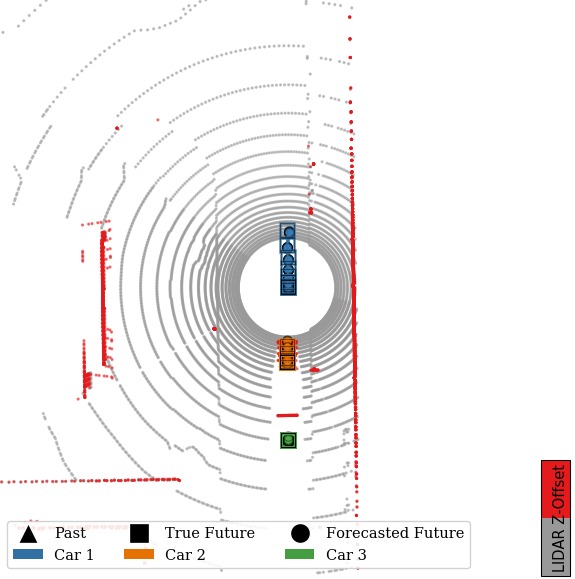}
     \caption{CARLA, PRECOG}
    \end{subfigure}
    \begin{subfigure}{\plwidth}
    \includegraphics[width=\textwidth]{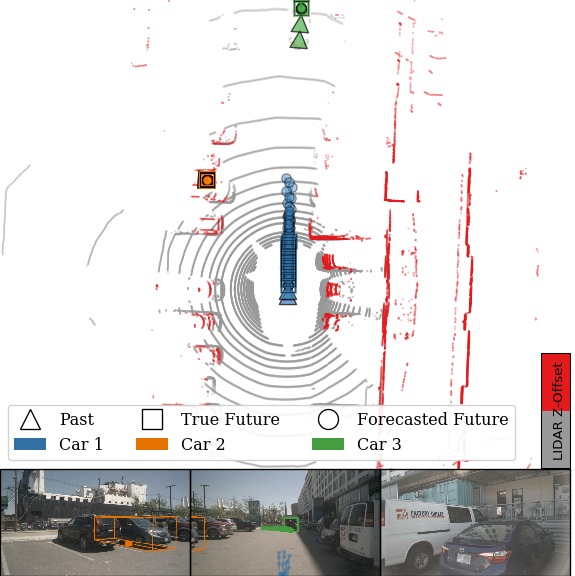}
      \caption{nuScenes, ESP}
    \end{subfigure}
    \begin{subfigure}{\plwidth}
    \includegraphics[width=\textwidth]{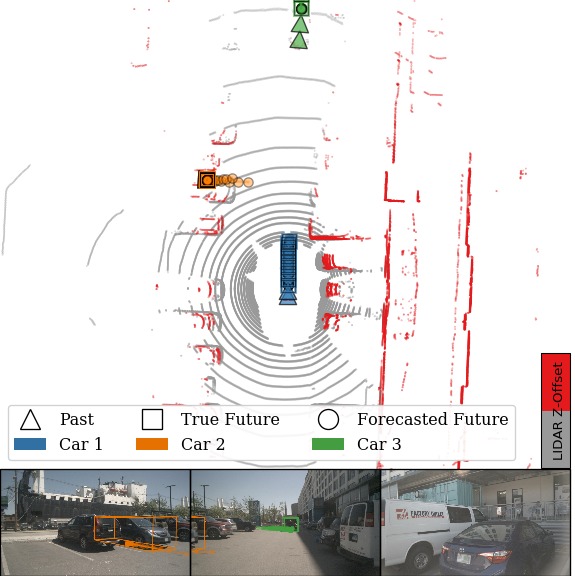}
         \caption{nuScenes, PRECOG}
    \end{subfigure}
    \caption{Examples of \emph{planned} multi-agent forecasting (PRECOG) with our learned model in CARLA and nuScenes. By using our planning approach and conditioning the robot on its true final position, our predictions of the other agents change, our predictions for the robot become more accurate, and sometimes our predictions of the other agent become more accurate.}
    \label{fig:example_carla_plan}
\end{figure*} 

\vspace{-2mm}
\subsubsection{CARLA and nuScenes PRECOG}
\vspace{-1mm}
\noindent{\bf DESIRE planning baseline}: We developed a straightforward planning baseline by feeding an input goal state and past encoding to a two-layer 200-unit ReLU MLP trained to predict the latent state of the robot given training tuples $(x\!=\!(\!\mathcal H_X,\State^r_T\sim\!q_{\text{DESIRE}}(\State | \phi, \latent^r)_T)), y=\latent^r)$. The latents for the other agents are samples from their DESIRE priors. 

\noindent{\bf Experiments:} We use the trained ESP models to run PRECOG on the test-sets in CARLA and nuScenes.
Here, we use both $\hat{m}_K$ and $\hat{m}_K^a$ to quantify joint sample quality in terms of all agents and each agent individually. 
In \tab{carla_planning_forecast} and \fig{example_carla_plan}, we report results of our planning experiments.
We observe that our planning approach significantly improves the quality of the joint trajectories. As expected, the forecasting performance improves the most for the planned agent ($\hat{m}_K^1$). Notably, the forecasting performance of the other agents improves across all datasets and all agents. We see the non-planned-agent improvements are usually greatest for Car 2 ($\hat{m}_K^2$). This result conforms to our intuitions: Car 2 is the \emph{closest} agent to the planned agent, and thus, it the agent that Car 1 influences the most. Qualitative examples of this planning are shown in \fig{example_carla_plan}. We observe trends similar to the CARLA planning experiments -- the forecasting performance improves the most for the planned agent, with the forecasting performance of the unplanned agent improving in response to the latent plans.  See \app{visualizations} for additional visualizations.

\section{Conclusions}
We present a multi-agent forecasting method, ESP, that outperforms state-of-the-art multi-agent forecasting methods on real (nuScenes) and simulated (CARLA) driving data. We also developed a novel algorithm, PRECOG, to condition forecasts on agent goals. We showed conditional forecasts improve joint-agent and per-agent predictions, compared to  unconditional forecasts used in prior work. Conditional forecasting can be used for planning, which we demonstrated with a novel multi-agent imitative planning objective.
Future directions include conditional forecasting \wrt multiple agent goals, useful for multi-AV coordination via communicated intent.

\vfill
\noindent{\bf Acknowledgements:} 
{
We thank K. Rakelly, A. Filos, A. Del Giorno, A. Dragan, and reviewers for their helpful feedback. Sponsored in part by IARPA (D17PC00340), ARL DCIST CRA W911NF-17-2-0181, DARPA via the Assured Autonomy Program, the ONR, and NVIDIA.
}
\clearpage

\bibliographystyle{ieee_fullname}
\bibliography{bibliography}

\appendix           

\clearpage 
\newcommand{\nuscappwidth}[0]{.325\textwidth}
\section{Planning and Forecasting Algorithms}
To execute planning, we perform gradient ascent to approximately solve the optimization problem \eqn{objectiveZ}. Recall the latent joint behavior is $\Latent \doteq \Latent^{1:A}_{1:T}$, the latent human behavior is $\Latent^h \doteq \Latent^{2:A}_{1:T}$, and the robot behavior is $\latent^r \doteq \latent^{1}_{1:T}$. We approximate the expectation in \eqn{objectiveL} with a sample expectation over $K$ samples from $p(\Latent^h)=\mathcal{N}(\Zero,\Eye)$, denoted ${}^{1:K}\!\latent^{h}$, where the $k^{\mathrm{th}}$ sample is ${}^{k}\!\latent^{h}$. Each of these samples for the latent human behavior is combined with the single latent robot plan, each denoted ${}^{k}\!\latent = [\latent^r, {}^{k}\!\latent^h]$. This batch is denoted ${}^{1:K}\!\latent$. The approximation of \eqn{objectiveL} is then given as
\begin{align} 
\!\hat L({}^{1:K}\!\latent,\mathcal{G},\!\phi)\!=\!\frac{1}{K}\sum_{k=1}^K\! &\log(q(f({}^{k}\!{\latent})| \phi) p(\mathcal G| f({}^{k}\!{\latent}),\phi)),  \label{eqn:approxL}
\end{align}
with $\hat L$ parameterized by $(q, f, p)$, and $f$'s dependence on $\phi$ dropped for notational brevity. The ${}^{1:K}\!\latent^{h}$ is redrawn before each gradient ascent step on \eqn{approxL}. This procedure is illustrated in Alg.~\ref{alg:latent-planning}.
\begin{algorithm}
    \caption{\, \sc{MultiImitativePlan}($q, f, p, \phi, K$)}
    \label{alg:latent-planning}
    \begin{algorithmic}[1]
        \STATE Define $\hat L$ with $q, f, p$
        \STATE Initialize $\latent_{1:T}^{r} \sim \mathcal N(0, I)$
        \WHILE{not converged} 
            \STATE ${}^{1:K}\!{\latent}^{h} \iidsim N(0, I)$
            \STATE $\latent_{1:T}^{r} \gets \latent_{1:T}^{r} + \nabla_{\latent_{1:T}^{r}} \hat L({}^{1:K}\!\latent,\mathcal{G},\phi)  $
        \ENDWHILE
        \STATE \textbf{return} $\latent_{1:T}^r$
    \end{algorithmic}
\end{algorithm}

In our implementation, we use $K\!=\!12$, track the $\latent^r_{1:T}$ that achieved the best $\hat L$ score, terminate the ascent if the best $\latent^r_{1:T}$ has not improved in $10$ steps, and return the corresponding best $\latent^r_{1:T}$. To initialize $\latent^r_{1:T}$ more robustly, we sample a full ${}^{1:K}\!\latent$ multiple times ($15$), and use the $\latent^r$ corresponding to the best $\hat L$. We can also run the planning over multiple initial samples of $\latent^r_{1:T}$. 

Now, we further detail how we can use this planning to perform goal-conditioned forecasting. As described in \sct{multiagentplanningexperiments}, we model goals in our experiments by defining our goal-likelihood $p(\mathcal G|\State_{1:T},\phi) = \mathcal N(\State^r_{T} ; \State^{*r}_T, 0.1\!\cdot\!\Eye)$, \emph{i.e.} a normal distribution at the controlled agent's last true future position, $\State^{*r}_T$. In general, we can pass any final desired robot position, $\state^{\dagger r}_T$ as the mean of this distribution. Then, we perform goal-conditioned forecasting on a specific scene $\phi$ to a specific robot goal $\state^{\dagger r}_T$, with our trained multi-agent density $q$, defined by $f$. This forecasting is performed by first planning $\latent^r$ according to Alg.~\ref{alg:latent-planning}, then sampling $\Latent^h$ again to generate stochastic joint outcomes, conditioned on the robot's plan. This procedure is illustrated in Algs.~\ref{alg:precog}~and~\ref{alg:stateprecog}.

\begin{algorithm}
    \caption{\, \sc{\algname}($q, f, p, \state^{\dagger r}_T, \phi, K$)}
    \label{alg:precog}
    \begin{algorithmic}[1]
        \STATE $\latent^r \gets$ \textsc{MultiImitativePlan}$(q, f, p, \phi, K)$
        \STATE Sample ${}^{1:K}\!\latent^h_{1:T} \iidsim N(0, I)$
        \STATE Forecast ${}^{1:K}\!\state_{1:T}^{1:A} \gets f({}^{1:K}\!\latent_{1:T}^{1:A}, \phi)$
        \STATE \textbf{return} ${}^{1:K}\!\state^{1:A}_{1:T}$
    \end{algorithmic}
\end{algorithm}

\begin{algorithm}
    \caption{\, \sc{Pos\algname}($q, f, \state^{\dagger r}_T, \phi, K$)}
    \label{alg:stateprecog}
    \begin{algorithmic}[1]
        \STATE Define $p(\mathcal G|\State_{1:T},\phi)=\mathcal N(\State^r_{T} ; \state^{\dagger r}_T, 0.1\!\cdot\!I)$
        \STATE \textbf{return} {\sc\algname}($q, f, p, \state^{\dagger r}_T, \phi, K$)
    \end{algorithmic}
\end{algorithm}

\section{Alternate Joint PDF forms}

The original joint can be expanded  over each agent:
\begin{equation}
    q(\State|\phi)\;=\;\prod_{t=1}^Tq(\State_t|\State_{1:t-1}\!,\phi) 
    \;=\;\prod_{t=1}^T\prod_{a=1}^A\mathcal{N}(\State_t^a; \bmu_t^a\!, \bSigma_t^a). \nonumber
\end{equation}
Additionally, the change-of-variables rule yields an equivalent density \cite{dinh2016realnvp,grathwohl2018ffjord,kingma2018glow,rhinehart2018r2p2}:

   \begin{align}\textstyle
   q(\State|\phi)  &= \mathcal N(f^{-1}(\State;\phi); 0, I) \left|\mathrm{det}\frac{\mathrm{d}f}{\mathrm{d}\Latent} |_{\Latent=f^{-1}(\State;\phi)} \right|^{-1}, \nonumber
\end{align}

We can derive expressions via the rollout equation \eqn{verlet}, reproduced here as \eqn{verlet_repr}, which implicitly defines $f$:

\begin{equation}
 \textstyle \State^a_{t}=\underbrace{2\State^a_{t\!-\!1}\!\!-\!\State^a_{t\!-\!2}\!\!+\!m^a_\theta(\State_{1:t\!-\!1}, \!\phi)}_{\bmu_{t}^a}+\underbrace{\sigma^a_\theta(\State_{1:t\!-\!1},\!\phi)}_{\bsigma_{t}^a}\cdot\Latent^a_{t}. \label{eqn:verlet_repr}
\end{equation}

The full Jacobian is given as:

$$\textstyle \frac{\mathrm{d} f}{\mathrm{d}\Latent} = \begin{bmatrix}
    \frac{\partial \State_1}{\partial\Latent_1} & 0 & \dots & 0 \\
    \frac{\partial \State_2}{\partial\Latent_1} & \frac{\partial \State_2}{\partial\Latent_2} & \dots & 0 \\
    \vdots & \vdots & \ddots & 0 \\
    \frac{\partial \State_T}{\partial  \Latent_1}  & \frac{\partial \State_T}{\partial  \Latent_2}  & \dots & \frac{\partial \State_T}{\partial\Latent_T} 
\end{bmatrix},$$

where 
\begin{align}\textstyle \frac{\partial \State_t}{\partial\Latent_t}\!=\!\begin{bmatrix}
    \bsigma_t^1 & 0 & \dots & 0\\
    \frac{\partial \State_t^2}{\partial \Latent_t^1} & \bsigma_t^2 & \dots & 0 \\
    \vdots & \vdots & \ddots & 0 \\
    \frac{\partial \State_t^A}{\partial \Latent_t^1}  & \frac{\partial \State_t^A}{\partial  \Latent_t^2}  & \dots & \bsigma_t^A
\end{bmatrix}\!=\!\begin{bmatrix}
    \bsigma_t^1 & 0 & \dots & 0\\
    0 & \bsigma_t^2 & \dots & 0 \\
    \vdots & \vdots & \ddots & 0 \\
    0  & 0  & \dots & \bsigma_t^A
\end{bmatrix}\nonumber
\end{align}

Due to the block triangular nature of the Jacobian and applying Laplace expansion along the diagonal:
\begin{align}
 \textstyle   \mathrm{det}\frac{\mathrm{d}f}{\mathrm{d}\Latent}=&\prod_t \mathrm{det}\;\frac{\partial \State_t}{\partial\Latent_t}=\prod_{t=1}^T\prod_{a=1}^A \mathrm{det}\;\sigma_t^a(\State_{1:t-1},\phi).\nonumber
\end{align}

$\Latent\!=\!f^{-1}(\State;\phi)$ is given by computing each $\Latent_t^a=\left(\sigma^a_{t}(\State_{1:t-1}, \phi)\right)^{-1}\left(\State_t^a-\mu_t^a(\State_{1:t-1}, \phi)\right).$ Algorithmically,  the functions $f$ and $f^{-1}$ are  implemented  separately,  each  with  a  \emph{double  for-loop  over  agents  and  time}. Note that since we use RNNs to produce $\bmu_t$ and $\bsigma_t$, the forward $f$ and its inverse must be computed in the same direction by stepping the RNN's forward in time over the input $\State$. To aid implementation, we use  the  following  checks  to  ensure $f$ is  a  bijection:	$||\Latent-f^{-1}(f(\Latent, \phi), \phi)||_\infty < \epsilon$, $||\State-f(f^{-1}(\State,\phi),\phi)||_\infty < \epsilon$. 

\section{Architecture and Training Details}\label{app:implementation}

Both past and future trajectories for each agent are represented in each agent's own \emph{local} coordinate frame at $t=0$, with agent's forward axis pointing along the agent's yaw at $t=0$. Each agent $a$ observes positions of the other agents in the coordinate frame of agent $a$. We use a 9-layer fully-convolutional network with stride $1$ and $32$ channels per layer, and kernel sizes of $3 \times 3$, to process $\chi$ into a feature grid $\bGamma$ at the same spatial resolution as $\chi$. The LIDAR is mounted on the first agent, thus it is generally more informative about nearby agents.  This enables the prediction to be learned \emph{relative} to the agent, with global context obtained by feature map interpolation. At each time step, each agent's predicted future position $\state_t^a$ is bilinearly-interpolated into $\bGamma$: $\bGamma(\state_t^{a})$, which ensures $\nicefrac{\mathrm{d}\bGamma(\state_t^{a})}{\mathrm{d}\state_t^{a}}$ exists.  The ``SocialMapFeat" component performs this interpolation by converting the positions (in $\mathrm{meters}$) to feature grid coordinates (in $0.5~\nicefrac{\mathrm{meters}}{\mathrm{cell}}$), and bilinearly-interpolating each into the feature map $\bGamma$. The interpolation is performed by retrieving the features at the corners of the nearest unit square to the current continuous position. 

We also employed an additional featurization scheme, termed ``whiskers''. Instead of interpolating only at $\state_t^a$, we interpolated at nearby positions subsampled from arcs relative to $\state_t^a$ at various radii. By letting $\state_t^a - \state_{t-1}^a$ define the predicted orientation, the arcs were generated by evenly sampling $7$ points along arcs of length $\nicefrac{5\pi}{4}$ at radii $[1, 2, 4, 8, 16, 32]$ meters, which loosely simulates the forecasted agent's future field-of-view. The midpoint of each arc lied along the ray from $\state_{t-1}^a$ through $\state_t^a$. After interpolating at points $\{\omega_n\}_{n=1}^{42}$, the resulting feature is of size $8\cdot7\cdot6$ ($8$ is the size of the last dimension of $\Gamma$, $7$ is the number of points per arc, and $6$ is the number of arcs). We found this approach to yield superior performance and employed it in the R2P2-MA baseline, as well as all of our methods. The full details of the architecture are provided in \tab{archdetailed}.

Finally, we performed additional featurization in the nuScenes setting by replacing $\chi$ with a \emph{signed-distance transform}, similar to \cite{rhinehart2018r2p2}. It provides a spatially-smoother input to the convolutional network, which we found augmented performance. The signed distance transform ($\mathrm{SDT}$) of $\chi_c \in \mathbb R^{H \times W}$ can be computed by first binarizing to $\chi_c \in \{0, 1\}^{H \times W}$ and using the Euclidean distance transform ($\mathrm{DT}$), which is commonly provided (\emph{e.g.} in \texttt{scipy}). We compute it by binarizing with threshold $\tau$: $\mathrm{SDT}(\chi_c, \tau) = \mathrm{DT}(\chi_c \geq \tau) - \mathrm{DT}(\chi_c < \tau)$, then clipping the result to $[-10, 1]$, and finally normalizing to $[0, 1]$. For LIDAR channels, we use $\tau=5$. When we use the already-binarized road prior, binarization is unnecessary.

\begin{figure*}[th!]
    \centering
    \begin{subfigure}[t]{0.19\textwidth}
        \vskip 0pt
        \centering
        \includegraphics[width=0.9\textwidth]{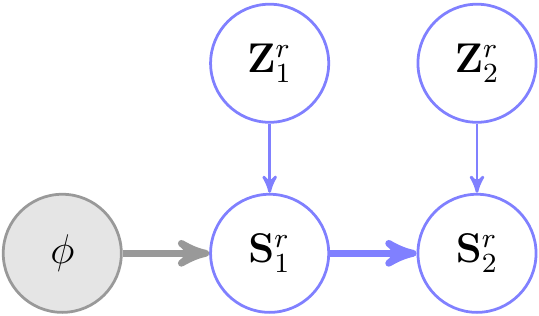}
        \vspace{21mm}
        \caption{R2P2 forecast \cite{rhinehart2018r2p2}}
        \label{fig:r2p2_forecast}
    \end{subfigure}
    \begin{subfigure}[t]{0.19\textwidth}
        \vskip 0pt
        \centering
        \includegraphics[width=0.9\textwidth]{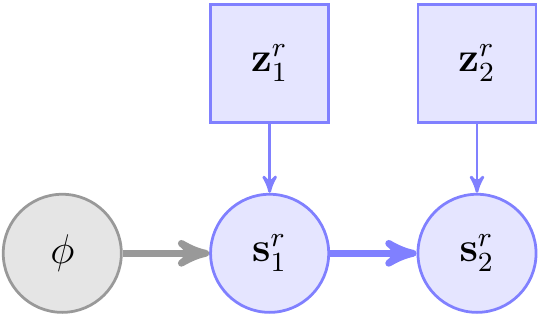}
        \vspace{21mm}
        \caption{DIM planning \cite{rhinehart2018deep}}
        \label{fig:dim_plan}
    \end{subfigure}
    \begin{subfigure}[t]{0.19\textwidth}
        \vskip 0pt
        \centering
        \includegraphics[width=0.9\textwidth]{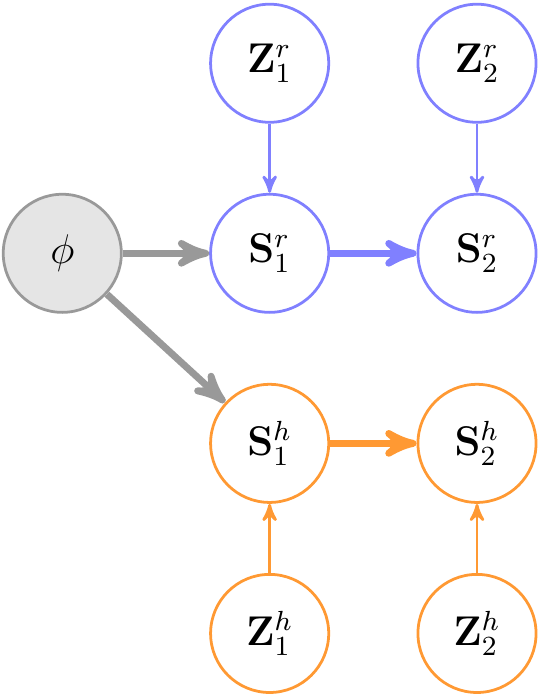}
        \caption{R2P2-MA forecast}
        \label{fig:r2p2star_forecast}
    \end{subfigure}
    \begin{subfigure}[t]{0.19\textwidth}
         \vskip 0pt
        \centering
        \includegraphics[width=0.9\textwidth]{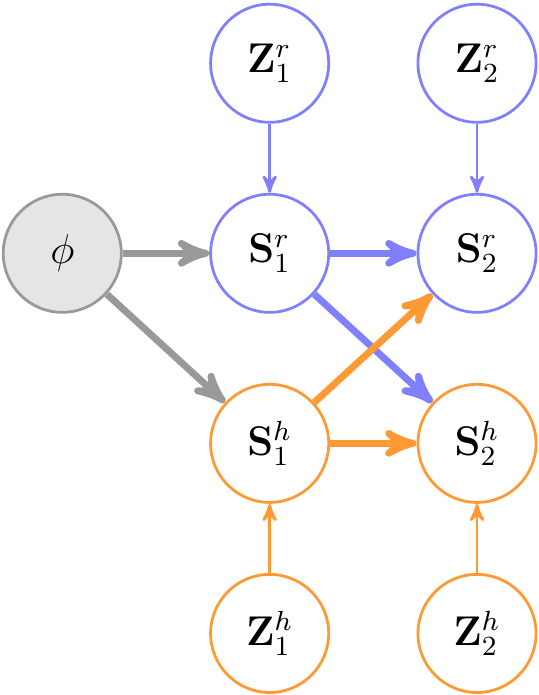}
        \caption{ESP forecast}
        \label{fig:esp_forecast}
    \end{subfigure}
    \begin{subfigure}[t]{0.19\textwidth}
        \vskip 0pt
        \centering
        \includegraphics[width=0.9\textwidth]{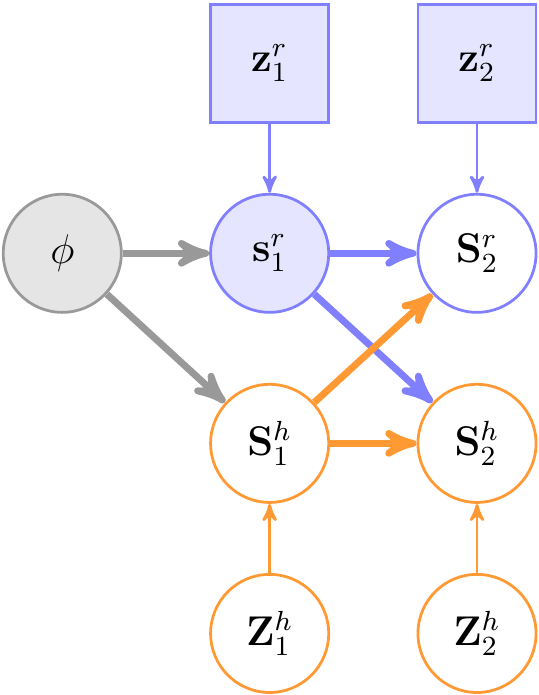}
        \caption{PRECOG planning}
        \label{fig:precog_plan}
    \end{subfigure}
    \caption{
    Graphical model comparison between prior work (\fig{r2p2_forecast}, \fig{dim_plan}); a baseline we used    (\fig{r2p2star_forecast}); and our proposed methods (\fig{esp_forecast}, \fig{precog_plan}). All figures show $A=2$ and two steps of the true $T$-step horizon. Shaded nodes represent observed variables, and square nodes represent robot decisions. Thick arrows represent non-Makovian ``carry-forward'' dependencies (i.e.\ a state can depend on multiple previous states): add a thin arrow for every two nodes connected by a chain of thick arrows. Future reactions are always unknown in the case of the human drivers (``h'' superscript), but can be decided in the case of robot (``r'' superscipt) planning. How vehicles react affects---and induces uncertianty into---the multi-agent system state $\State$. 
    }
    \label{fig:forecast-and-planning-graph-full} 
\end{figure*}

\begin{figure*}[th!]
    \centering
    \begin{subfigure}[t]{0.4\textwidth}
         \vskip 0pt
        \centering
        \includegraphics[width=\textwidth]{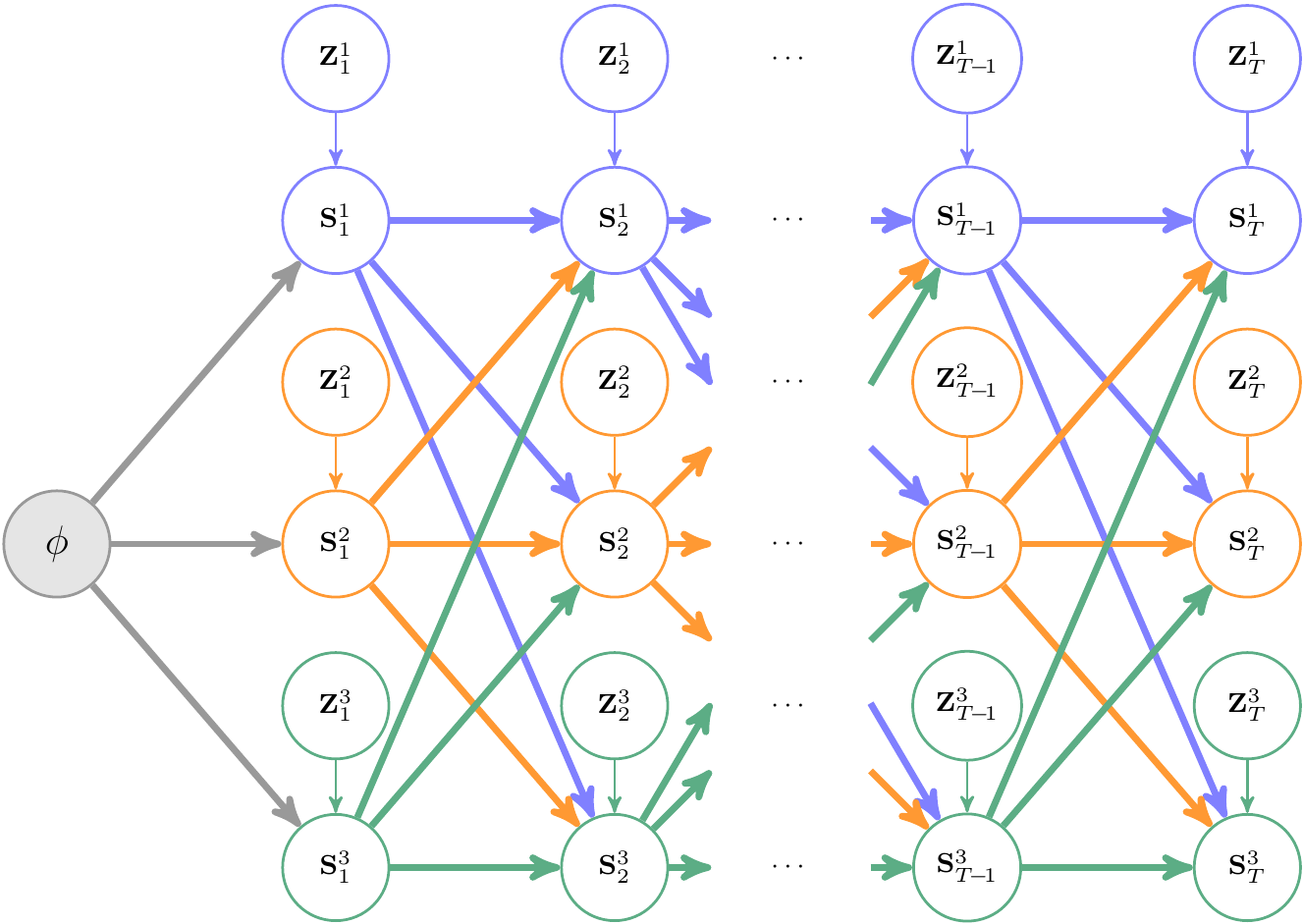}
        \caption{ESP forecast}
        \label{fig:esp_forecast_A3}
    \end{subfigure}
    \hfill
    \begin{subfigure}[t]{0.4\textwidth}
        \vskip 0pt
        \centering
        \includegraphics[width=\textwidth]{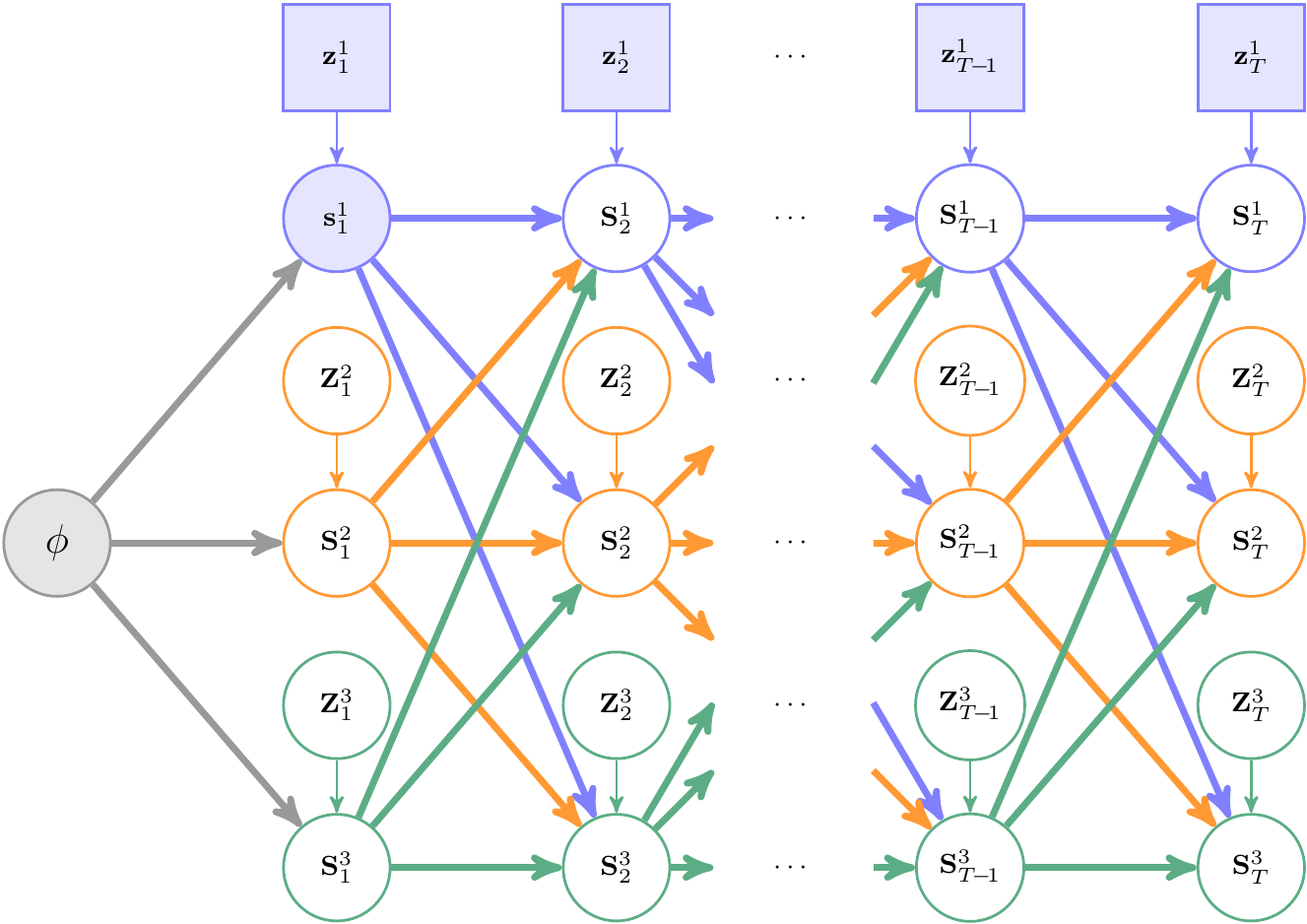}
        \caption{PRECOG planning}
        \label{fig:precog_plan_A3}
    \end{subfigure}
    \caption{
    Graphical models of ESP and PRECOG for $A=3$. See \fig{forecast-and-planning-graph-full}'s caption for notation.
    }
    \label{fig:forecast-and-planning-graph-A3} 
\end{figure*}

In \fig{forecast-and-planning-graph-full}, we illustrate various forms of the probabilistic graphical models corresponding to our main model (ESP), a baseline model (R2P2-MA), and how the assignment of latent variables ($\Latent$) in these models affects the production of the states ($\State$). In \fig{forecast-and-planning-graph-A3}, we illustrate the graphical models of ESP and PRECOG for $A=3$.

We trained our model with stochastic gradient descent using the Adam optimizer with learning rate $1\cdot10^{-4}$, with minibatch size $10$, until validation-set performance (of $\hat e$, as discussed in the main paper) showed no improvement for a period of $10$ epochs.

\begin{table*}[th]
\centering
\caption{Detailed ESP Architecture that implements $\state_{1:T}^{1:A}=f(\latent_{1:T}^{1:A}, \phi)$. Typically, $T=20$, $A=5$, $D=2$. In CARLA, $H = W = 100$. In nuScenes, $H = W = 200$. An asterisk $~({}^*)$ denotes a component whose output is masked in the flexible-count version of ESP by using the agent-presence mask $M\!\in\!\{0,1\}^{A_{\text{train}}}$, discussed in Sec.~\ref{sct:implementation}.}
\label{tab:archdetailed} 
\resizebox{.98\textwidth}{!}{
\begin{tabular}{lllllllll}
\toprule
Component & Input [dimensionality] & Layer or Operation  & Output [dimensionality] & Details \\
\midrule
\multicolumn{4}{l}{\emph{Static featurization of context:} $\phi=\{\chi,\state_{-\tau:0}^{1:A}\}$. \emph{Shared parameters for each agent.}} &\\
\midrule
MapFeat  & $\chi~[H,W,2]$ & 2D Convolution & ${}^{1}\chi~[H,W, 32]$ & $3\times3$ stride 1, ReLu \\
MapFeat  & ${}^{i-1}\chi~[H,W,32]$ & 2D Convolution & ${}^{i}\chi~[H, W, 32]$ & $3\times3$ stride 1, ReLu, $i \in [2, \dots, 8]$ \\
MapFeat  & ${}^{8}\chi~[H,W,32]$ & 2D Convolution & $\bGamma~[H, W, 8]$ & $3\times3$ stride 1, ReLu\\
PastRNN$^*$ & $\state_{-\tau:0}^{1:A}~[\tau+1, A, D]$ & RNN & ${}^1\alpha^{1:A}~[A, 128]$ & GRU across time dimension \\
PastRNN & $\alpha~[A, 128]$ & ${}^1\alpha^a \oplus \sum_{b \in \{1..A\} \setminus a}{}^1\alpha^b$& ${}^2\alpha^a~[256]$ & Index, Concat, \& Sum for agent-$a$ context \\
\midrule
\multicolumn{4}{l}{\emph{Dynamic generation via double loop:} $\mathrm{for}~t\in\{0,\dots,T-1\},\mathrm{for}~a\in\{1,\dots,A\}$. \emph{Shared or separate parameters for each agent.}}\\
\midrule
SocialFeat$^*$ & $\state_{t}^{1:A}~[AD]$ & $\state_{t}^a-\state_{t}^b,b\in\{1..A\}\setminus a$ & ${}^{0}\eta_t^a~[AD-D]$ & Agent displacements\\
SocialFeatMLP & ${}^{0}\eta_t^a~[AD-D]$ & Affine (FC) & ${}^{1}\eta_t^a~[200]$ & Tanh activation\\
SocialFeatMLP & ${}^{1}\eta_t^a~[200]$ & Affine (FC) & ${}^{2}\eta_t^a~[50]$ & Identity activation\\
WhiskerMapFeat & $\omega_1\dots\omega_N~[42,D]$ &  Interpolate & $w_t^a=\bGamma(\omega_1)\oplus\dots\oplus\bGamma(\omega_N)~[8\cdot42]$ & Interpolate ahead of the sample's P.O.V. \\
SocialMapFeat$^*$ & $\state_t^{1:A}~[AD]$ & Interpolate & $\gamma_t^a=\bGamma(\state_t^1)\oplus\dots\oplus\bGamma(\state_t^A)~[8A]$ & Differentiable interpolation, concat. ($\oplus$)\\
JointFeat & $\gamma_t^a,\state_0^{1:A},{}^{2}\eta_a,{}^2\alpha^a,w_t^a$ & $\gamma_t^a\oplus \state_0^{1:A}\oplus{}^{2}\eta_a\oplus{}^2\alpha^a\oplus{}w_t^a$  & $\rho^a_t~[8A+AD+50+256+336]$ &  Concatenate ($\oplus$) \\
FutureRNN  & $\rho^a_t~[8A+AD+50+256]$ & RNN & ${}^{1}\rho^a_t~[50]$  & GRU\\
FutureMLP  & ${}^{1}\rho^a_t~[50]$ & Affine (FC) & ${}^{2}\rho^a_t~[200]$  & Tanh activation\\
FutureMLP  & ${}^{2}\rho^a_t[200]$ & Affine (FC) & $m_t^a~[D],\;\xi_t^a~[D,D]$  & Identity activation\\
MatrixExp & $\xi_t^a~[D,D]$& $\mathrm{expm}(\xi_t^a + \xi_t^{a,\mathrm{transpose}})$ & $\bsigma_t^a~[D,D]$ & Differentiable Matrix Exponential \cite{rhinehart2018r2p2} \\
VerletStep & $\state_{t},\state_{t-1},m_t^a,\bsigma_t^a,\latent_t^a$ & $2\state_{t}-\state_{t-1}+m_t^a+\bsigma_t^a\latent_t^a$ & $\state_{t+1}^a~[D]$ & (Eq.~\ref{eqn:verlet})\\
\bottomrule
\end{tabular}
}
\end{table*}

 \begin{figure}[t]
     \centering
     \begin{subfigure}[t]{.45\columnwidth}
     \includegraphics[width=\textwidth]{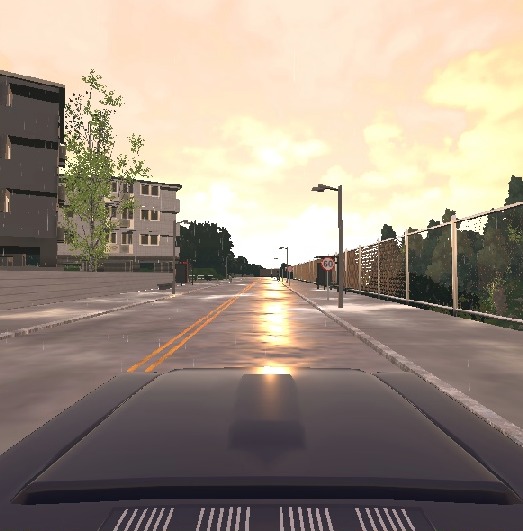}
    \end{subfigure}
         \begin{subfigure}[t]{.45\columnwidth}
     \includegraphics[width=\textwidth,angle=90]{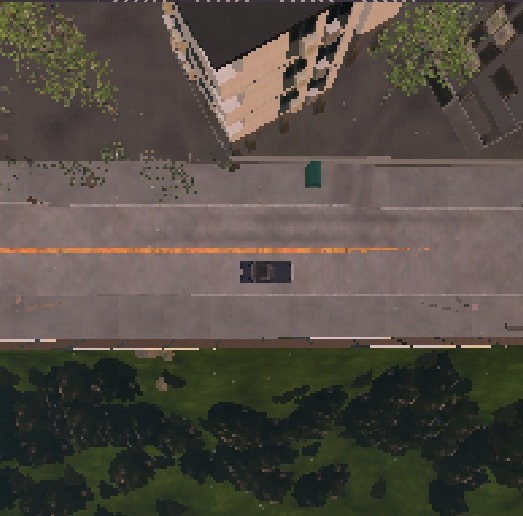}
    \end{subfigure}
    \caption{Images from the CARLA simulator \cite{dosovitskiy17carla}. \emph{Left:} frontal view. \emph{Right:} overhead view.} \label{fig:carlaims}
 \end{figure}

\section{Baseline Implementations}
\noindent \textbf{SocialGAN}  We used the public implementation {\small \url{https://github.com/agrimgupta92/sgan}}. We found the default {\small \tt train.py} parameters yielded poor performance. We achieved significantly better SocialGAN performance by using the network parameters in the {\small \tt run\_traj.sh} script. In contrast to SocialGAN, we model joint trajectories, and can compute likelihoods for planning (and for $\hat e$). 

\noindent \textbf{DESIRE}  We re-implemented DESIRE following the authors' description in the paper and supplement, as there is no open-source version available. In our domain, $\chi$ is purely LIDAR-based, whereas their model combines image-based semantic segmentation features into the same coordinate frame. We found most provided parameters to work well, except those related to the re-ranking component. The re-ranking often did not improve the trajectories. The best results were obtained with 1 re-ranking step. Whereas DESIRE is trained with a single-agent evidence lower bound (ELBO), our model jointly models multiple agents with an exact likelihood.  As DESIRE does not compute multi-agent likelihoods, we cannot compute its $\hat e$, nor use it for planning in a multi-agent setting.

\noindent \textbf{R2P2}  We re-implemented R2P2 following the authors' description in the paper and supplement, as there is no open-source version available. We extended R2P2 to the multi-agent setting and use it as our R2P2-MA model; R2P2 does not jointly model agents. We can compute R2P2's likelihood, and therefore $\hat e$, by assuming independence across agents: $q(\State|\phi)=\prod_{a=1}^Aq^a(\State^a|\phi)$. Note that since this joint likelihood does not model agent's future actions to influence each other, R2P2 cannot be used for planning in a multi-agent setting. \fig{forecast-and-planning-graph-full} compares the R2P2 baseline to our ESP model.

\begin{table*}[thb]
\centering
\caption{CARLA multi-agent forecasting evaluation. All CARLA-trained models use {\tt Town01 Train} only, and are tested on {\tt Town01 Test}. Mean scores (and their standard errors) of sample quality $\hat{m}$ \eqn{mhat}, and log likelihood $\hat{e}$ \eqn{ehat}, are shown. The en-dash (--) indicates if an approach cannot compute likelihoods. The R2P2-MA generalizes the single-agent forecasting approach of \cite{rhinehart2018r2p2}. Variants of our ESP method (highlighted gray) mostly outperform prior work in the multi-agent CARLA setting. For single agent evaluations, see \tab{carla_forecast_A1}.} 
\label{tab:carla_forecast_town1} 
\resizebox{1.0\textwidth}{!}{
\begin{tabular}{lccccccccccc}
\toprule
Approach  & Test $\hat{m}_{K\!=\!12}$ & Test $\hat{e}$    &  & Test $\hat{m}_{K\!=\!12}$ & Test $\hat{e}$     &  & Test $\hat{m}_{K\!=\!12}$ & Test $\hat{e}$ &  & Test $\hat{m}_{K\!=\!12}$ & Test $\hat{e}$   \\
                   & (minMSD)                  & (extra nats)      &  & (minMSD)                  & (extra nats)       &  & (minMSD)                  & (extra nats)  &  & (minMSD)                  & (extra nats)   \\
\midrule

\textbf{CARLA Town01 Test}                   & \multicolumn{2}{c}{2 agents} &  & \multicolumn{2}{c}{3 agents} & & \multicolumn{2}{c}{4 agents} &  & \multicolumn{2}{c}{5 agents} \\
                     \cmidrule{2-3}                                     \cmidrule{5-6}                                      \cmidrule{8-9}    \cmidrule{11-12}
DESIRE \hfilll \cite{lee2017desire} &  $\num{1.942505} \pm \num{0.032771}$ & --  & & $\num{1.586585} \pm \num{0.020020}$ & -- & & $\num{2.234055} \pm \num{0.023480}$ & -- & & $\num{2.421706} \pm \num{0.017378}$ & -- \\
SocialGAN \hfilll \cite{gupta2018social} & $0.977 \pm 0.016$ & --  & & $0.812 \pm 0.013$ & -- & & $1.098 \pm 0.014 $ & -- & & $1.141 \pm 0.015$ & --\\
R2P2-MA \hfilll \cite{rhinehart2018r2p2} & $\num{0.539781} \pm \num{0.008945}$ & $\num{0.624733} \pm \num{0.001983}$ & & $\num{0.387280} \pm \num{0.008461}$ & $\num{0.644931} \pm \num{0.002254}$ & & $\num{0.689926} \pm \num{0.008749}$ &  $\num{0.621101} \pm \num{0.001736}$ & & $\num{0.770273} \pm \num{0.008075}$ & $\num{0.617563} \pm \num{0.001543}$\\
\rowcolor{ourmethod}
Ours: ESP, no LIDAR                            & $\num{0.724062} \pm \num{0.012838}$ &  $\num{0.687721} \pm \num{0.002532}$ & & $\num{0.718911} \pm \num{0.011194}$ & $\num{0.639706} \pm \num{0.002110}$ & & $\num{0.919307} \pm \num{0.011470}$ & $\num{0.650486} \pm \num{0.001871}$ & & $\num{1.101755} \pm \num{0.011160}$ & $\num{0.652004} \pm \num{0.001689}$ \\
\rowcolor{ourmethod}
Ours: ESP                                   & $\boldmath \num{0.311238} \pm \num{0.007892}$ & $\num{0.614608} \pm \num{0.002236}$ & & $\boldmath \num{0.385151} \pm \num{0.006941}$ & $\num{0.584910} \pm \num{0.001915}$ & & $\num{0.509251} \pm \num{0.007067}$ & $ \num{0.598852} \pm \num{0.001710}$& & $\num{0.675212} \pm \num{0.007193}$ & $\num{0.630041} \pm \num{0.001452}$  \\
\rowcolor{ourmethod}
Ours: ESP, flex. count & $\num{0.414851} \pm \num{0.013651}$ & $\boldmath \num{0.531100} \pm \num{0.001521}$ && $\num{0.397603} \pm \num{0.011015}$ & $\boldmath \num{0.512697} \pm \num{0.001322}$ && $\boldmath \num{0.410654} \pm \num{0.009709}$ & $\boldmath \num{0.507109} \pm \num{0.001181}$ && $\boldmath \num{0.446775} \pm \num{0.009190}$ & $\boldmath \num{0.508599} \pm \num{0.001097}$ \\
\bottomrule
\end{tabular}
}
\end{table*} 

\section{CARLA Dataset Details} \label{app:carla_dataset}
To remind the reader, we generated a realistic dataset for multi-agent trajectory forecasting and planning with the CARLA simulator \cite{dosovitskiy17carla}. Images from the simulator are shown \fig{carlaims}. We ran the autopilot in {\tt Town01} for over 900 episodes of 100 seconds each in the presence of 100 other vehicles, and recorded the trajectory of every vehicle and the autopilot's LIDAR observation. We randomized episodes to either train, validation, or test sets. We created sets of $\num{60701}$ train, $\num{7586}$ validation, and $\num{7567}$ test scenes, each with 2 seconds of past and 4 seconds of future position information at 5Hz.  The dataset also includes 100 episodes obtained by following the same procedure in {\tt Town02}. We used this data to further evaluate our ESP model. We applied our saved models (the same models used to report results in the paper) to this data.  We generated the CARLA data using version $0.8.4$. We used the default vehicle. We used a LIDAR position of $(0.0, 0.0, 2.5)$, with $32$ channels, a range of $50$, $\num{100000}$ points per second, a rotation frequency of $10$, an upper FOV limit of $10$, and a lower FOV limit of $-30$. We will release the 100GB of collected data.

\begin{table}[htb]
\centering
\caption{Performance in CARLA $A=1$ (n.b. here the model is identical to R2PA-MA (denoted by ${}^*$)). }
\label{tab:carla_forecast_A1} 
\resizebox{.95\linewidth}{!}{
\begin{tabular}{lcc}
\toprule
Approach  & Test $\hat{m}_{K\!=\!12}$ & Test $\hat{e}$  \\
& (minMSD)                  & (extra nats)   \\
\midrule
\textbf{CARLA Town01 Test} & \multicolumn{2}{c}{1 agent} \\
\cmidrule{2-3}                                    
DESIRE \hfilll \cite{lee2017desire}                      & $\num{1.0665} \pm \num{0.03996}$             & --                                         \\                             
SocialGAN \hfilll \cite{gupta2018social}                 & $0.921 \pm 0.031$                            & --                                         \\
R2P2-MA \hfilll \cite{rhinehart2018r2p2}  & ${}^*$ & ${}^*$  \\
\rowcolor{ourmethod}
Ours: ESP, no LIDAR                               & $\num{0.49561} \pm \num{0.023711}$           & $\num{0.69926} \pm \num{0.006148}$         \\
\rowcolor{ourmethod}
Ours: ESP                                      & $\boldmath \num{0.13589} \pm \num{.010101}$  & $\boldmath \num{0.63369} \pm \num{.0056}$        \\
\bottomrule
\end{tabular}
}
\end{table}

\section{Additional Evaluation}\label{app:additional_evaluation}

\subsection{Robustness to Agent Localization Errors}
\begin{figure}[htb]
\centering
 \begin{subfigure}{\columnwidth}
 \centering
    \includegraphics[width=\textwidth]{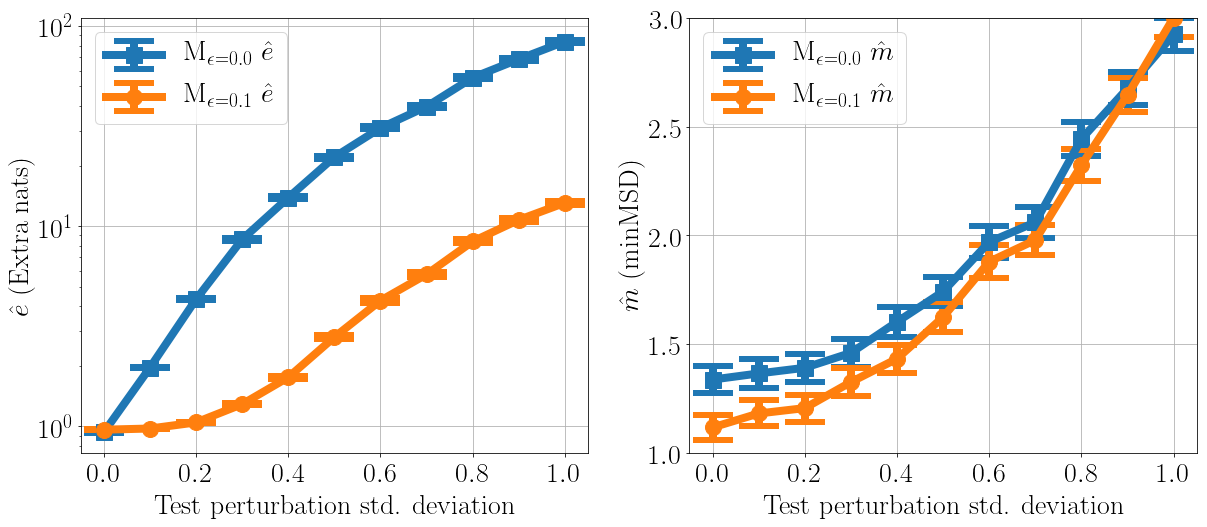}
    \end{subfigure}
\caption{Evaluating the effects of noisy localization on nuScenes $A=2$.} \label{fig:robust_models}
\end{figure}

In real-world data, there may be error in the localization of the other agents ($\state_{-\tau:0}$). We can simulate this error in our test-set by perturbing $\state^a_{-\tau:0}$ with a random vector $v^a\!\sim\!\mathcal N(\Zero,\!\epsilon I_{D\times D})$. We also train a model by injecting noise generated similarly. In \fig{robust_models} we compare nuScenes $A=2$ ESP models trained without ($M_{\epsilon=0.0}$) and with ($M_{\epsilon=0.1}$) noise injection. We observe that $M_{\epsilon=0.0}$ is much more sensitive to test-time noise than $M_{\epsilon=0.1}$ at all perturbation scales, which shows noise injection is an effective strategy to mitigate the effects of localization error. We also note $M_{\epsilon=0.1}$ improves performance even when the test-data is not perturbed.

\begin{figure}[htb]
\centering
 \begin{subfigure}{\columnwidth}
\includegraphics[width=\textwidth]{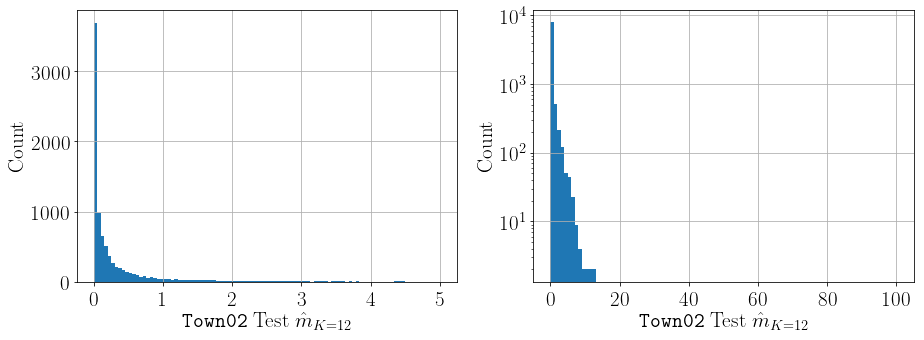}
\end{subfigure}
\caption{Histogram of $\hat m_{K=12}$ of forecasts made by the ESP flexible-count model on CARLA {\tt Town02} Test $A=5$, $T=20$ at $10$Hz (2 seconds of future). The median $\hat m_{K=12}$ is $0.09$.}  \label{fig:town02testhist}
\end{figure}

\begin{figure}[htb]
\centering
 \begin{subfigure}{\columnwidth}
\includegraphics[width=\textwidth]{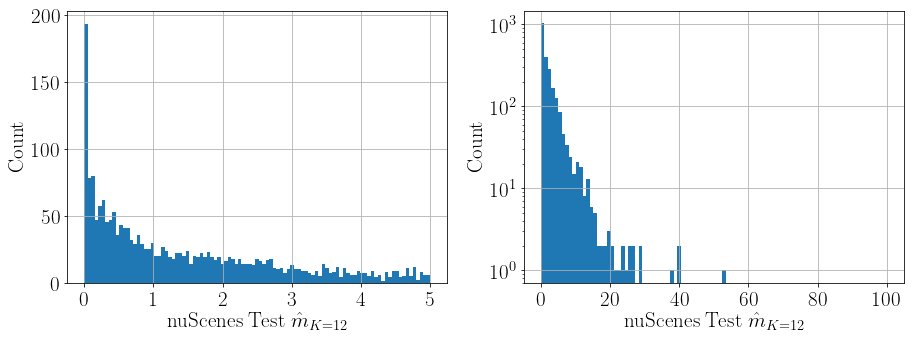}
\end{subfigure}
\caption{Histogram of $\hat m_{K=12}$ of forecasts made by the ESP flexible-count model on nuScenes Test $A=5$, $T=20$ at $5$Hz (4 seconds of future). The median $\hat m_{K=12}$ is $1.31$.}  \label{fig:nuscenestesthist}
\end{figure}

\begin{figure}[htb]
\centering
 \begin{subfigure}{.45\columnwidth}
\includegraphics[width=\textwidth]{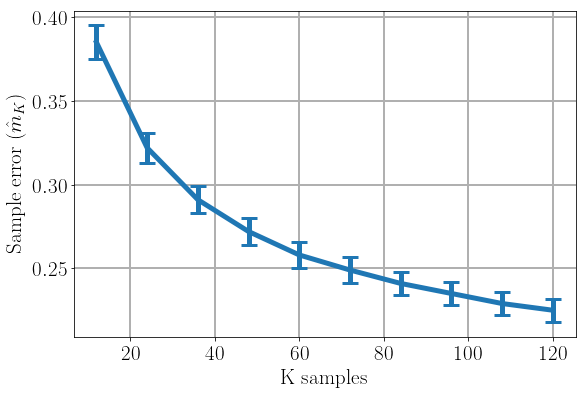}
\caption{Plot of $\hat m_K$ vs. $K$ of the ESP flexible-count model on CARLA {\tt Town02} Test $A=5$, $T=20$ at $10$Hz (2s).} 
\end{subfigure}
\hfill 
 \begin{subfigure}{.45\columnwidth}
\includegraphics[width=\textwidth]{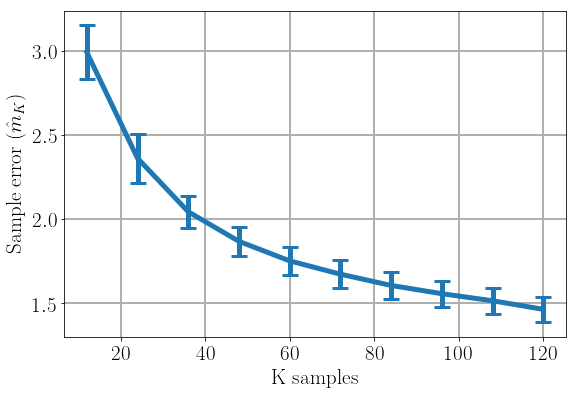}
\caption{Plot of $\hat m_K$ vs. $K$ of the ESP flexible-count model on nuScenes Test $A=5$, $T=20$ at $5$Hz (4s).} 
\end{subfigure} 
\caption{Mean $\hat m_K$ and its standard error vs.\ $K$  in two settings.} \label{fig:testmvsK}
\end{figure}

\begin{figure*}[th]
\centering
\captionsetup{type=table} 
\caption{Forecasting evaluation of our model on CARLA {\tt Town01 Test} and nuScenes Test data. Planning the robot to a goal position (PRECOG) generates better predictions for all agents. Means and their standard errors are reported. The en-dash (--) represents statistics of agents that are not present in a dataset.}
\label{tab:planning_forecast_app} 
\resizebox{1.0\textwidth}{!}{
\begin{tabular}{clcccccc}
\toprule
Data & Approach & \multirow{1}{*}{Test $\hat{m}_{K=12}$} & \multirow{1}{*}{Test $\hat{m}_{K=12}^{a=1}$} & \multirow{1}{*}{Test $\hat{m}_{K=12}^{a=2}$} & \multirow{1}{*}{Test $\hat{m}_{K=12}^{a=3}$} & \multirow{1}{*}{Test $\hat{m}_{K=12}^{a=4}$} & \multirow{1}{*}{Test $\hat{m}_{K=12}^{a=5}$} \\ 
\midrule
\multirow{4}{*}{CARLA\, $A\!=\!2$} & DESIRE & $\num{1.83703} \pm \num{0.04810}$ &	$\num{1.99149} \pm \num{0.06605}$ & $\num{1.68257} \pm \num{0.05013}$ & -- & -- & -- \\\precogrow
 & DESIRE-plan & $\num{1.85810} \pm \num{0.04614}$ & $\num{0.91834} \pm \num{0.04441}$ & $\num{2.79785} \pm \num{0.07331}$ & -- & -- & -- \\\precogrow
& ESP & $\num{0.33705} \pm \num{0.0132}$ & $\num{	0.196} \pm \num{0.009}$ &$\num{0.478} \pm \num{0.024}$ & -- & -- & --\\ \precogrow
& PRECOG & ${\boldmath \num{0.2406397} \pm \num{0.01207}}$	& ${\boldmath \num{0.055} \pm \num{0.003}}$ & ${\boldmath \num{0.426} \pm \num{0.024 }}$ & -- & -- & -- \\ \precogrow
\midrule
\multirow{4}{*}{CARLA\, $A\!=\!3$} & DESIRE & $\num{1.69854} \pm \num{0.03154}$ & $\num{1.56990} \pm \num{0.03743}$ & $\num{1.66092} \pm \num{0.04731}$ & $\num{1.86479} \pm \num{0.04663}$ & -- & -- \\\precogrow
 & DESIRE-plan & $\num{2.34311} \pm \num{0.04724}$ & $\num{0.23222} \pm \num{0.00896}$ & $\num{3.12998} \pm \num{0.07784}$ & $\num{3.66712} \pm \num{0.09646}$	& --	& -- \\ \precogrow
& ESP & $\num{0.42641} \pm \num{0.01320}$ &	$\num{0.20427} \pm \num{0.00856}$	& $\num{0.55607} \pm \num{0.02707} $ & $\num{0.51887} \pm \num{0.02124}$ & -- & --  \\ \precogrow
& PRECOG & ${\boldmath  \num{0.35464} \pm \num{0.01170}}$	& ${\boldmath \num{0.05175} \pm \num{0.00339}}$ & ${\boldmath \num{0.51927}\pm \num{0.02536}} $ & ${\boldmath  \num{0.49290} \pm \num{0.02036}}$ & -- & -- \\ \precogrow
 \midrule
\multirow{4}{*}{CARLA\, $A\!=\!4$} & DESIRE & $\num{2.40163} \pm \num{0.03761}$ & $\num{2.42204} \pm \num{0.05373}$ & $\num{2.06456} \pm \num{0.04388}$ & $\num{2.53084} \pm \num{0.07142}$ & $\num{2.58909} \pm \num{0.06445}$ & -- \\
& DESIRE-plan & $\num{1.82801} \pm \num{0.03488}$ & $\num{0.14943} \pm \num{0.00366}$ & $\num{2.48027} \pm \num{0.06241}$ & $\num{1.25603} \pm \num{0.04731}$ & $\num{3.42629} \pm \num{0.09763}$ & -- \\
& ESP & $\num{0.53743} \pm \num{0.01120}$ & 	$\num{0.23587} \pm \num{0.00893}$ &	$\num{0.61460} \pm \num{0.02142}$ &	$\num{0.65640} \pm \num{0.02283}$ &	$\num{0.64285} \pm \num{0.02342}$ 
 & --  \\ \precogrow
 & PRECOG & ${\boldmath \num{0.47798} \pm \num{0.01078}}$	& ${\boldmath \num{0.05397}\pm \num{0.00323}} $	& 	${\boldmath \num{0.58304} \pm \num{0.02056}}$	& 	${\boldmath \num{0.63710} \pm \num{0.02208}}$		& ${\boldmath \num{0.63779} \pm \num{0.02291}}$ & -- \\ \precogrow
  \midrule
\multirow{4}{*}{CARLA\, $A\!=\!5$} & DESIRE & $\num{2.62247} \pm \num{0.02953}$ & $\num{2.62066} \pm \num{0.04547}$ & $\num{2.42165} \pm \num{0.04842}$ & $\num{2.71018} \pm \num{0.06613}$ & $\num{2.96901} \pm \num{0.05700}$ & $\num{2.39086} \pm \num{0.04891}$ \\
& DESIRE-plan & $\num{2.32850} \pm \num{0.03837}$ & $\num{0.19446} \pm \num{0.00359}$ & $\num{2.23880} \pm \num{0.05707}$ & $\num{3.11885} \pm \num{0.09775}$ & $\num{3.33222} \pm \num{0.09021}$ & $\num{2.75818} \pm \num{0.08278}$ \\ 
& ESP & $\num{0.71759} \pm \num{0.01187}$ & $\num{0.34031} \pm \num{0.01146}$ & $\num{0.75908} \pm \num{0.02449}$ & $\num{0.80909} \pm \num{0.02504}$ & $\num{0.85122} \pm \num{0.02335}$ & $\num{0.82827} \pm \num{0.02411}$ \\ \precogrow
 & PRECOG & ${\boldmath \num{0.64039} \pm \num{0.01109}}$ & ${\boldmath \num{0.06557}\pm \num{0.00348}} $ & ${\boldmath \num{0.74088} \pm \num{0.02363}}$ & ${\boldmath \num{0.79019}\pm \num{0.02449}} $ & ${\boldmath \num{0.80444} \pm \num{0.02238}}$ & ${\boldmath \num{0.80085} \pm \num{0.02401}}$\\ \precogrow
\midrule
\multirow{4}{*}{nuScenes\, $A\!=\!2$} & DESIRE & $\num{3.30741} \pm \num{0.09280}$ & $\num{3.00155} \pm \num{0.08819}$ & $\num{3.61328} \pm \num{0.13969}$ & -- & -- & -- \\ \precogrow
& DESIRE-plan & $\num{4.52804} \pm \num{0.15055}$ & $\num{0.45641} \pm \num{0.01544}$ & $\num{8.59966} \pm \num{0.29765}$ & -- &	-- &	-- \\\precogrow
& ESP & $\num{1.09373} \pm \num{0.05281}$ & $\num{0.95464} \pm \num{0.05651}$ & $\num{1.23282} \pm \num{0.07814}$ & -- &  -- & --\\ \precogrow
& PRECOG &  $\boldmath \num{0.51431} \pm \num{0.03667}$ & $\boldmath \num{0.15778} \pm \num{0.01621}$ & $\boldmath \num{0.87085} \pm \num{0.07002}$ & -- & -- & --\\ \precogrow
\midrule
\multirow{4}{*}{nuScenes\, $A\!=\!3$} & DESIRE & $\num{4.84047} \pm \num{0.13493}$ & $\num{3.93144} \pm \num{0.12722}$ & $\num{4.98354} \pm \num{0.20695}$ & $\num{5.60645} \pm \num{0.23404}$ & --	& -- \\ \precogrow
& DESIRE-plan & $\num{5.88704} \pm \num{0.18713}$ & $\num{0.40910} \pm \num{0.01451}$ & $\num{7.73086} \pm \num{0.33673}$ & $\num{9.52117} \pm \num{0.39943}$ &  -- & -- \\
& ESP & $\num{1.51103} \pm \num{0.07741}$ & $\num{1.12817} \pm \num{0.06052}$ & $\num{1.54337} \pm \num{0.11801}$ & $\num{1.86155} \pm \num{0.14696}$ & --& --\\ \precogrow
 & PRECOG & $\boldmath \num{1.01569} \pm \num{0.06249}$ & $\boldmath \num{0.12111} \pm \num{0.00457}$ & $\boldmath \num{1.32020} \pm \num{0.10501}$ & $\boldmath \num{1.60576} \pm \num{0.12173}$ & -- & -- \\ \precogrow
 \midrule
\multirow{4}{*}{nuScenes\, $A\!=\!4$} & DESIRE  & $\num{5.77079} \pm \num{0.15115}$ & $\num{4.19479} \pm \num{0.15876}$ & $\num{5.85411} \pm \num{0.24286}$ & $\num{6.13803} \pm \num{0.28031}$ & $\num{6.89623} \pm \num{0.32431}$ 	& -- \\\precogrow
& DESIRE-plan & $\num{5.04545} \pm \num{0.15842}$ & $\num{0.47126} \pm \num{0.01900}$ & $\num{5.56701} \pm \num{0.24471}$ & $\num{5.49193} \pm \num{0.25697}$ & $\num{8.65161} \pm \num{0.40677}$&	-- \\\precogrow
&ESP & $\num{2.20022} \pm \num{0.08957}$& $\num{1.60400} \pm \num{0.09882}$& $\num{1.94018} \pm \num{0.12261}$& $\num{2.40535} \pm \num{0.14867}$& $\num{2.85133} \pm \num{0.21334}$& --\\ \precogrow
 & PRECOG & $\boldmath \num{1.75499} \pm \num{0.08254}$ & $\boldmath \num{0.13314} \pm \num{0.00591}$ & $\boldmath \num{1.80359} \pm \num{0.12562}$ & $\boldmath \num{2.31905} \pm \num{0.14119}$ & $\boldmath \num{2.76419} \pm \num{0.23126}$ & -- \\ \precogrow
  \midrule
\multirow{4}{*}{nuScenes\, $A\!=\!5$} & DESIRE & $\num{6.82986} \pm \num{0.20435}$ & $\num{4.99886} \pm \num{0.21870}$ & $\num{6.41475} \pm \num{0.29442}$ & $\num{7.02696} \pm \num{0.36014}$ & 	$\num{7.41826} \pm \num{0.32365}$ & $\num{8.29048} \pm \num{0.53172}$ \\\precogrow
& DESIRE-plan & $\num{6.56219} \pm \num{0.20719}$ & $\num{2.26053} \pm \num{0.09984}$ & $\num{6.64410} \pm \num{0.31435}$ & $\num{6.18376} \pm \num{0.32513}$ & $\num{9.20305} \pm \num{0.44800}$ & 	$\num{8.51950} \pm \num{0.51390}$ \\\precogrow
& ESP & $\num{2.92126} \pm \num{0.17499}$ & $ \num{1.86066} \pm \num{0.10935}$ & $\num{2.36853} \pm \num{0.18780}$ & $ \num{2.81241} \pm \num{0.18794}$ & $\num{3.20137} \pm \num{0.25363}$ & $\num{4.36335} \pm \num{0.65235}$ \\ \precogrow
& PRECOG & $\boldmath \num{2.50763} \pm \num{0.15214}$ & $\boldmath \num{0.14913} \pm \num{0.02075}$ & $\boldmath \num{2.32361} \pm \num{0.18743}$ & $\boldmath \num{2.65441} \pm \num{0.19017}$ & $\boldmath \num{3.15719} \pm \num{0.27262}$ & $\boldmath \num{4.25379} \pm \num{0.58602}$ \\
 \bottomrule
\end{tabular}
}
\end{figure*}
 
 \subsection{Additional CARLA and nuScenes Evaluations.}
\begin{figure}[H]
\centering
 \begin{subfigure}{\columnwidth}
\includegraphics[width=\textwidth]{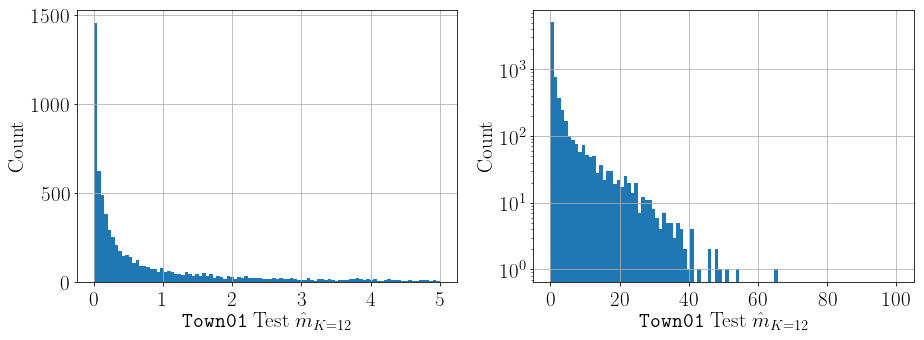}
\end{subfigure}
\caption{Histogram of $\hat m_{K=12}$ of forecasts made by the ESP flexible-count model on CARLA {\tt Town01} Test $A=5$, $T=40$ at $10$Hz (4 seconds of future). The median $\hat m_{K=12}$ is $0.38$.}  \label{fig:town01testhist}
\end{figure}

We show additional evaluations on CARLA in \tab{carla_forecast_town1}. Table~\ref{tab:carla_forecast_town1} shows the {\tt Town01} of the models trained on {\tt Town01} (on separate episodes). We show single-agent CARLA forecasting results in \tab{carla_forecast_A1}. We show histograms of $\hat m$ in \fig{town02testhist}, \fig{nuscenestesthist}, and \fig{town01testhist}. We show a comparison to longer time-horizon forecasting in \tab{carla_forecast_T40}. We show a plot of means and their standard errors of $\hat m_K$ vs. $K$ in \fig{testmvsK}.

\begin{table}[htb]
\centering
\caption{Performance on CARLA Town01 Test with $T=40$ at $10$Hz (4 seconds of future). This data has larger dimensionality than CARLA $T=20$, $10$Hz (2 seconds) data and the nuScenes $T=20$, $5$Hz (4 seconds) data.}
\label{tab:carla_forecast_T40} 
\resizebox{1.0\linewidth}{!}{
\begin{tabular}{lcc}
\toprule
Approach  & Test $\hat{m}_{K\!=\!12}$ & Test $\hat{e}$  \\
& (minMSD)                  & (extra nats)   \\
\midrule
\textbf{Town01 Test}, $T=20$, $10$Hz (2s) & \multicolumn{2}{c}{5 agent} \\
\cmidrule{2-3}      
ESP, flex. count & $\num{0.446775} \pm \num{0.009190}$ & $\num{0.508599} \pm \num{0.001097}$ \\
\midrule
\textbf{Town02 Test}, $T=20$, $10$Hz (2s) & \multicolumn{2}{c}{5 agent} \\
\cmidrule{2-3}    
ESP, flex. count & $\num{0.435110} \pm \num{0.010517}$ & $\num{0.495612} \pm \num{0.001090}$\\
\midrule
\textbf{Town01 Test}, $T=40$, $10$Hz (4s) & \multicolumn{2}{c}{5 agent} \\
\cmidrule{2-3}                                    
ESP, flex. count                                      & $\num{2.500401} \pm \num{.076503}$  & $\num{0.491603} \pm \num{0.001252}$        \\
\midrule
\textbf{nuScenes Test}, $T=20$, $5$Hz (4s) & \multicolumn{2}{c}{5 agent} \\
\cmidrule{2-3}  
ESP, flex. count & $\num{2.933192} \pm \num{0.128841}$ & $\num{1.028862} \pm \num{0.002387}$\\
\bottomrule
\end{tabular}
}
\end{table}

\subsection{Full Conditional Forecasting Experiments}
Due to main-text space limits, we report some remaining results (i.e. for $A=\{3,4\}$) in \tab{planning_forecast_app}. We observe similar trends in these results as in $A=2$ and $A=5$: PRECOG improves predictions of all agents' future trajectories, and that knowledge of the ego-agent's goal provides improves predictions for closer agents more than farther agents.

 \section{Additional Visualizations} \label{app:visualizations}
\begin{figure*}[htpb]
    \centering
    \begin{subfigure}{\nuscappwidth}
    \FBox{ 
    \begin{overpic}[width=\textwidth]{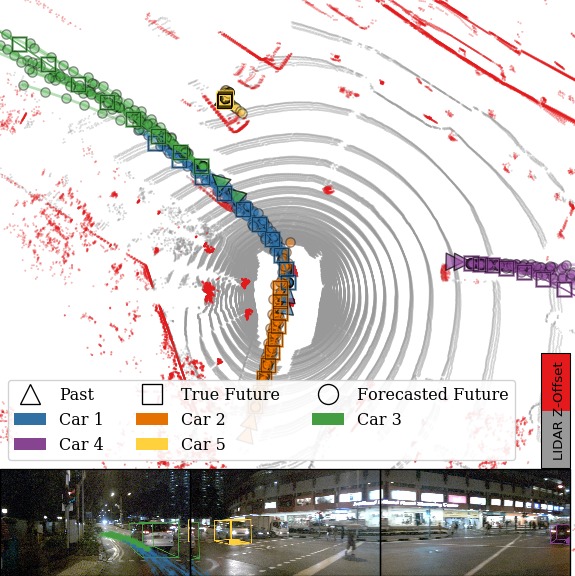}
    \put(0,15.5){{\setlength{\fboxsep}{1pt}\fontsize{5pt}{0pt}\selectfont\transparent{0.7}\colorbox{white}{\textcolor{black}{\texttransparent{1.0}{Left\vphantom{g}}}}}}
    \put(33,15.5){{\setlength{\fboxsep}{1pt}\fontsize{5pt}{0pt}\selectfont\transparent{0.7}\colorbox{white}{\textcolor{black}{\texttransparent{1.0}{Front\vphantom{g}}}}}}
    \put(66,15.5){{\setlength{\fboxsep}{1pt}\fontsize{5pt}{0pt}\selectfont\transparent{0.7}\colorbox{white}{\textcolor{black}{\texttransparent{1.0}{Right}}}}}
    \end{overpic}
    }
    \end{subfigure}
    \begin{subfigure}{\nuscappwidth}
    \FBox{ 
    \begin{overpic}[width=\textwidth]{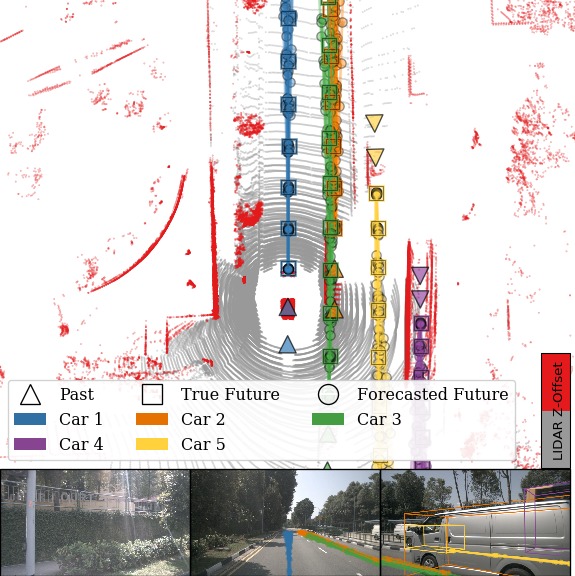}
      \put(0,15.5){{\setlength{\fboxsep}{1pt}\fontsize{5pt}{0pt}\selectfont\transparent{0.7}\colorbox{white}{\textcolor{black}{\texttransparent{1.0}{Left\vphantom{g}}}}}}
    \put(33,15.5){{\setlength{\fboxsep}{1pt}\fontsize{5pt}{0pt}\selectfont\transparent{0.7}\colorbox{white}{\textcolor{black}{\texttransparent{1.0}{Front\vphantom{g}}}}}}
    \put(66,15.5){{\setlength{\fboxsep}{1pt}\fontsize{5pt}{0pt}\selectfont\transparent{0.7}\colorbox{white}{\textcolor{black}{\texttransparent{1.0}{Right}}}}}
    \end{overpic}
     }
    \end{subfigure}
        \begin{subfigure}{\nuscappwidth}
    \FBox{ 
    \begin{overpic}[width=\textwidth]{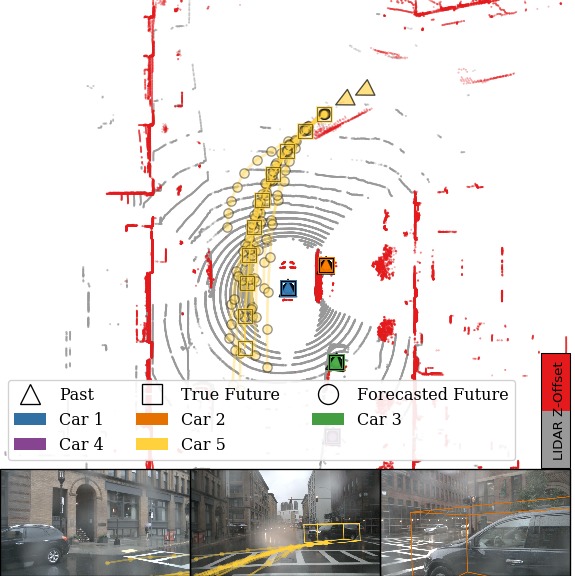}
      \put(0,15.5){{\setlength{\fboxsep}{1pt}\fontsize{5pt}{0pt}\selectfont\transparent{0.7}\colorbox{white}{\textcolor{black}{\texttransparent{1.0}{Left\vphantom{g}}}}}}
    \put(33,15.5){{\setlength{\fboxsep}{1pt}\fontsize{5pt}{0pt}\selectfont\transparent{0.7}\colorbox{white}{\textcolor{black}{\texttransparent{1.0}{Front\vphantom{g}}}}}}
    \put(66,15.5){{\setlength{\fboxsep}{1pt}\fontsize{5pt}{0pt}\selectfont\transparent{0.7}\colorbox{white}{\textcolor{black}{\texttransparent{1.0}{Right}}}}}
    \end{overpic}
     }
    \end{subfigure}
    \begin{subfigure}{\nuscappwidth}
    \FBox{ 
    \begin{overpic}[width=\textwidth]{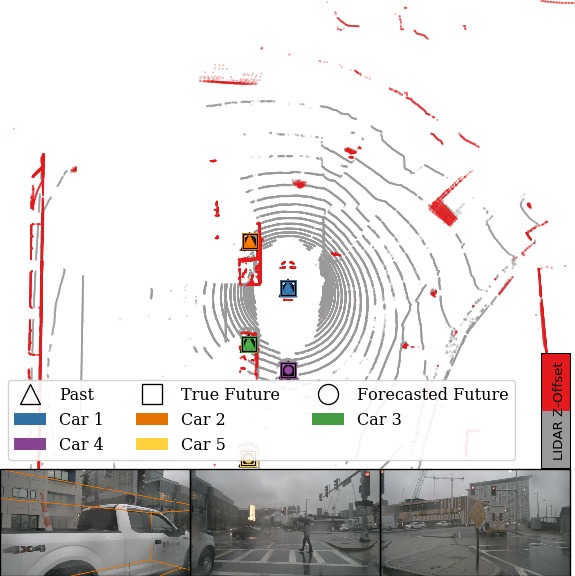}
  \put(0,15.5){{\setlength{\fboxsep}{1pt}\fontsize{5pt}{0pt}\selectfont\transparent{0.7}\colorbox{white}{\textcolor{black}{\texttransparent{1.0}{Left\vphantom{g}}}}}}
    \put(33,15.5){{\setlength{\fboxsep}{1pt}\fontsize{5pt}{0pt}\selectfont\transparent{0.7}\colorbox{white}{\textcolor{black}{\texttransparent{1.0}{Front\vphantom{g}}}}}}
    \put(66,15.5){{\setlength{\fboxsep}{1pt}\fontsize{5pt}{0pt}\selectfont\transparent{0.7}\colorbox{white}{\textcolor{black}{\texttransparent{1.0}{Right}}}}}
    \end{overpic}
     }
    \end{subfigure} 
        \begin{subfigure}{\nuscappwidth}
    \FBox{ 
    \begin{overpic}[width=\textwidth]{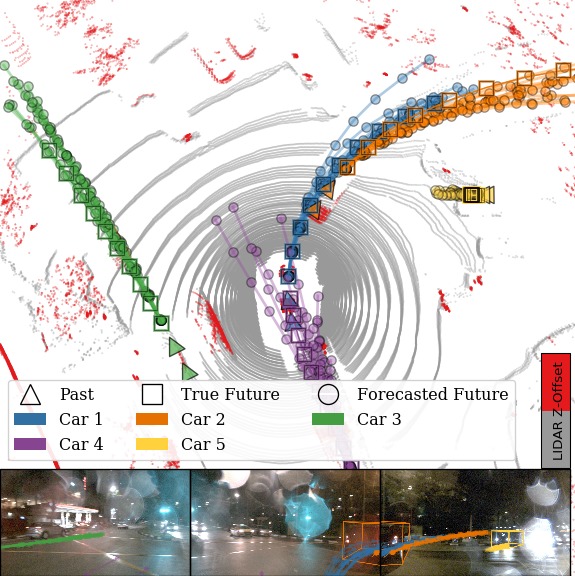}
  \put(0,15.5){{\setlength{\fboxsep}{1pt}\fontsize{5pt}{0pt}\selectfont\transparent{0.7}\colorbox{white}{\textcolor{black}{\texttransparent{1.0}{Left\vphantom{g}}}}}}
    \put(33,15.5){{\setlength{\fboxsep}{1pt}\fontsize{5pt}{0pt}\selectfont\transparent{0.7}\colorbox{white}{\textcolor{black}{\texttransparent{1.0}{Front\vphantom{g}}}}}}
    \put(66,15.5){{\setlength{\fboxsep}{1pt}\fontsize{5pt}{0pt}\selectfont\transparent{0.7}\colorbox{white}{\textcolor{black}{\texttransparent{1.0}{Right}}}}}
    \end{overpic}
     }
    \end{subfigure} 
            \begin{subfigure}{\nuscappwidth}
    \FBox{ 
    \begin{overpic}[width=\textwidth]{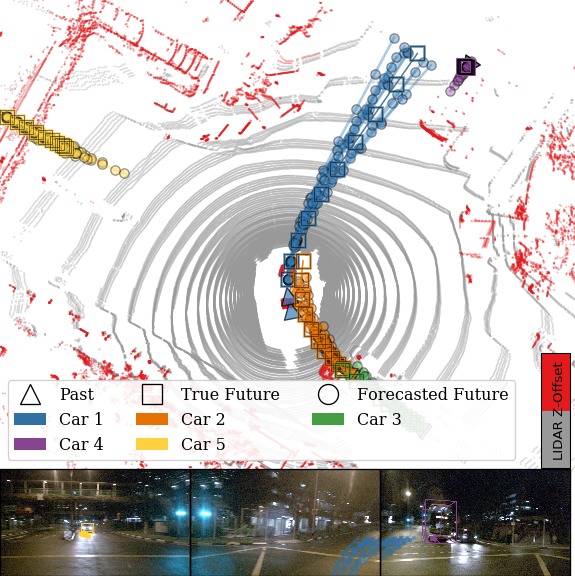}
  \put(0,15.5){{\setlength{\fboxsep}{1pt}\fontsize{5pt}{0pt}\selectfont\transparent{0.7}\colorbox{white}{\textcolor{black}{\texttransparent{1.0}{Left\vphantom{g}}}}}}
    \put(33,15.5){{\setlength{\fboxsep}{1pt}\fontsize{5pt}{0pt}\selectfont\transparent{0.7}\colorbox{white}{\textcolor{black}{\texttransparent{1.0}{Front\vphantom{g}}}}}}
    \put(66,15.5){{\setlength{\fboxsep}{1pt}\fontsize{5pt}{0pt}\selectfont\transparent{0.7}\colorbox{white}{\textcolor{black}{\texttransparent{1.0}{Right}}}}}
    \end{overpic}
     }
    \end{subfigure} 
            \begin{subfigure}{\nuscappwidth}
    \FBox{ 
    \begin{overpic}[width=\textwidth]{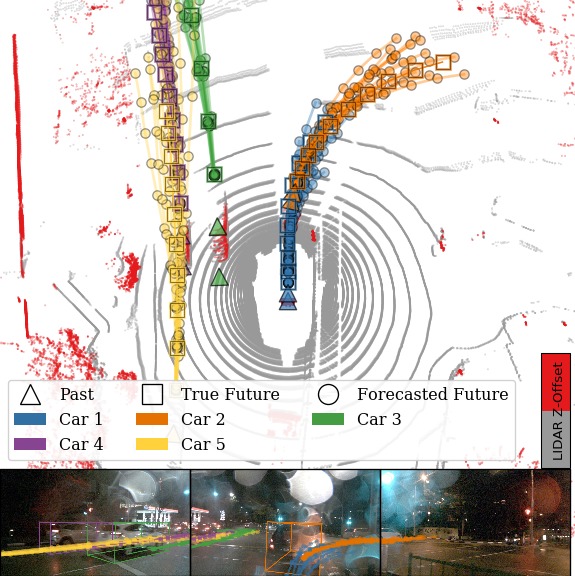}
  \put(0,15.5){{\setlength{\fboxsep}{1pt}\fontsize{5pt}{0pt}\selectfont\transparent{0.7}\colorbox{white}{\textcolor{black}{\texttransparent{1.0}{Left\vphantom{g}}}}}}
    \put(33,15.5){{\setlength{\fboxsep}{1pt}\fontsize{5pt}{0pt}\selectfont\transparent{0.7}\colorbox{white}{\textcolor{black}{\texttransparent{1.0}{Front\vphantom{g}}}}}}
    \put(66,15.5){{\setlength{\fboxsep}{1pt}\fontsize{5pt}{0pt}\selectfont\transparent{0.7}\colorbox{white}{\textcolor{black}{\texttransparent{1.0}{Right}}}}}
    \end{overpic}
     }
    \end{subfigure} 
    \begin{subfigure}{\nuscappwidth}
    \FBox{ 
    \begin{overpic}[width=\textwidth]{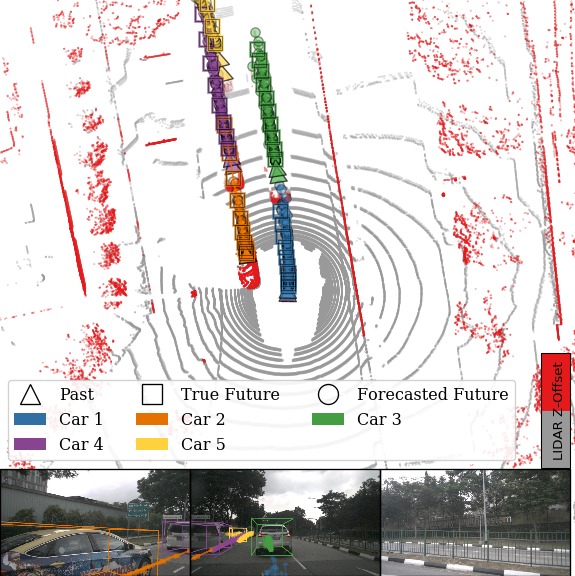}
  \put(0,15.5){{\setlength{\fboxsep}{1pt}\fontsize{5pt}{0pt}\selectfont\transparent{0.7}\colorbox{white}{\textcolor{black}{\texttransparent{1.0}{Left\vphantom{g}}}}}}
    \put(33,15.5){{\setlength{\fboxsep}{1pt}\fontsize{5pt}{0pt}\selectfont\transparent{0.7}\colorbox{white}{\textcolor{black}{\texttransparent{1.0}{Front\vphantom{g}}}}}}
    \put(66,15.5){{\setlength{\fboxsep}{1pt}\fontsize{5pt}{0pt}\selectfont\transparent{0.7}\colorbox{white}{\textcolor{black}{\texttransparent{1.0}{Right}}}}}
    \end{overpic}
     }
    \end{subfigure} 
    \begin{subfigure}{\nuscappwidth}
    \FBox{ 
    \begin{overpic}[width=\textwidth]{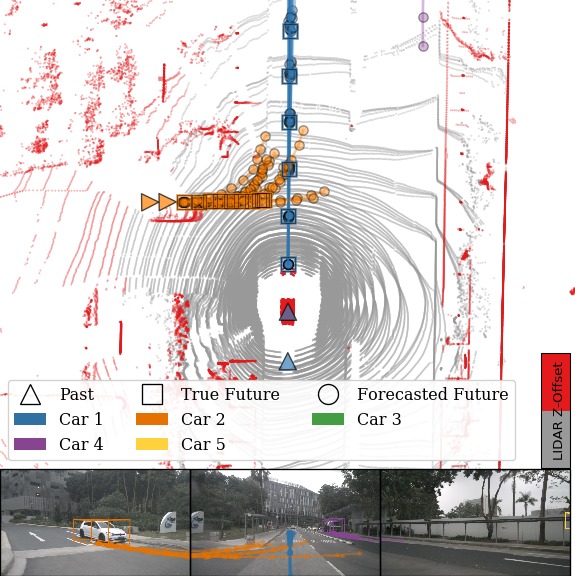}
  \put(0,15.5){{\setlength{\fboxsep}{1pt}\fontsize{5pt}{0pt}\selectfont\transparent{0.7}\colorbox{white}{\textcolor{black}{\texttransparent{1.0}{Left\vphantom{g}}}}}}
    \put(33,15.5){{\setlength{\fboxsep}{1pt}\fontsize{5pt}{0pt}\selectfont\transparent{0.7}\colorbox{white}{\textcolor{black}{\texttransparent{1.0}{Front\vphantom{g}}}}}}
    \put(66,15.5){{\setlength{\fboxsep}{1pt}\fontsize{5pt}{0pt}\selectfont\transparent{0.7}\colorbox{white}{\textcolor{black}{\texttransparent{1.0}{Right}}}}}
    \end{overpic}
     }
    \end{subfigure}
    \begin{subfigure}{\nuscappwidth}
    \FBox{ 
    \begin{overpic}[width=\textwidth]{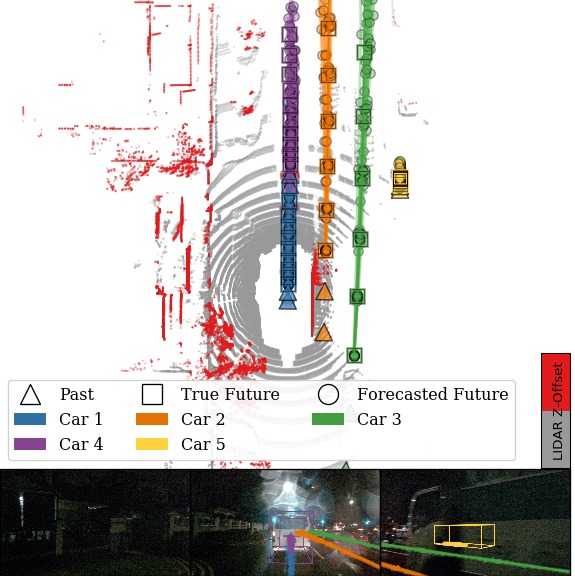}
  \put(0,15.5){{\setlength{\fboxsep}{1pt}\fontsize{5pt}{0pt}\selectfont\transparent{0.7}\colorbox{white}{\textcolor{black}{\texttransparent{1.0}{Left\vphantom{g}}}}}}
    \put(33,15.5){{\setlength{\fboxsep}{1pt}\fontsize{5pt}{0pt}\selectfont\transparent{0.7}\colorbox{white}{\textcolor{black}{\texttransparent{1.0}{Front\vphantom{g}}}}}}
    \put(66,15.5){{\setlength{\fboxsep}{1pt}\fontsize{5pt}{0pt}\selectfont\transparent{0.7}\colorbox{white}{\textcolor{black}{\texttransparent{1.0}{Right}}}}}
    \end{overpic}
     }
    \end{subfigure}
    \begin{subfigure}{\nuscappwidth}
    \FBox{ 
    \begin{overpic}[width=\textwidth]{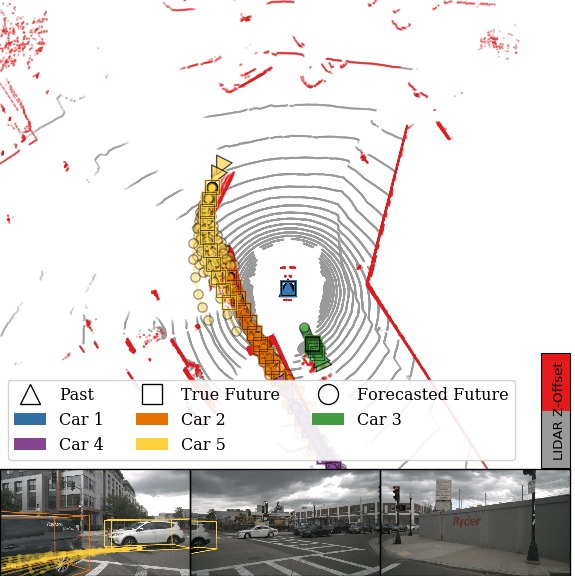}
  \put(0,15.5){{\setlength{\fboxsep}{1pt}\fontsize{5pt}{0pt}\selectfont\transparent{0.7}\colorbox{white}{\textcolor{black}{\texttransparent{1.0}{Left\vphantom{g}}}}}}
    \put(33,15.5){{\setlength{\fboxsep}{1pt}\fontsize{5pt}{0pt}\selectfont\transparent{0.7}\colorbox{white}{\textcolor{black}{\texttransparent{1.0}{Front\vphantom{g}}}}}}
    \put(66,15.5){{\setlength{\fboxsep}{1pt}\fontsize{5pt}{0pt}\selectfont\transparent{0.7}\colorbox{white}{\textcolor{black}{\texttransparent{1.0}{Right}}}}}
    \end{overpic}
     }
    \end{subfigure}
    \begin{subfigure}{\nuscappwidth}
    \FBox{ 
    \begin{overpic}[width=\textwidth]{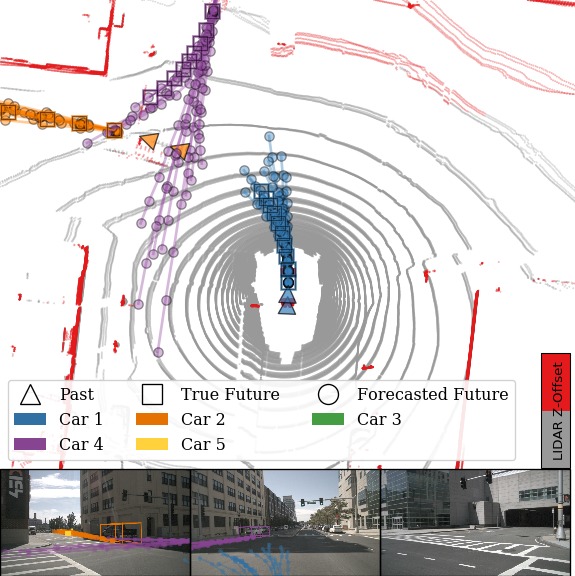}
  \put(0,15.5){{\setlength{\fboxsep}{1pt}\fontsize{5pt}{0pt}\selectfont\transparent{0.7}\colorbox{white}{\textcolor{black}{\texttransparent{1.0}{Left\vphantom{g}}}}}}
    \put(33,15.5){{\setlength{\fboxsep}{1pt}\fontsize{5pt}{0pt}\selectfont\transparent{0.7}\colorbox{white}{\textcolor{black}{\texttransparent{1.0}{Front\vphantom{g}}}}}}
    \put(66,15.5){{\setlength{\fboxsep}{1pt}\fontsize{5pt}{0pt}\selectfont\transparent{0.7}\colorbox{white}{\textcolor{black}{\texttransparent{1.0}{Right}}}}}
    \end{overpic}
     }
    \end{subfigure}
    
    \caption{Example forecasting results on held-out nuScenes data with our learned ESP model. In each scene, 12 joint samples are shown, and LIDAR colors are discretized to near-ground and above-ground}
    \label{fig:nuscenes_test_vis}
\end{figure*}

\begin{figure*}[htp]
    \centering
    \FBox{ 
    \begin{subfigure}[t]{\nuscappwidth}
       \includegraphics[width=\textwidth,clip]{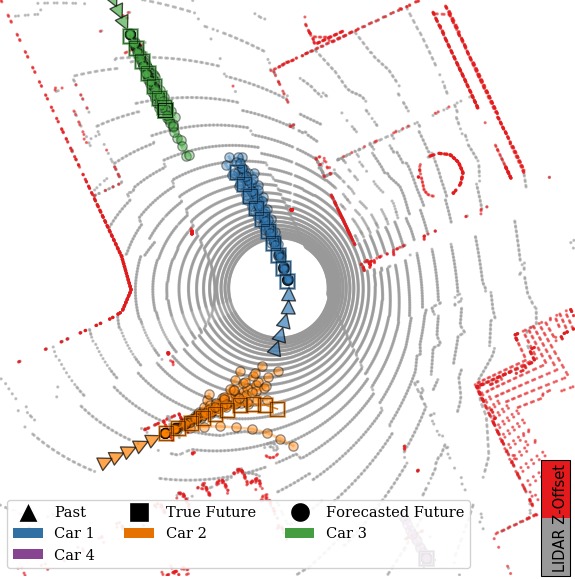}
       \end{subfigure}
    }
    \FBox{     \begin{subfigure}[t]{\nuscappwidth}
        \includegraphics[width=\textwidth]{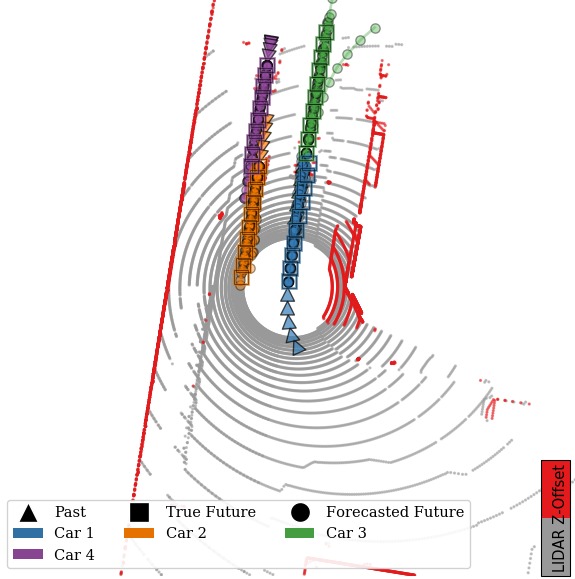}
    \end{subfigure} }
      \FBox{     \begin{subfigure}[t]{\nuscappwidth}
        \includegraphics[width=\textwidth]{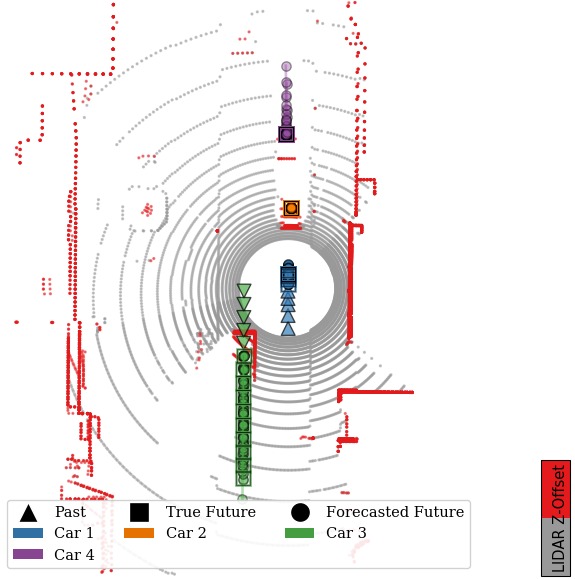}
    \end{subfigure} }
        \FBox{ 
    \begin{subfigure}[t]{\nuscappwidth}
       \includegraphics[width=\textwidth,clip]{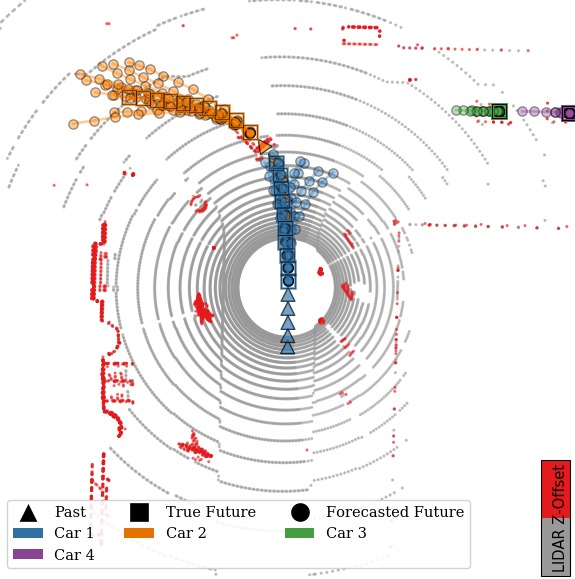}
       \end{subfigure}
    }
    \FBox{     \begin{subfigure}[t]{\nuscappwidth}
        \includegraphics[width=\textwidth]{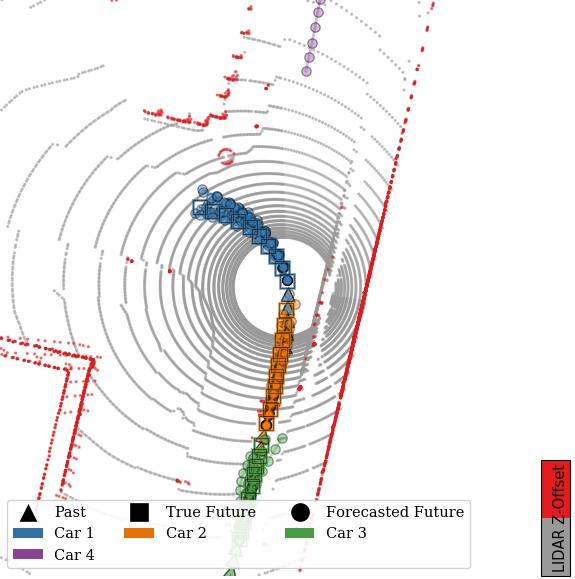}
    \end{subfigure} }
      \FBox{     \begin{subfigure}[t]{\nuscappwidth}
        \includegraphics[width=\textwidth]{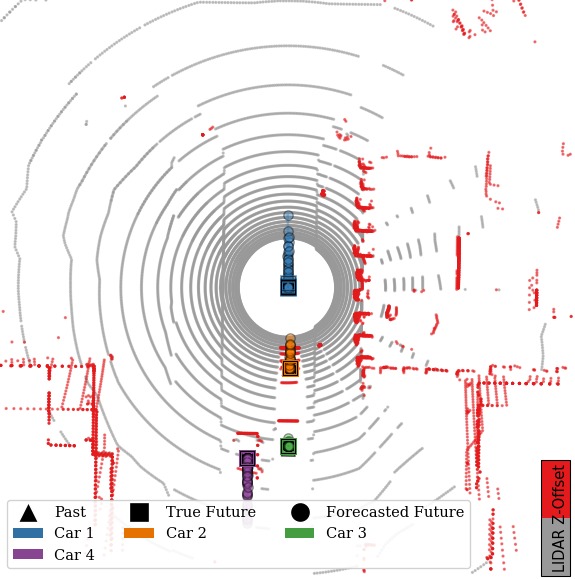}
    \end{subfigure} }
    \FBox{     \begin{subfigure}[t]{\nuscappwidth}
    \includegraphics[width=\textwidth]{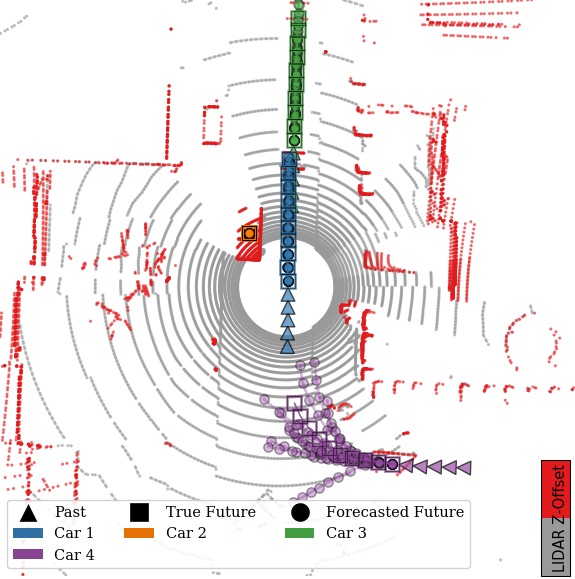}
    \end{subfigure}     }
        \FBox{     \begin{subfigure}[t]{\nuscappwidth}
    \includegraphics[width=\textwidth]{fig/carla/a5/frame_00000_00000051_008.jpg}
    \end{subfigure}     }
            \FBox{     \begin{subfigure}[t]{\nuscappwidth}
    \includegraphics[width=\textwidth]{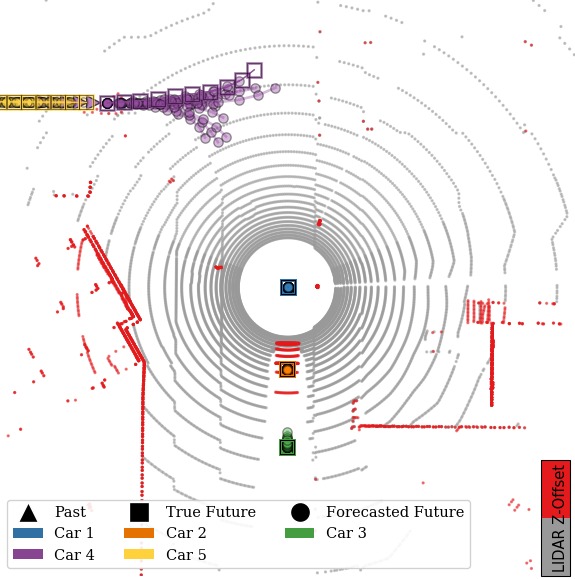}
    \end{subfigure}     }
    \FBox{     \begin{subfigure}[t]{\nuscappwidth}
    \includegraphics[width=\textwidth]{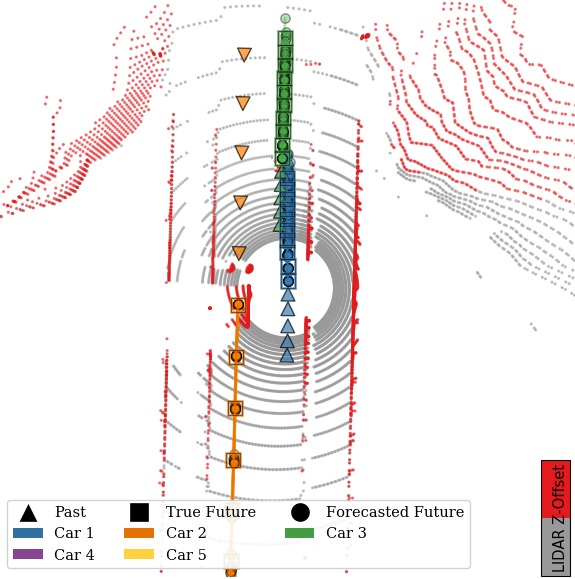}
    \end{subfigure}     }
        \FBox{     \begin{subfigure}[t]{\nuscappwidth}
    \includegraphics[width=\textwidth]{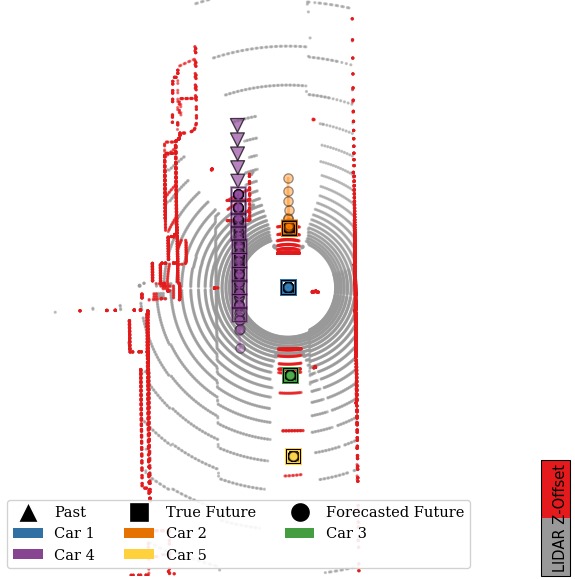}
    \end{subfigure}     }
            \FBox{     \begin{subfigure}[t]{\nuscappwidth}
    \includegraphics[width=\textwidth]{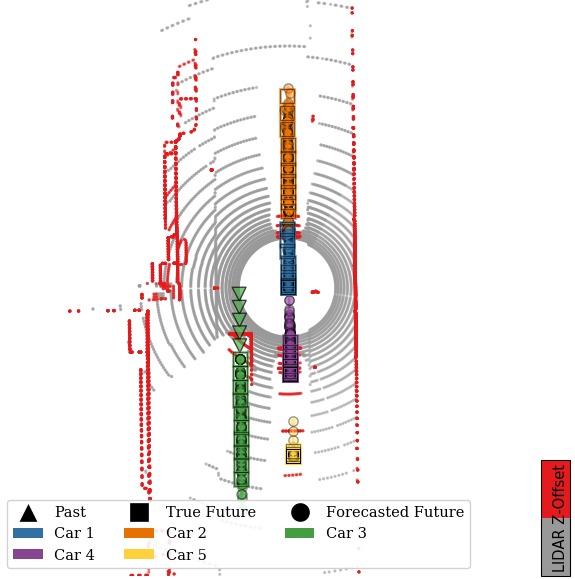}
    \end{subfigure}     }
    \caption{
    Examples of multi-agent forecasting with our learned ESP model. In each scene, 12 joint samples are shown, and LIDAR colors are discretized to near-ground and above-ground. 
    } 
    \label{fig:example_forecast_app}
\end{figure*}

We display additional visualization of our results in Figures~\ref{fig:nuscenes_test_vis}, \ref{fig:example_forecast_app}, \ref{fig:app_example_carla_plan}, \ref{fig:app_example_nuscenes_plan}, and \ref{fig:posterior_scan_carla}. In \fig{nuscenes_test_vis}, we show additional forecasting results on the nuScenes dataset. In \fig{example_forecast_app}, we show additional forecasting results on the CARLA dataset. In \fig{app_example_carla_plan}, we show additional planning results on the CARLA dataset. In \fig{app_example_nuscenes_plan}, we show additional planning results on the nuScenes dataset. 
In \fig{best_middle_worst_carla}, we show qualitative results of high, medium, and low quality on the CARLA $A=5$ dataset (ordered by $\hat m$), paired with their corresponding $\hat m$ scores. 
In \fig{best_middle_worst_nuscenes}, we show qualitative results of high, medium, and low quality on the nuScenes $A=5$ dataset (ordered by $\hat m$), paired with their corresponding $\hat m$ scores. 
In \fig{posterior_scan_carla}, we visualize the planning criterion ($\hat L$) across many different spatio-temporal goal positions in CARLA, which gives a qualitative interpretation of where the model prefers goal. In \fig{posterior_scan_nuscenes}, we visualize the same posterior on nuScenes.

\begin{figure}[H]
\centering
 \begin{subfigure}{\columnwidth}
\includegraphics[width=.5\textwidth]{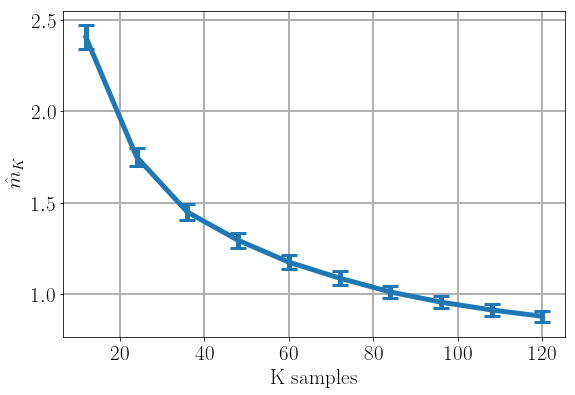}
\caption{Plot of $\hat m_K$ vs. $K$ of the ESP flexible-count model on CARLA {\tt Town01} Test $A=5$, $T=40$ at $10$Hz (4s).} 
\end{subfigure}
\end{figure}

\clearpage 
\pagestyle{empty} 

 \newcommand{\plappwidth}[0]{.28\textwidth}

 \begin{figure*}[htp]
    \centering
    \begin{subfigure}{\plappwidth}
    \includegraphics[width=\textwidth]{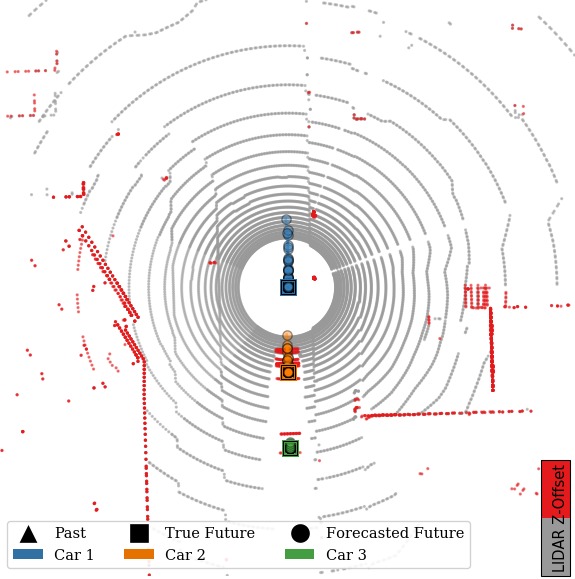}
    \caption{Scene 1, forecasted}
    \end{subfigure}
    \begin{subfigure}{\plappwidth}
    \includegraphics[width=\textwidth]{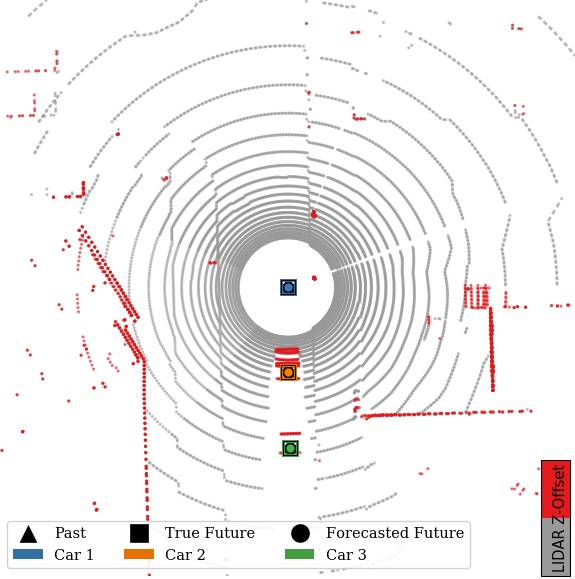}
     \caption{Scene 1, planned}
    \end{subfigure} \\

            \begin{subfigure}{\plappwidth}
    \includegraphics[width=\textwidth]{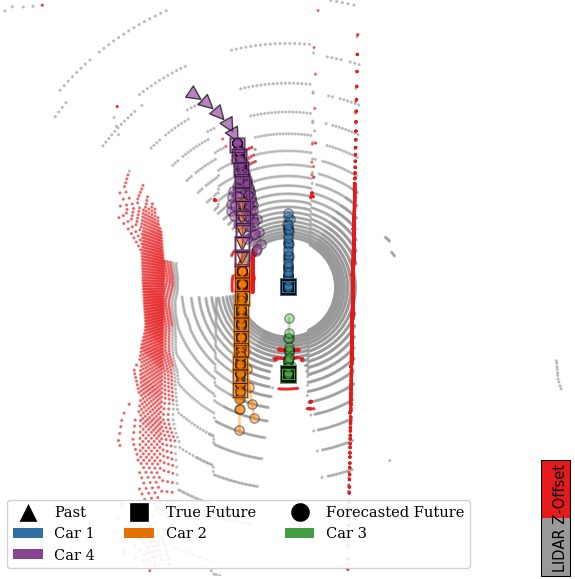}
      \caption{Scene 2, forecasted}
    \end{subfigure}
    \begin{subfigure}{\plappwidth}
    \includegraphics[width=\textwidth]{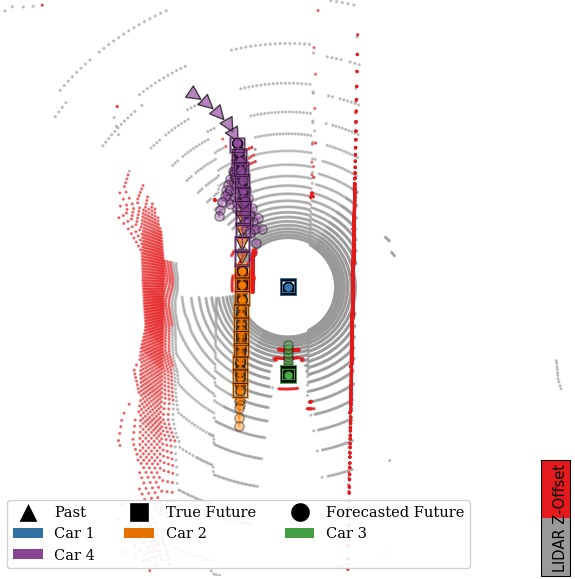}
         \caption{Scene 2, planned}
    \end{subfigure} \\
    
                \begin{subfigure}{\plappwidth}
    \includegraphics[width=\textwidth]{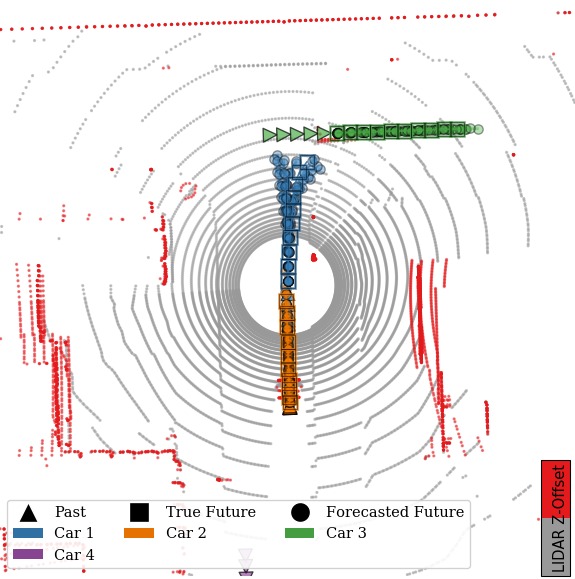}
      \caption{Scene 3, forecasted}
    \end{subfigure}
    \begin{subfigure}{\plappwidth}
    \includegraphics[width=\textwidth]{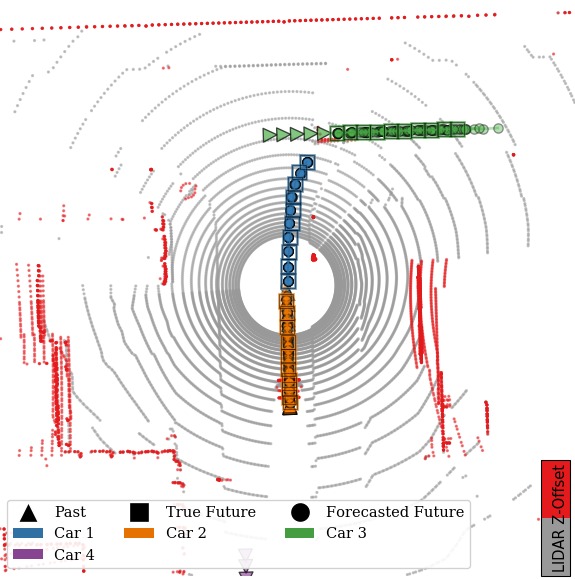}
         \caption{Scene 3, planned}
    \end{subfigure} \\
    
            \begin{subfigure}{\plappwidth}
    \includegraphics[width=\textwidth]{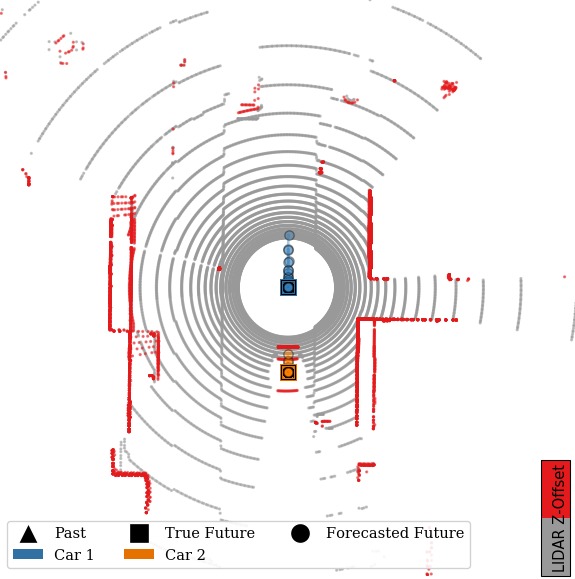}
      \caption{Scene 4, forecasted}
    \end{subfigure}
    \begin{subfigure}{\plappwidth}
    \includegraphics[width=\textwidth]{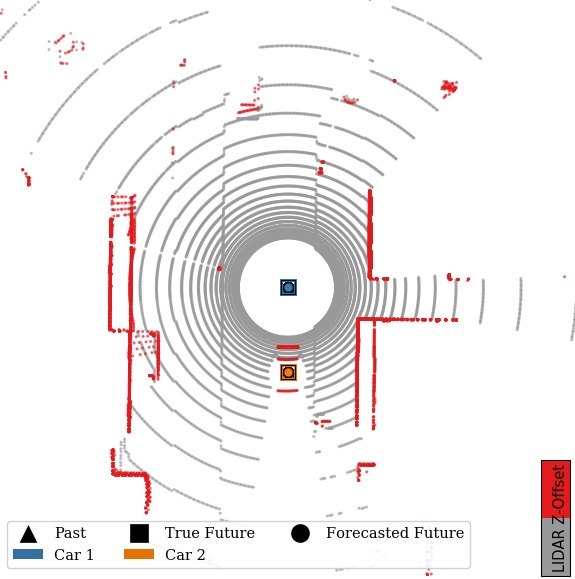}
         \caption{Scene 4, planned} 
    \end{subfigure} 
    
    \caption{Additional examples of \emph{planned} multi-agent forecasting (PRECOG) with our learned model in CARLA. By using our planning approach and conditioning the robot on its true final position, our predictions for the robot become more accurate, and often our predictions of the other agent become more accurate.}
    \label{fig:app_example_carla_plan}
\end{figure*} 

 \newcommand{\nuscplappwidth}[0]{.35\textwidth}
 \begin{figure*}[htp]
    \centering
    \begin{subfigure}{\nuscplappwidth}
    \includegraphics[width=\textwidth]{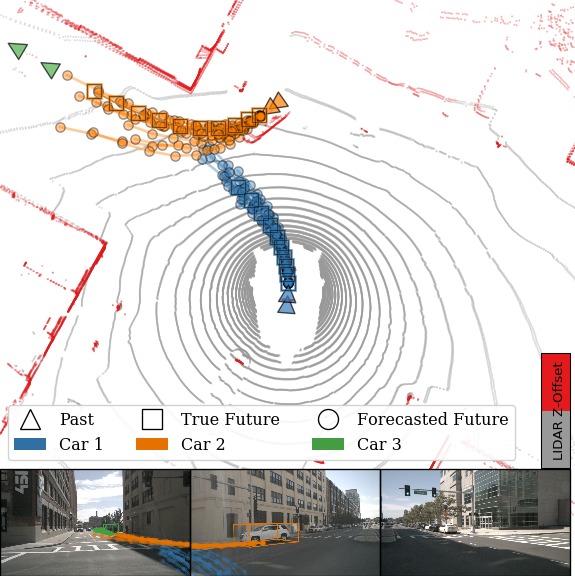}
    \caption{Scene 1, forecasted}
    \end{subfigure}
    \begin{subfigure}{\nuscplappwidth}
    \includegraphics[width=\textwidth]{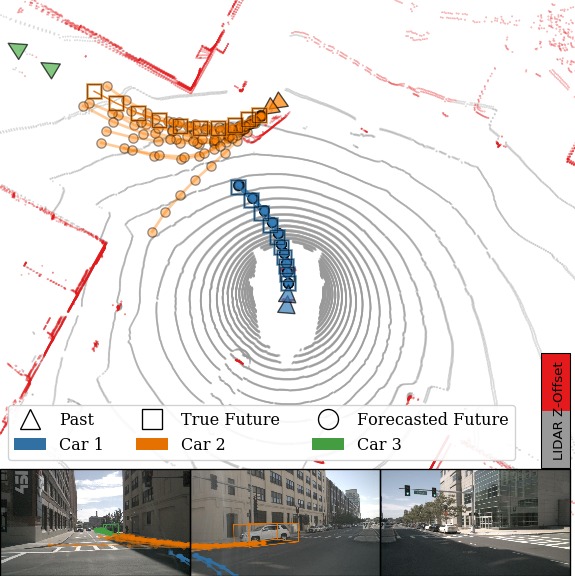}
     \caption{Scene 1, planned}
    \end{subfigure}
        \begin{subfigure}{\nuscplappwidth}
    \includegraphics[width=\textwidth]{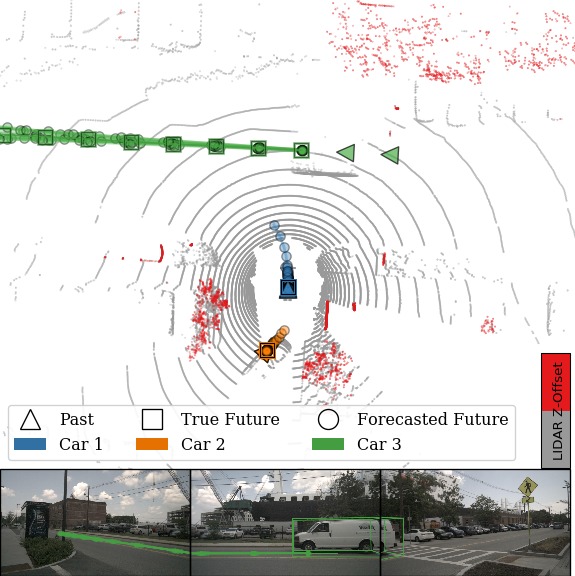}
      \caption{Scene 2, forecasted}
    \end{subfigure}
    \begin{subfigure}{\nuscplappwidth}
    \includegraphics[width=\textwidth]{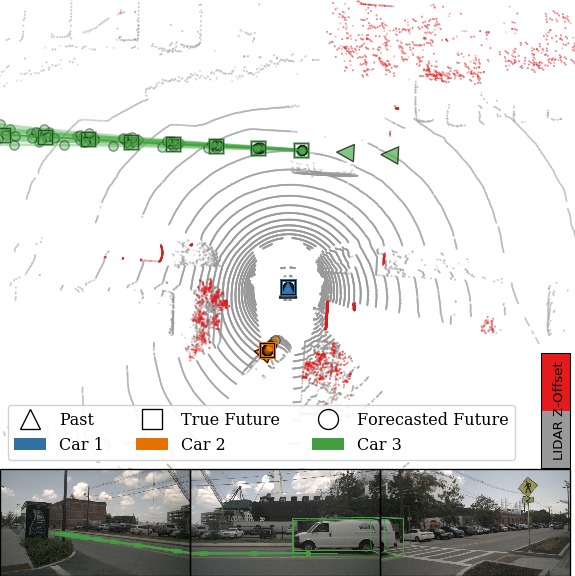}
         \caption{Scene 2, planned}
    \end{subfigure}
            \begin{subfigure}{\nuscplappwidth}
    \includegraphics[width=\textwidth]{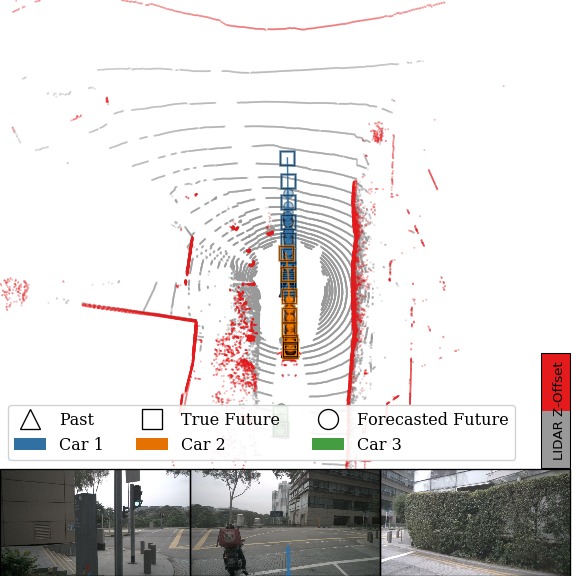}
      \caption{Scene 3, forecasted}
    \end{subfigure}
    \begin{subfigure}{\nuscplappwidth}
    \includegraphics[width=\textwidth]{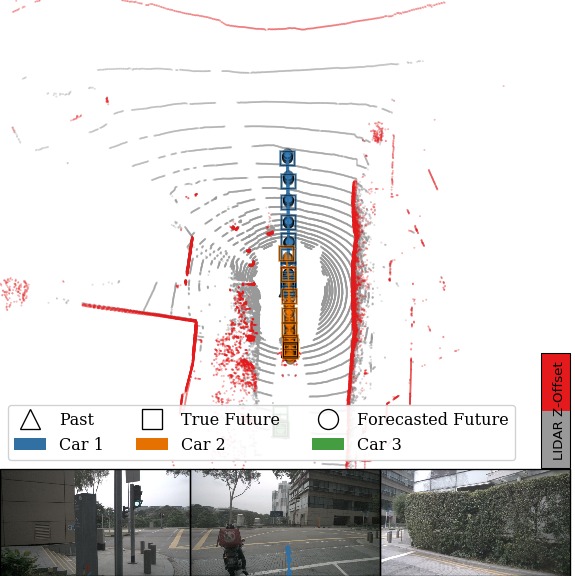}
         \caption{Scene 3, planned}
    \end{subfigure}
    \caption{Additional examples of \emph{planned} multi-agent forecasting (PRECOG) with our learned model in nuScenes. By using our planning approach and conditioning the robot on its true final position, our predictions for the robot become more accurate, and often our predictions of the other agent become more accurate.}
    \label{fig:app_example_nuscenes_plan}
\end{figure*}

\newcommand{\bestworst}[0]{.32\textwidth}
\begin{figure*}[htp]
    \centering
    \FBox{     \begin{subfigure}[t]{\bestworst}
        \includegraphics[width=\textwidth]{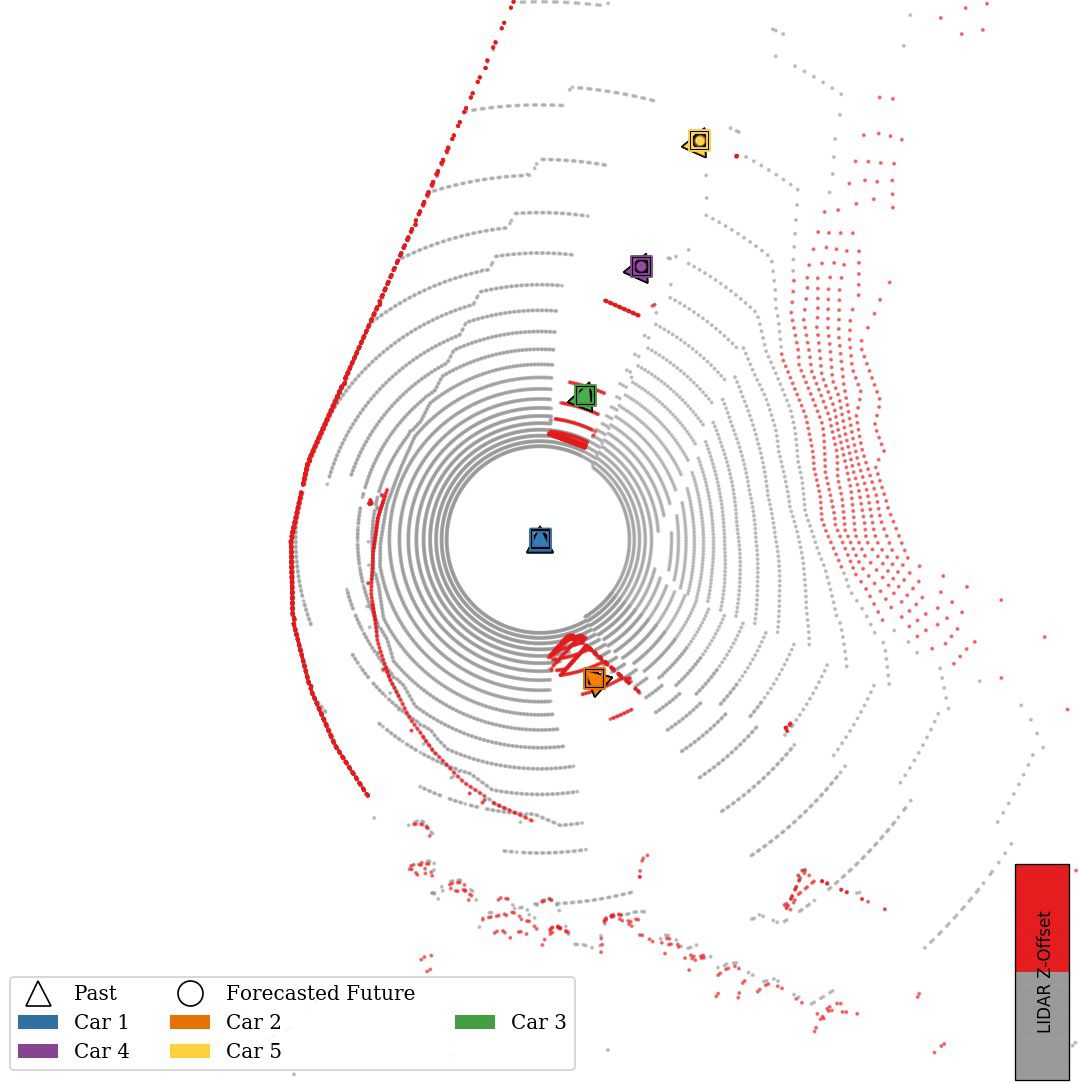}
            \caption{{\bf Best} ($>99\%$). $\hat m_{K=12}=\num[round-precision=1,scientific-notation=true]{0.00038554936}$}
    \end{subfigure} }
    \FBox{     \begin{subfigure}[t]{\bestworst}
    \includegraphics[width=\textwidth]{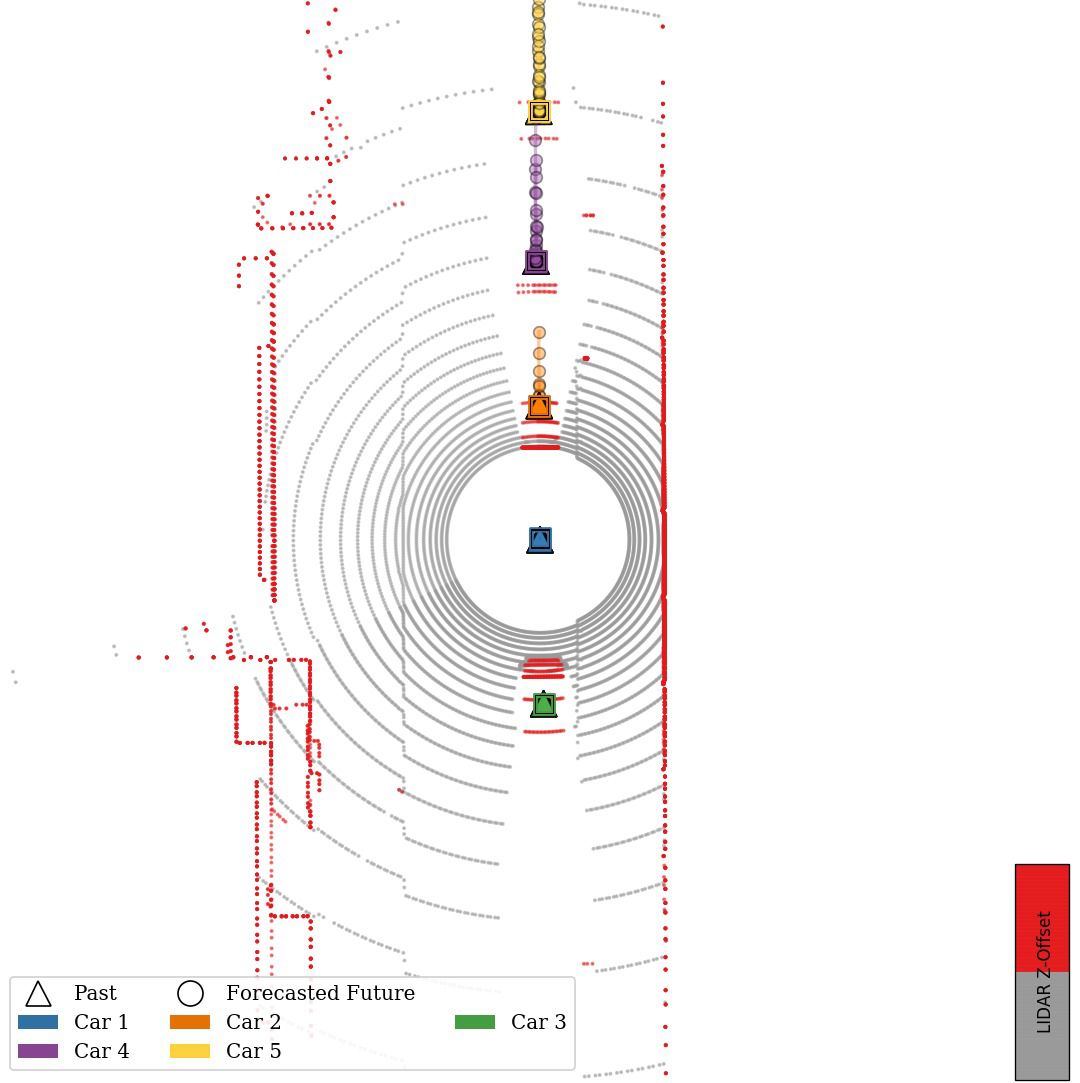}
    \caption{{\bf Best} ($>99\%$). $\hat m_{K=12}=\num[round-precision=1,scientific-notation=true]{0.00040327205}$}
    \end{subfigure}     }
        \FBox{     \begin{subfigure}[t]{\bestworst}
    \includegraphics[width=\textwidth]{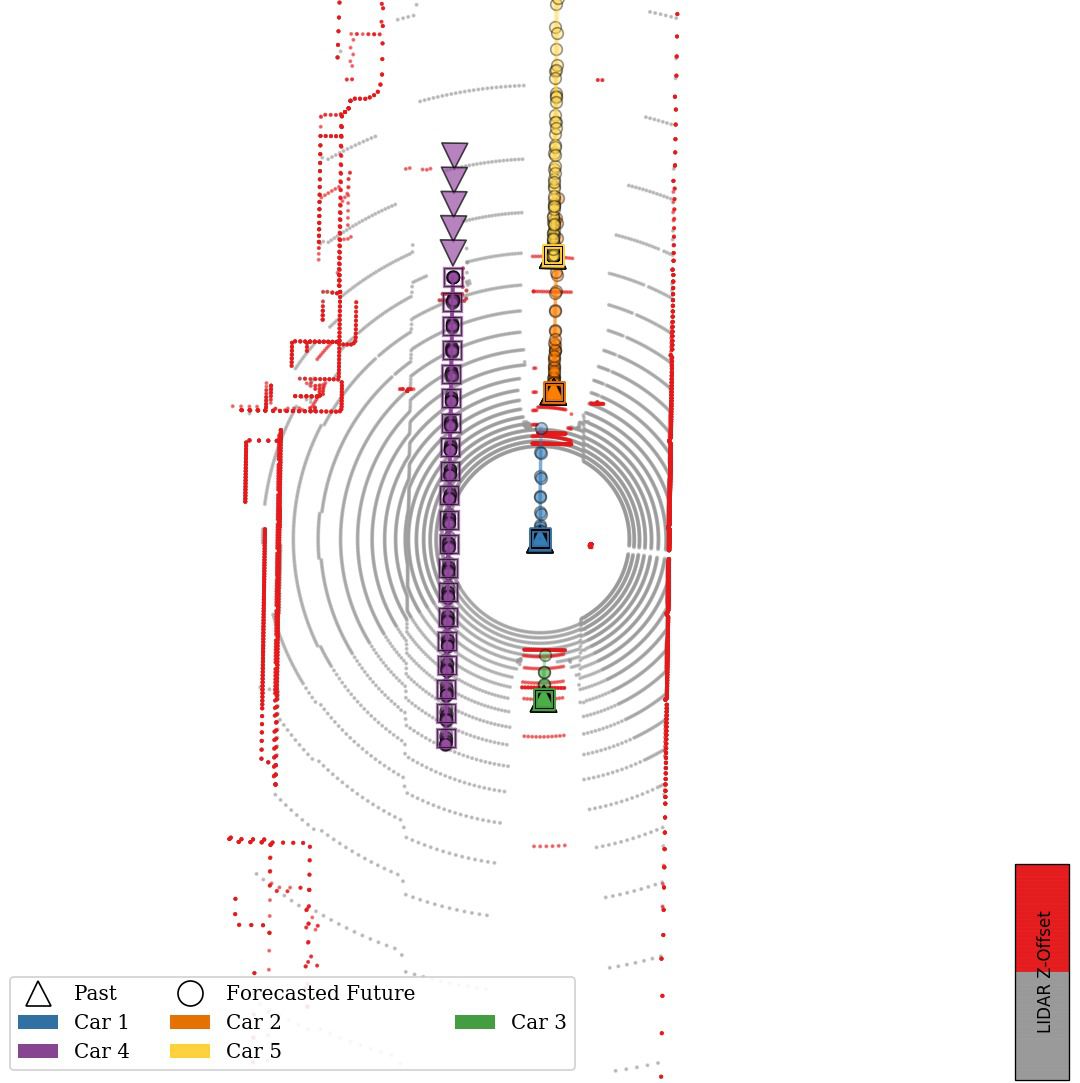}
    \caption{{\bf Best} ($>99\%$). $\hat m_{K=12}=\num[round-precision=1,round-precision=1,scientific-notation=true]{0.00054206065}$}
    \end{subfigure}     }

    \FBox{     \begin{subfigure}[t]{\bestworst}
    \includegraphics[width=\textwidth]{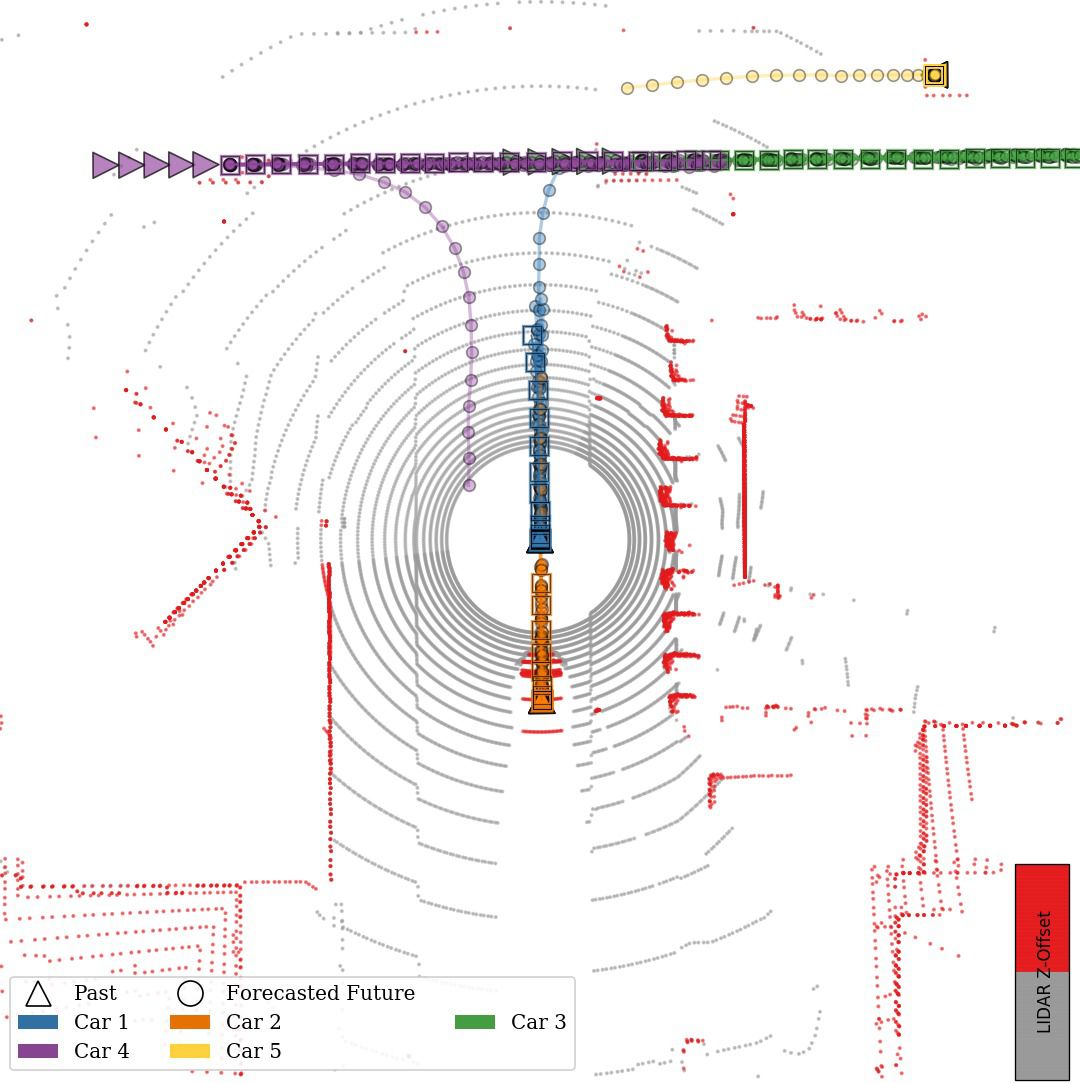}
    \caption{{\bf Median} ($\approx 50\%$). $\hat m_{K=12}=\num[round-precision=2]{0.3816643}$}
    \end{subfigure}     }
        \FBox{     \begin{subfigure}[t]{\bestworst}
    \includegraphics[width=\textwidth]{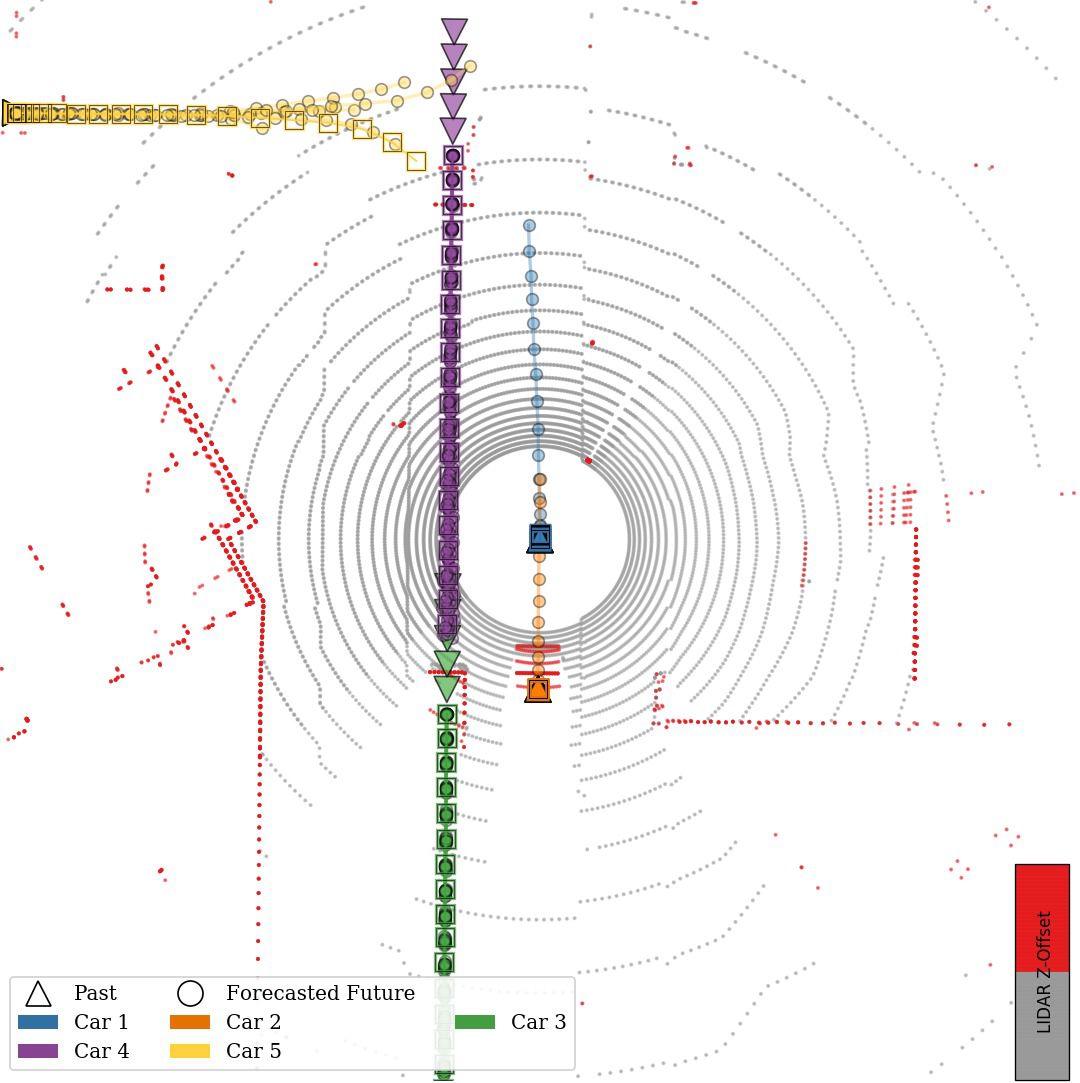}
        \caption{{\bf Median} ($\approx 50\%$). $\hat m_{K=12}=\num[round-precision=2]{0.37981653}$}
    \end{subfigure}     }
        \FBox{     \begin{subfigure}[t]{\bestworst}
    \includegraphics[width=\textwidth]{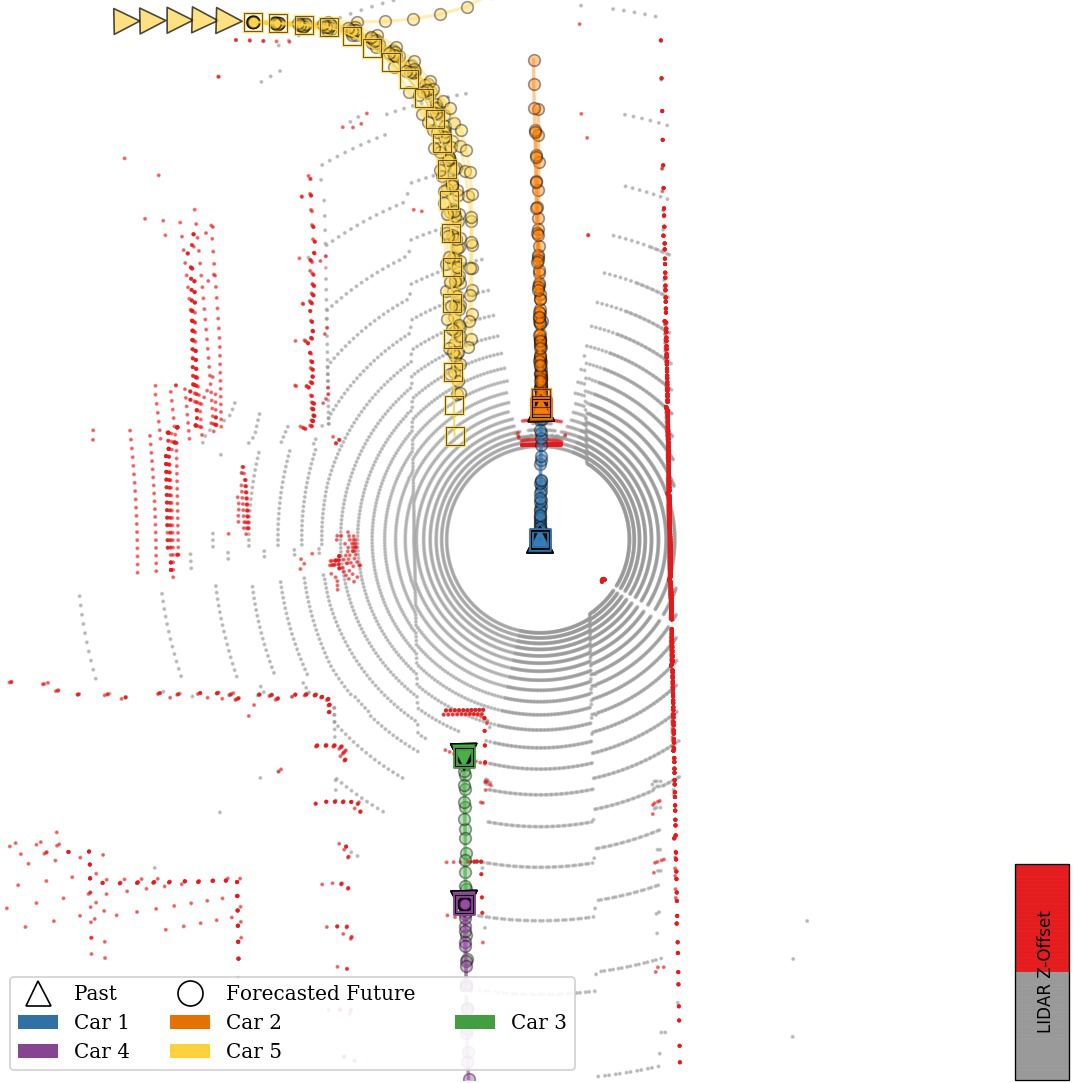}
    \caption{{\bf Median} ($\approx 50\%$). $\hat m_{K=12}=\num[round-precision=2]{0.3687198}$}
    \end{subfigure}     }
        \FBox{ 
    \begin{subfigure}[t]{\bestworst}
       \includegraphics[width=\textwidth,clip]{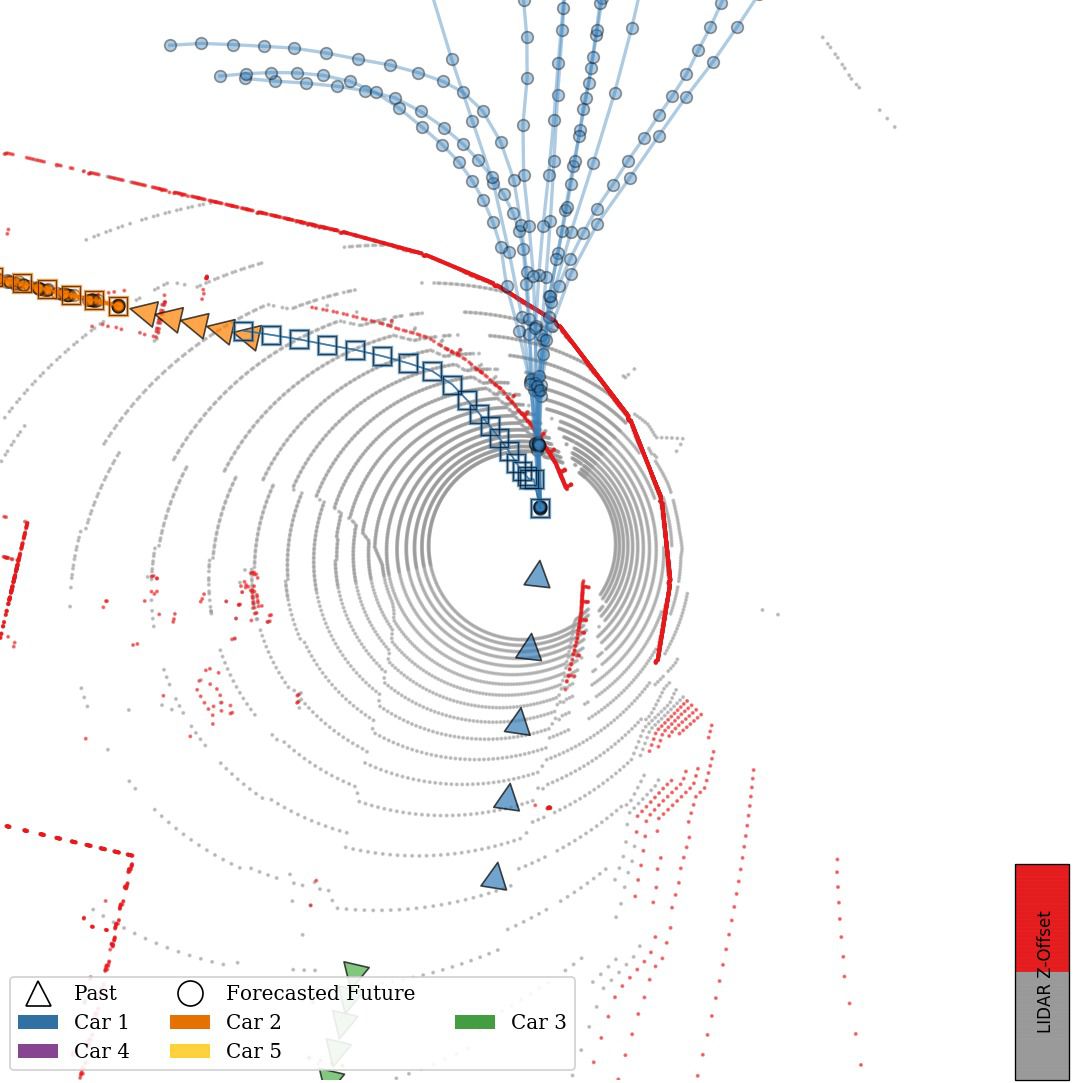}
       \caption{{\bf Worst} ($\!<\!0.2\%$). $\hat m_{K=12}\!=\!\num[round-precision=1]{64.738914}$}
       \end{subfigure}
    }
        \FBox{ 
    \begin{subfigure}[t]{\bestworst}
       \includegraphics[width=\textwidth,clip]{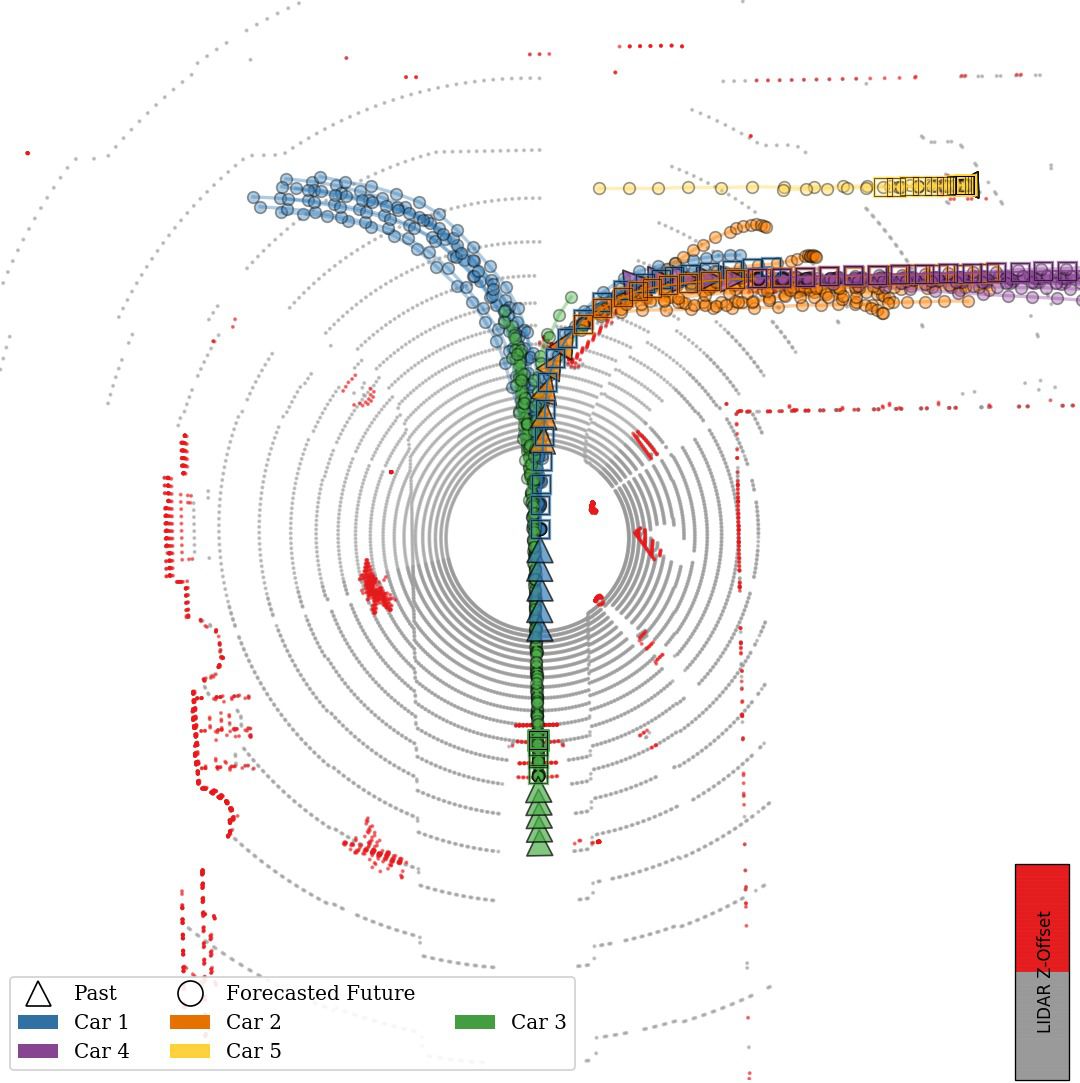}
            \caption{{\bf Worst} ($<0.2\%$). $\hat m_{K=12}=\num[round-precision=1]{41.256767}$}
       \end{subfigure}
    }
    \FBox{     \begin{subfigure}[t]{\bestworst}
        \includegraphics[width=\textwidth]{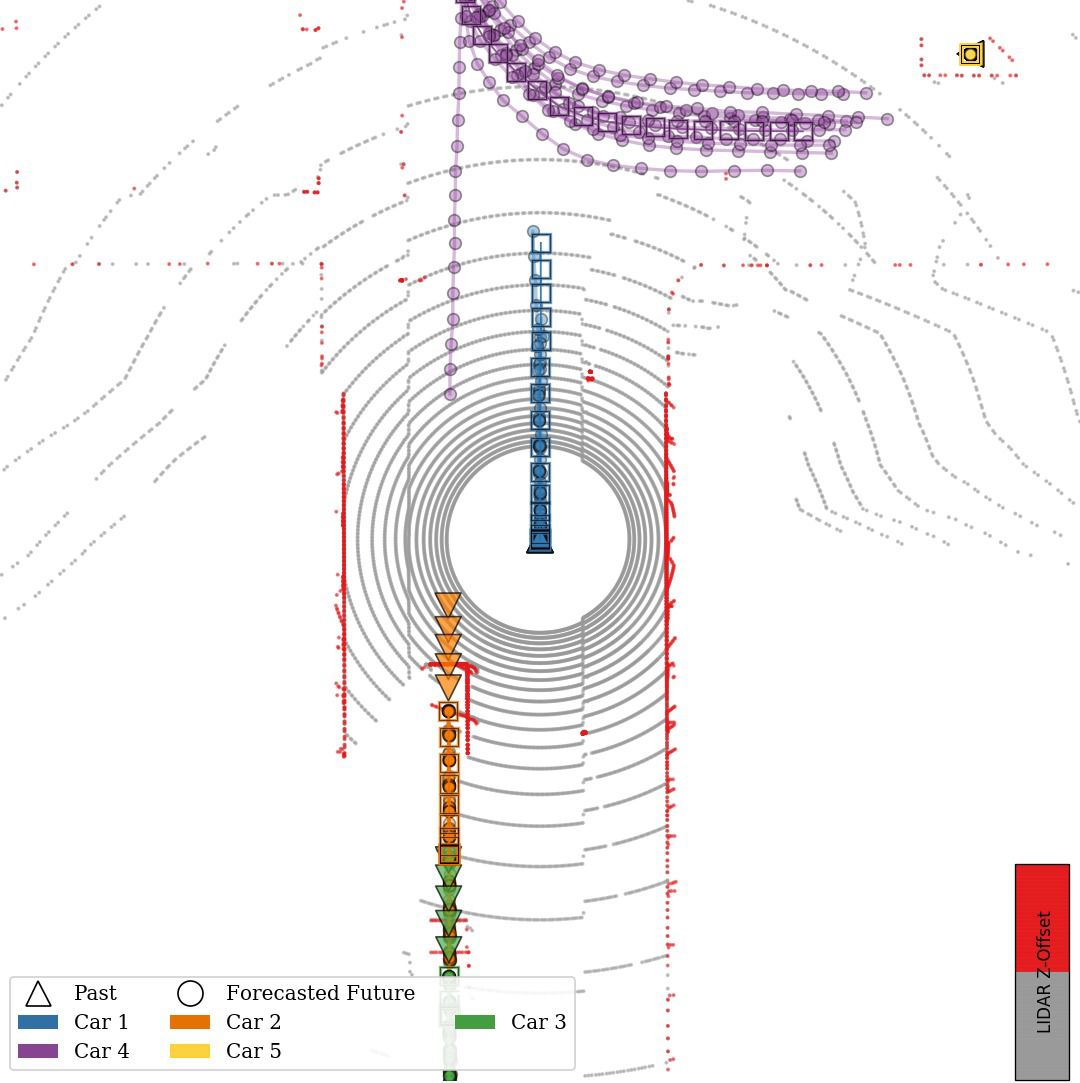}
            \caption{{\bf Worst} ($<0.2\%$). $\hat m_{K=12}=\num[round-precision=1]{41.18337}$}
    \end{subfigure} } 
    \caption{
    Various qualities (\emph{Row 1: } $\approx 100\%$, \emph{Row 2:} $\approx 50\%$, and \emph{Row 3:} $\approx 0\%$) of qualitative results of the ESP flex. count model on {\tt Town01 Test}, $A=5$, $T=40$ at $10$Hz (4 seconds of future), ordered by $\hat m_{K=12}$. Recall since $\hat m$ is a \emph{joint-agent} statistic, per-agent trajectory sample coverage is insufficient for a good $\hat m$ score. Also, recall $\hat m$ measures the error of the \emph{closest} joint trajectory to the true future, as opposed to the error of \emph{all} joint trajectories, which is key to its property of not penalizing otherwise-plausible trajectories.
    } 
    \label{fig:best_middle_worst_carla}
\end{figure*} 

\begin{figure*}[htp]
    \centering
        \FBox{ 
    \begin{subfigure}[t]{\bestworst}
       \includegraphics[width=\textwidth,clip]{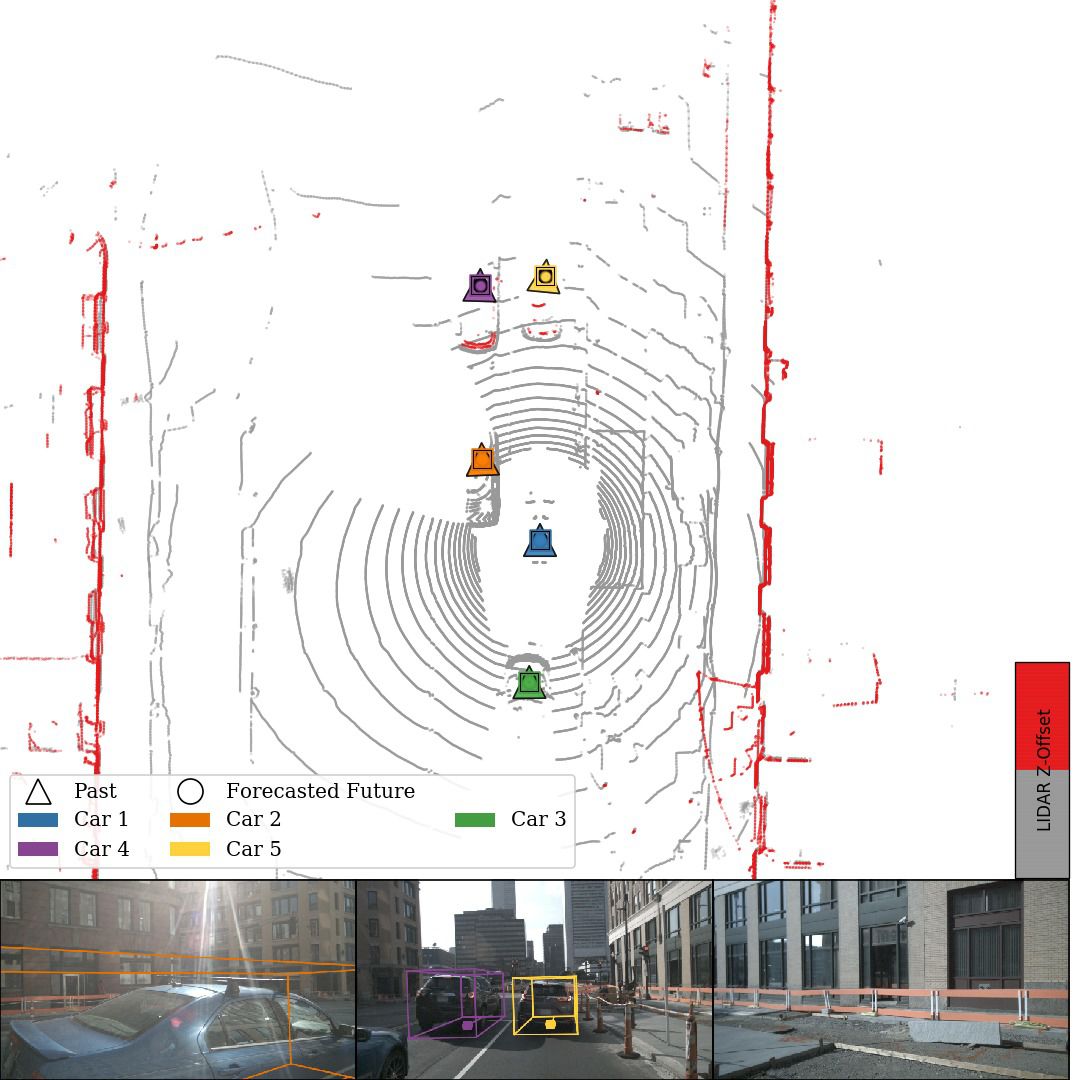}
       \caption{{\bf Best} ($\approx 99\%$). $\hat m_{K=12}\!=\!\num[round-precision=1,scientific-notation=true]{0.00031447396}$}
       \end{subfigure}
    }
        \FBox{ 
    \begin{subfigure}[t]{\bestworst}
       \includegraphics[width=\textwidth,clip]{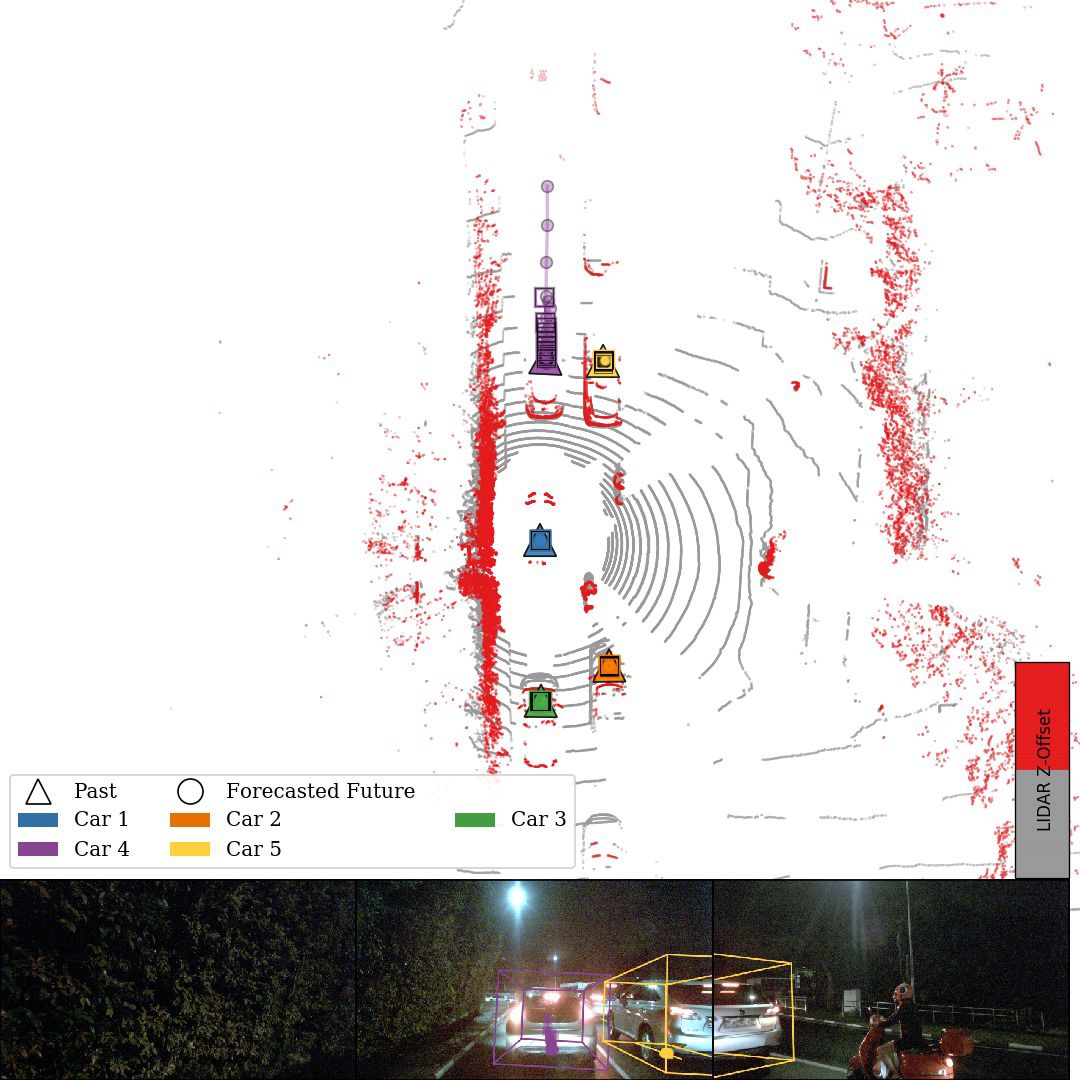}
       \caption{{\bf Best} ($\approx 94\%$). $\hat m_{K=12}\!=\!\num[round-precision=1,scientific-notation=true]{ 0.023256572}$}
       \end{subfigure}
    }
        \FBox{ 
    \begin{subfigure}[t]{\bestworst}
       \includegraphics[width=\textwidth,clip]{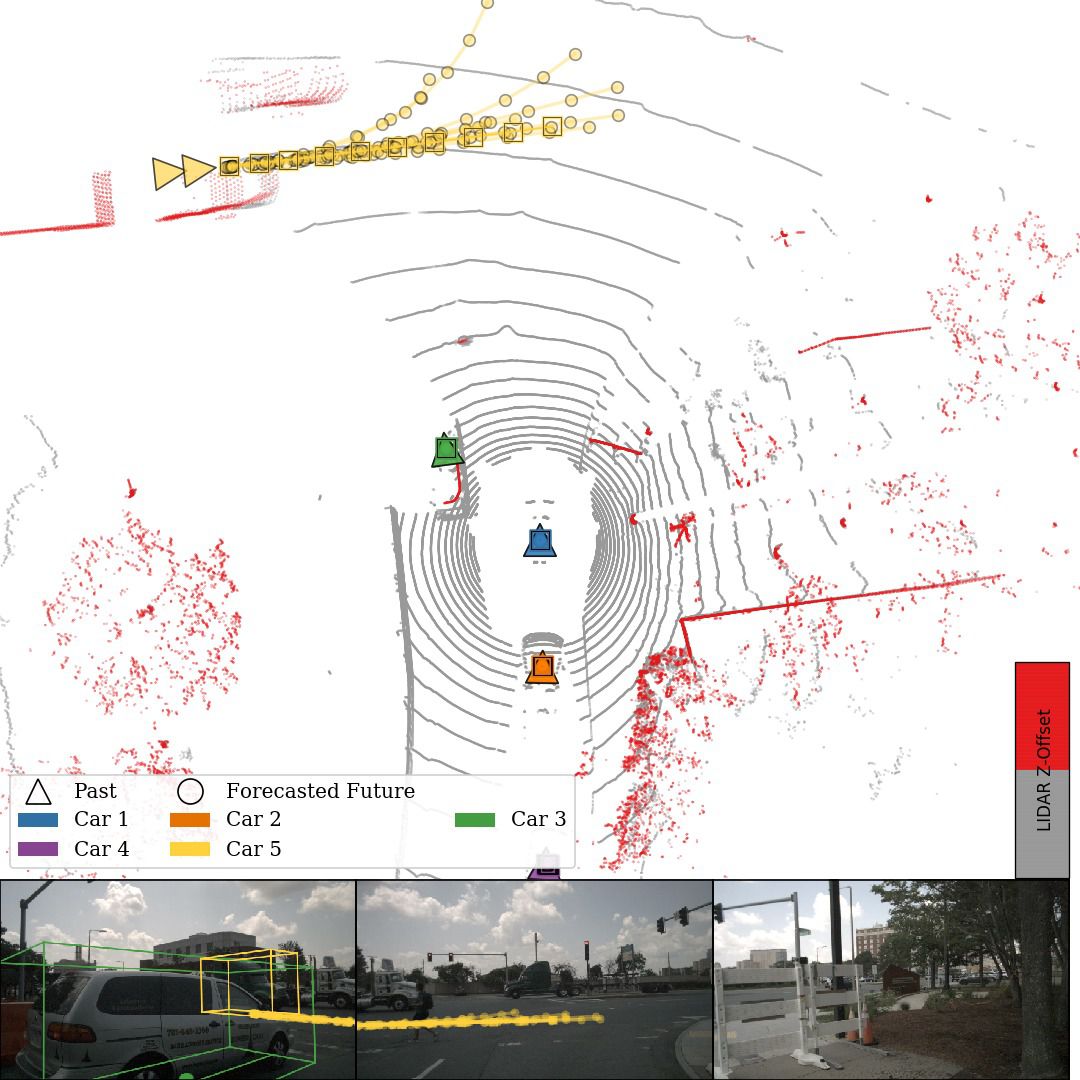}
       \caption{{\bf Best} ($\approx 93\%$). $\hat m_{K=12}\!=\!\num[round-precision=1,scientific-notation=true]{0.028836906}$}
       \end{subfigure}
    }
        \FBox{ 
    \begin{subfigure}[t]{\bestworst}
       \includegraphics[width=\textwidth,clip]{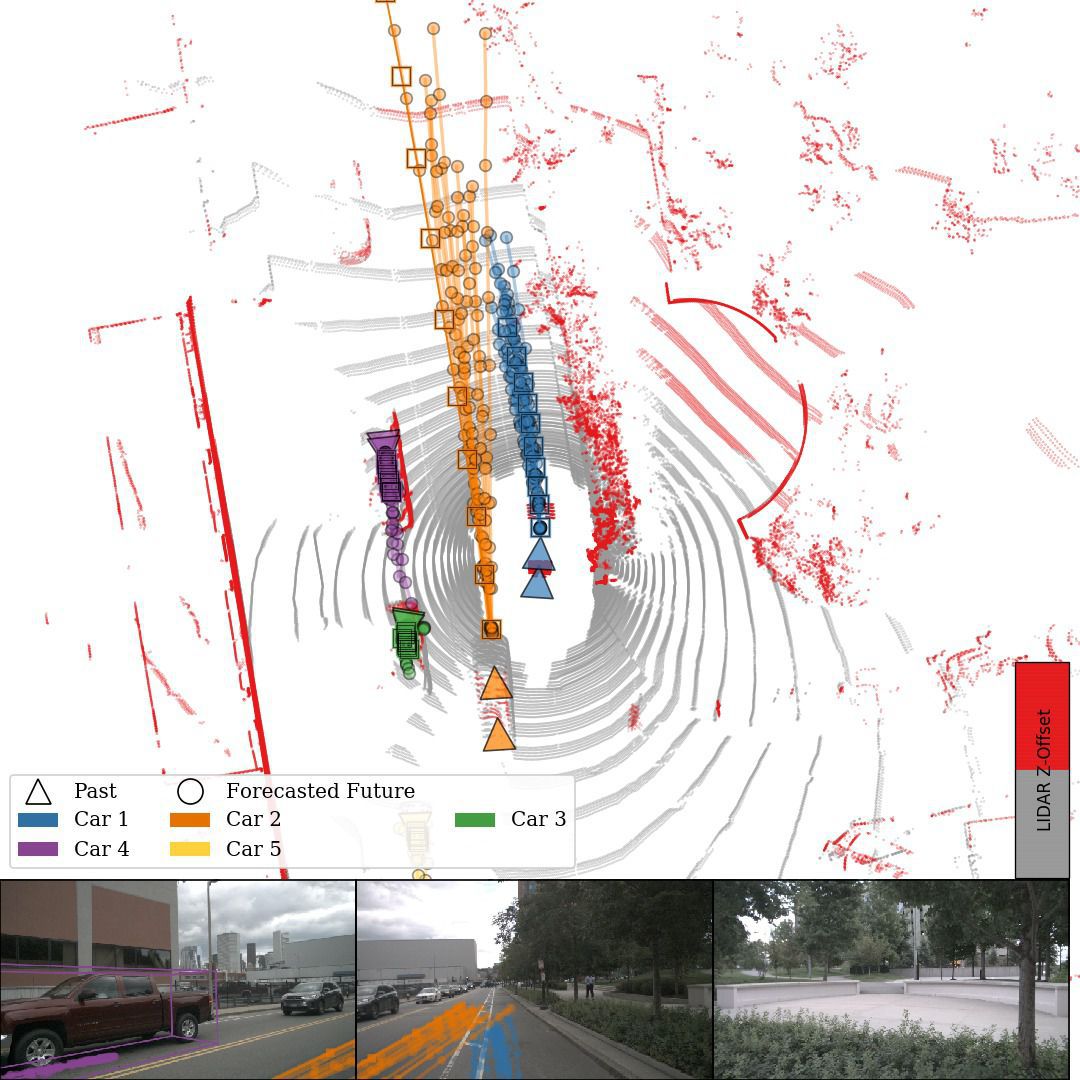}
       \caption{{\bf Median} ($\approx 50\%$). $\hat m_{K=12}\!=\!\num[round-precision=1]{1.3064103}$}
       \end{subfigure}
    }
        \FBox{ 
    \begin{subfigure}[t]{\bestworst}
       \includegraphics[width=\textwidth,clip]{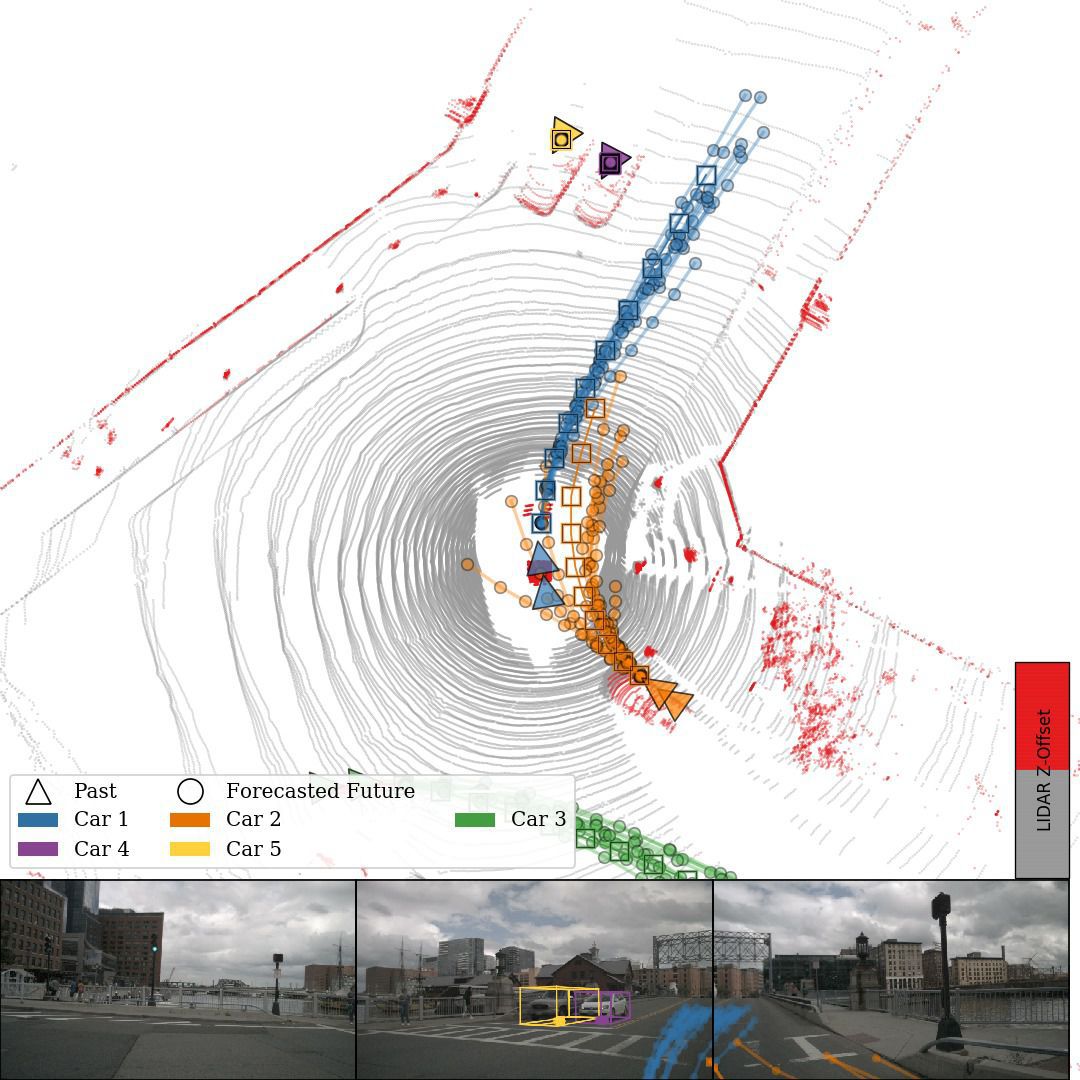}
       \caption{{\bf Median} ($\approx 50\%$). $\hat m_{K=12}\!=\!\num[round-precision=1]{1.3085703}$}
       \end{subfigure}
    }
        \FBox{ 
    \begin{subfigure}[t]{\bestworst}
       \includegraphics[width=\textwidth,clip]{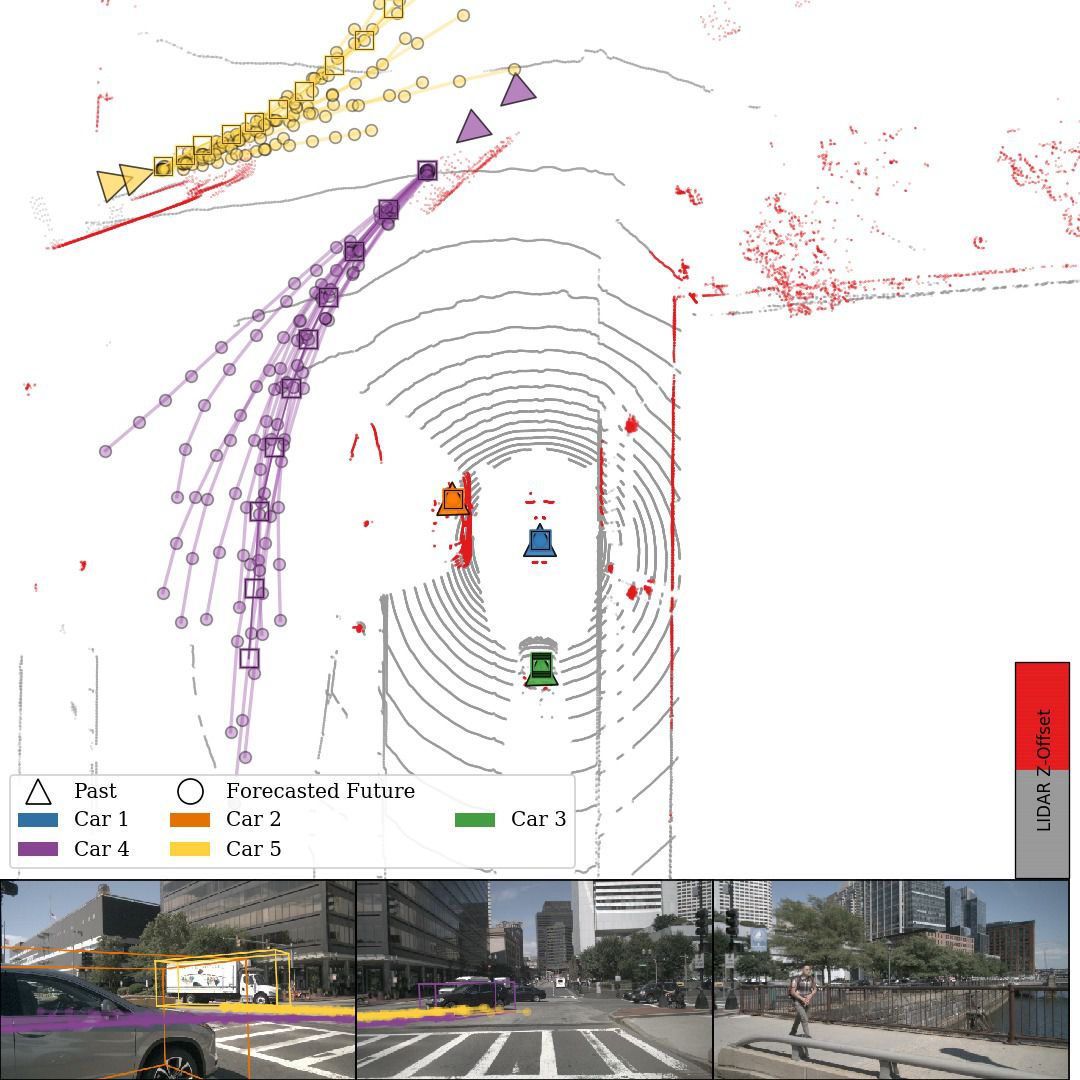}
       \caption{{\bf Median} ($\approx 50\%$). $\hat m_{K=12}\!=\!\num[round-precision=1]{1.3270347}$}
       \end{subfigure}
    }
            \FBox{ 
    \begin{subfigure}[t]{\bestworst}
       \includegraphics[width=\textwidth,clip]{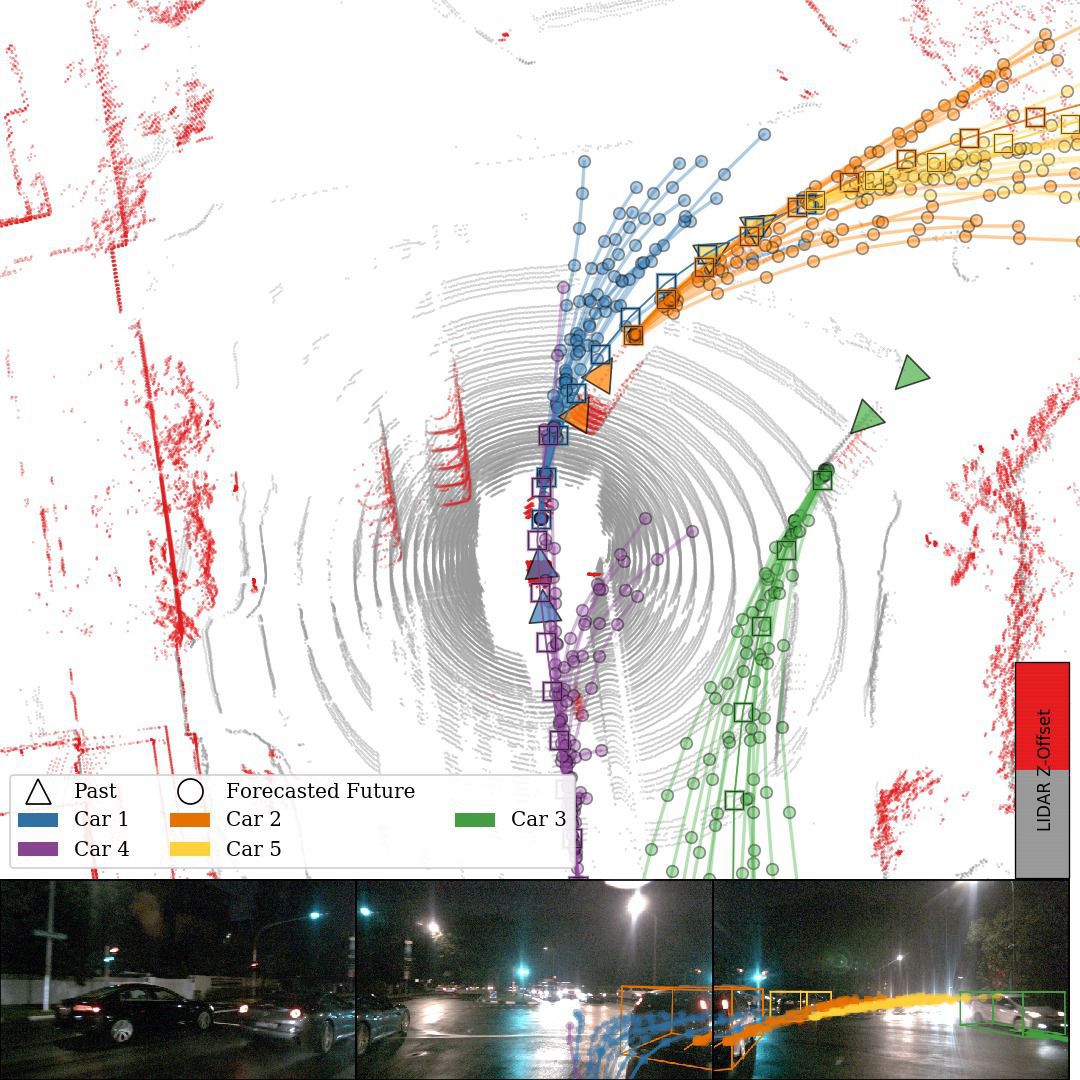}
       \caption{{\bf Worst} ($\!<\!3\%$). $\hat m_{K=12}\!=\!\num[round-precision=1]{15.592783}$}
       \end{subfigure}
    }
            \FBox{ 
    \begin{subfigure}[t]{\bestworst}
       \includegraphics[width=\textwidth,clip]{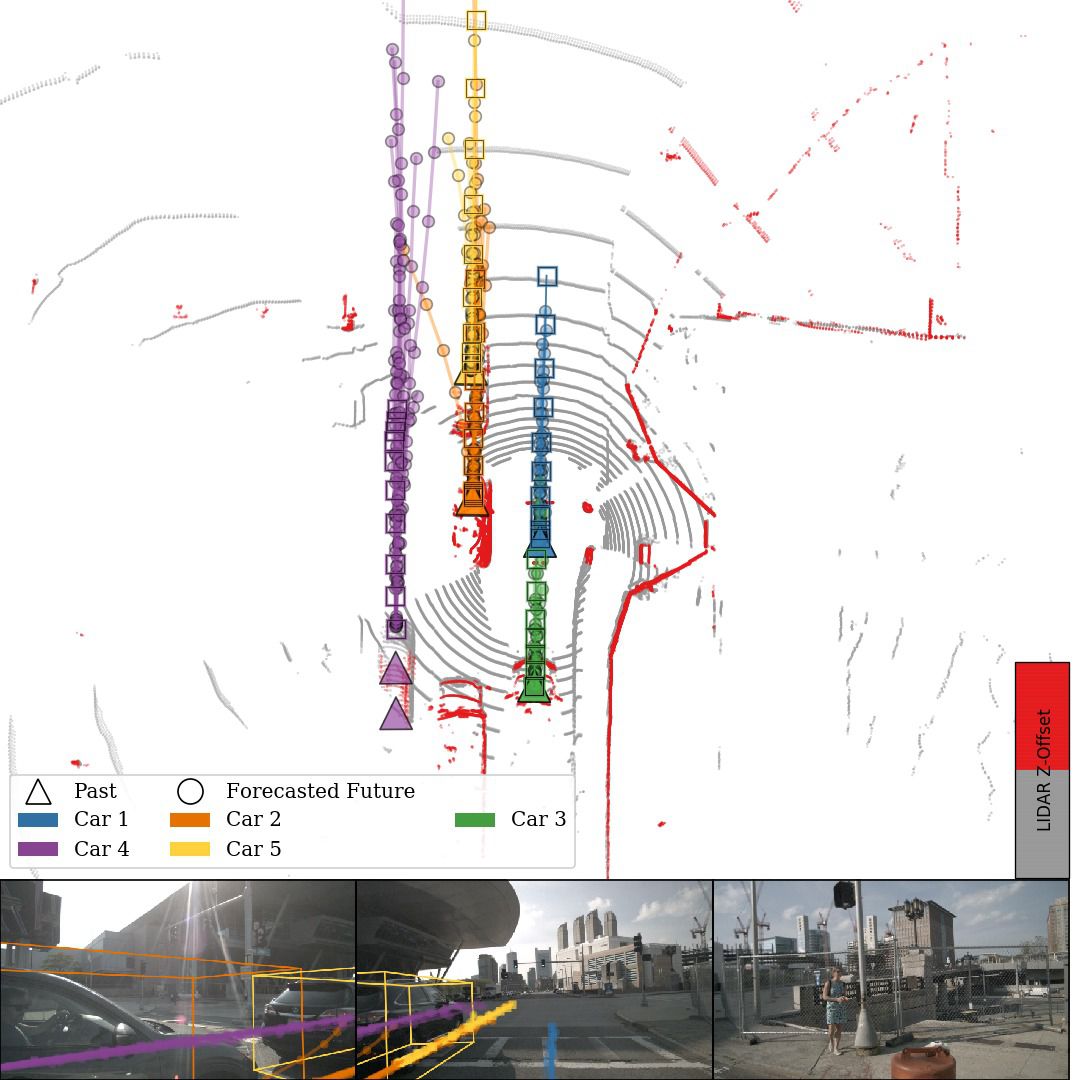}
       \caption{{\bf Worst} ($\!<\!3\%$). $\hat m_{K=12}\!=\!\num[round-precision=1]{13.391655}$}
       \end{subfigure}
    }
            \FBox{ 
    \begin{subfigure}[t]{\bestworst}
       \includegraphics[width=\textwidth,clip]{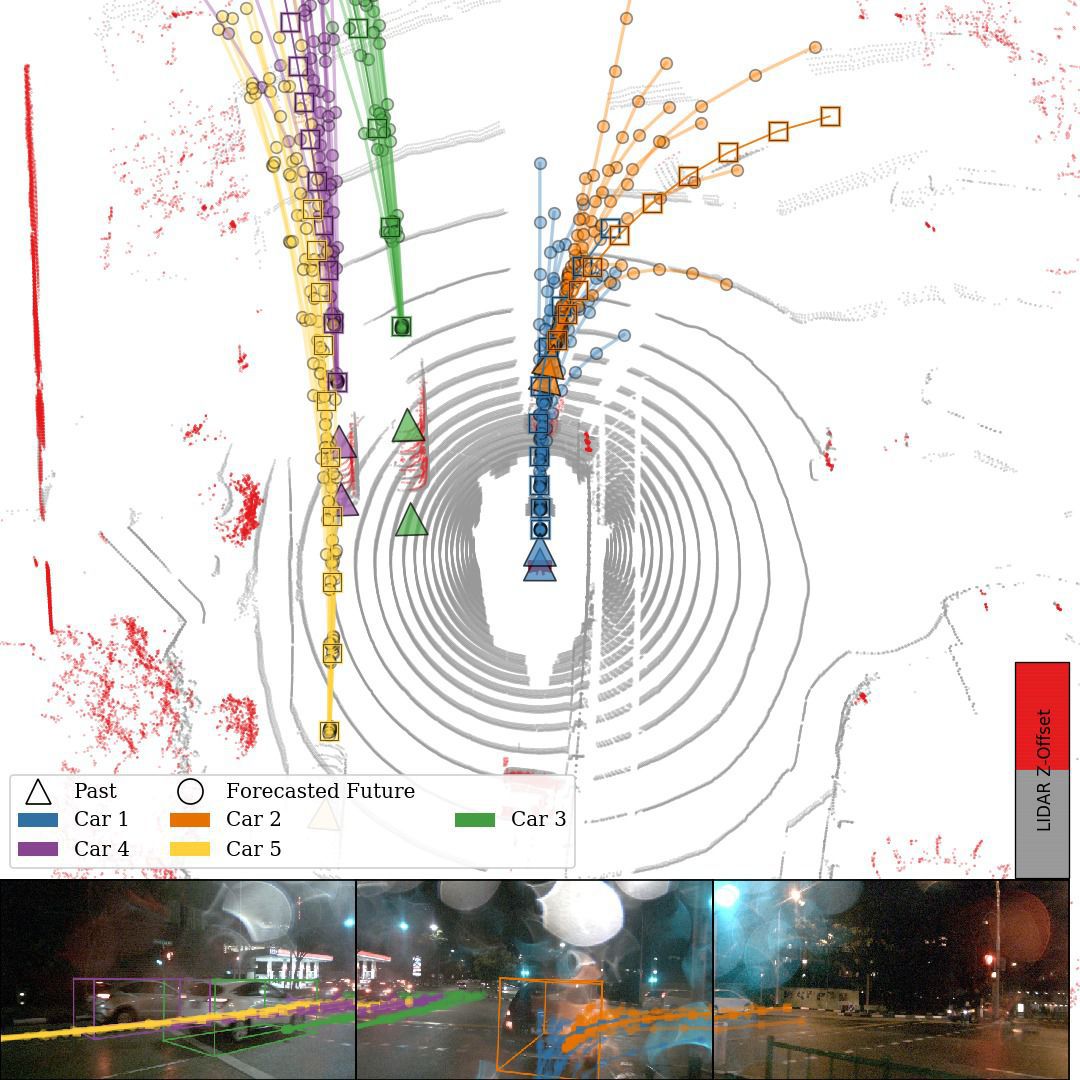}
       \caption{{\bf Worst} ($\!<\!3\%$). $\hat m_{K=12}\!=\!\num[round-precision=1]{11.520764}$}
       \end{subfigure}
    }
    \caption{
    Various qualities (\emph{Row 1: } $\approx 100\%$, \emph{Row 2:} $\approx 50\%$, and \emph{Row 3:} $\approx 0\%$) of qualitative results of the ESP flex. count model on {nuScenes Test}, $A=5$, $T=20$ at $5$Hz (4 seconds of future), ordered by $\hat m_{K=12}$. Recall since $\hat m$ is a \emph{joint-agent} statistic, per-agent trajectory sample coverage is insufficient for a good $\hat m$ score. Also, recall $\hat m$ measures the error of the \emph{closest} joint trajectory to the true future, as opposed to the error of \emph{all} joint trajectories, which is key to its property of not penalizing otherwise-plausible trajectories.
    } 
    \label{fig:best_middle_worst_nuscenes}
\end{figure*} 

\newcommand{\postscanwidth}[0]{.29\textwidth}
\begin{figure*}[htp]
    \centering
    \FBox{ 
    \begin{subfigure}[t]{\postscanwidth}
           \includegraphics[width=\textwidth,clip]{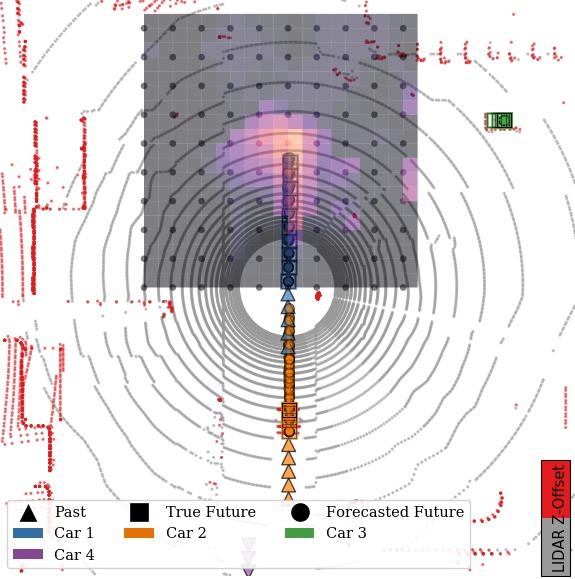}
       \end{subfigure}
    }
    \hfill
    \FBox{
        \begin{subfigure}[t]{\postscanwidth}
           \includegraphics[width=\textwidth,clip]{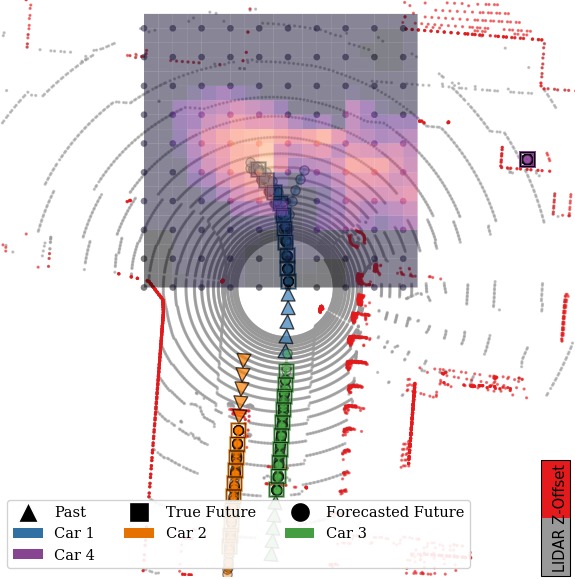}
       \end{subfigure}
    }
    \hfill
        \FBox{
        \begin{subfigure}[t]{\postscanwidth}
           \includegraphics[width=\textwidth,clip]{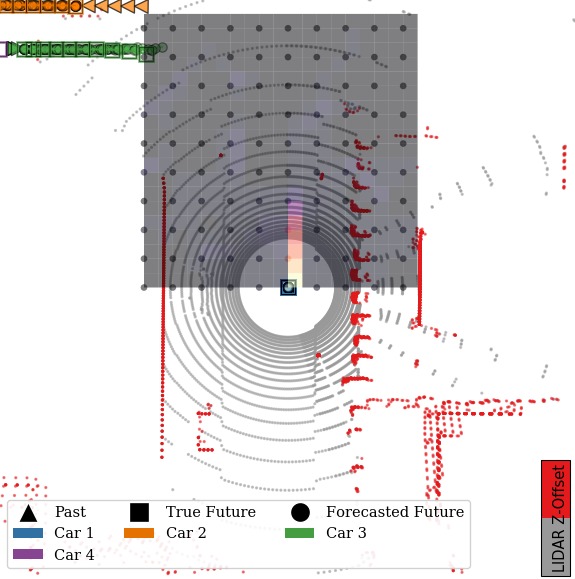}
       \end{subfigure}
    }
    
        \FBox{
        \begin{subfigure}[t]{\postscanwidth}
           \includegraphics[width=\textwidth,clip]{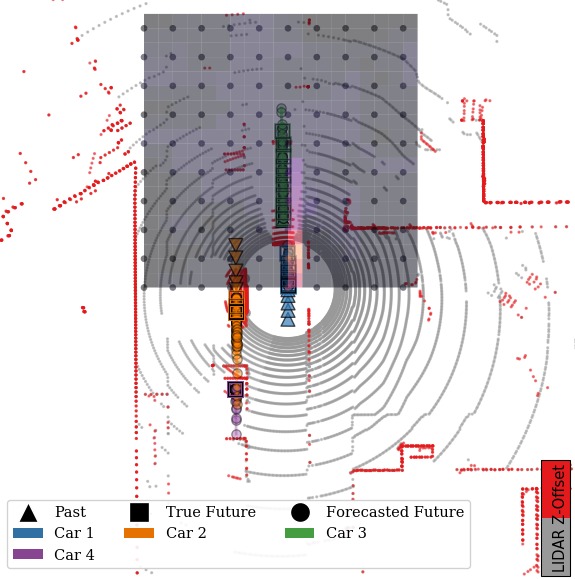}
       \end{subfigure}
    }
    \hfill
        \FBox{
        \begin{subfigure}[t]{\postscanwidth}
           \includegraphics[width=\textwidth,clip]{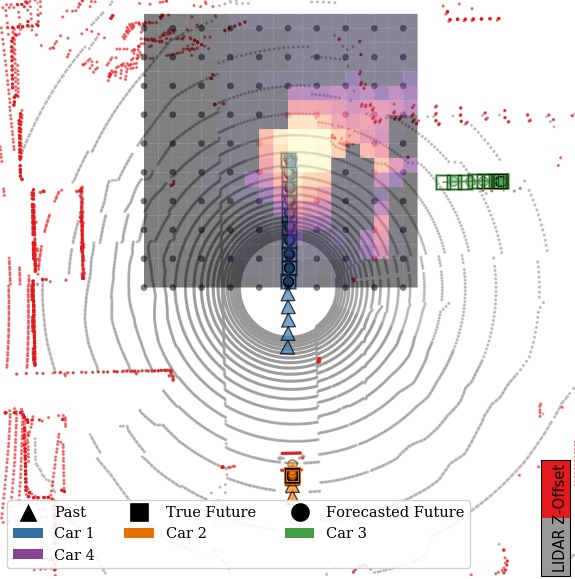}
       \end{subfigure}
    }
    \hfill
        \FBox{
        \begin{subfigure}[t]{\postscanwidth}
           \includegraphics[width=\textwidth,clip]{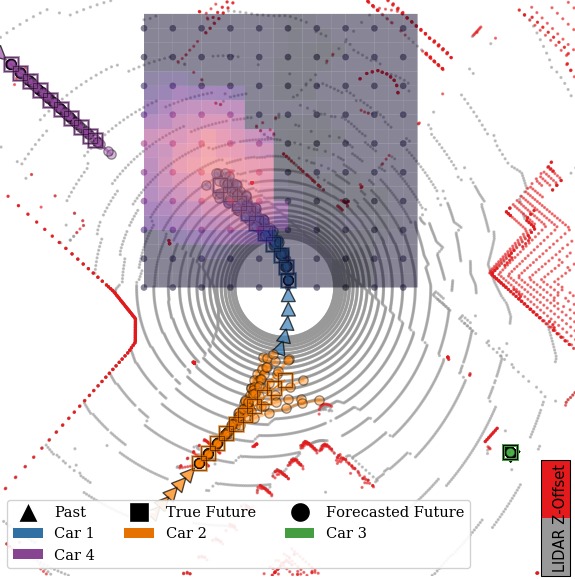}
       \end{subfigure}
    }
    
        \FBox{
        \begin{subfigure}[t]{\postscanwidth}
           \includegraphics[width=\textwidth,clip]{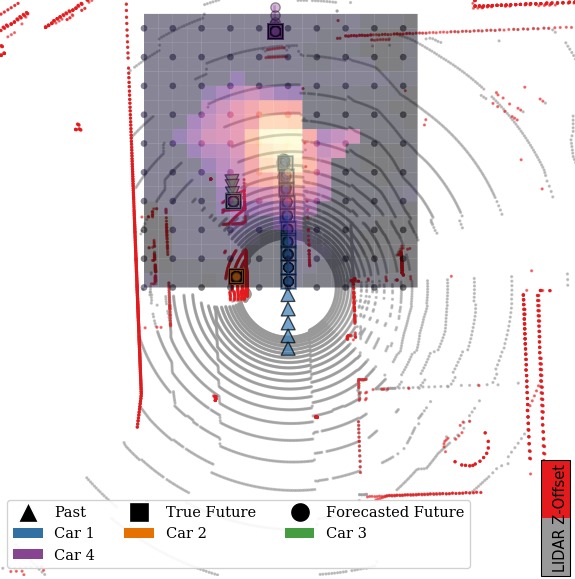}
       \end{subfigure}
    }
    \hfill
        \FBox{
        \begin{subfigure}[t]{\postscanwidth}
           \includegraphics[width=\textwidth,clip]{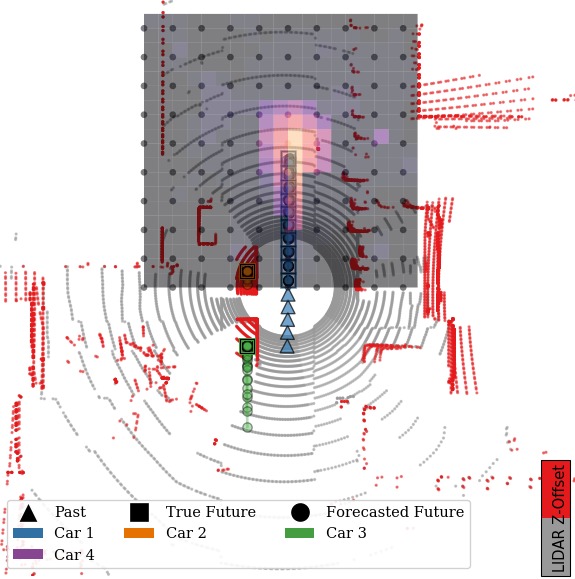}
       \end{subfigure}
    }
    \hfill
        \FBox{
        \begin{subfigure}[t]{\postscanwidth}
           \includegraphics[width=\textwidth,clip]{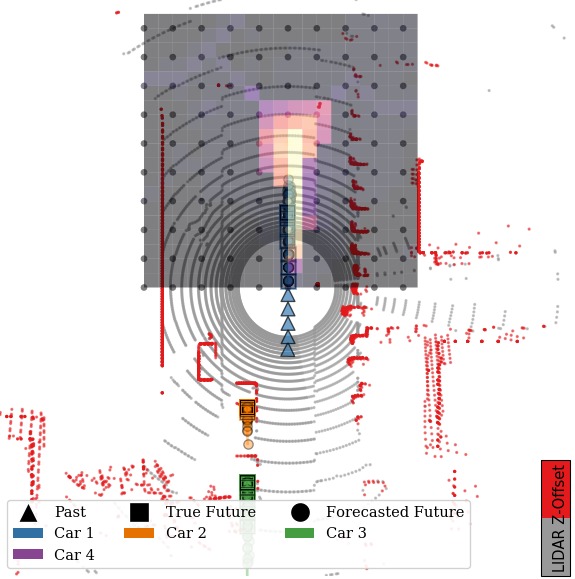}
       \end{subfigure}
    }
    
         \FBox{
        \begin{subfigure}[t]{\postscanwidth}
           \includegraphics[width=\textwidth,clip]{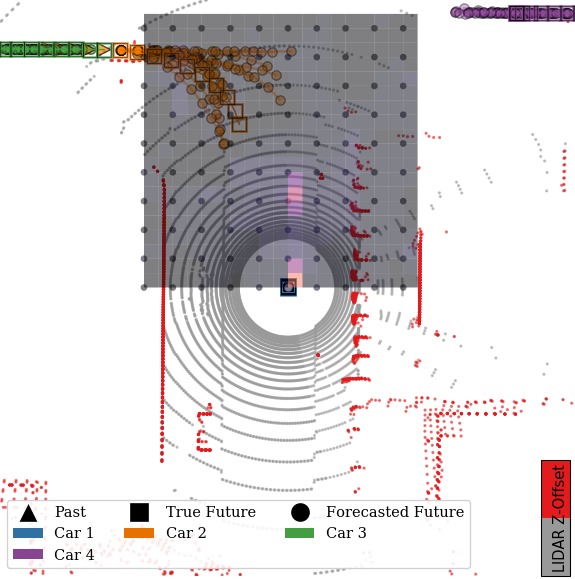}
       \end{subfigure}
    }
    \hfill
        \FBox{
        \begin{subfigure}[t]{\postscanwidth}
           \includegraphics[width=\textwidth,clip]{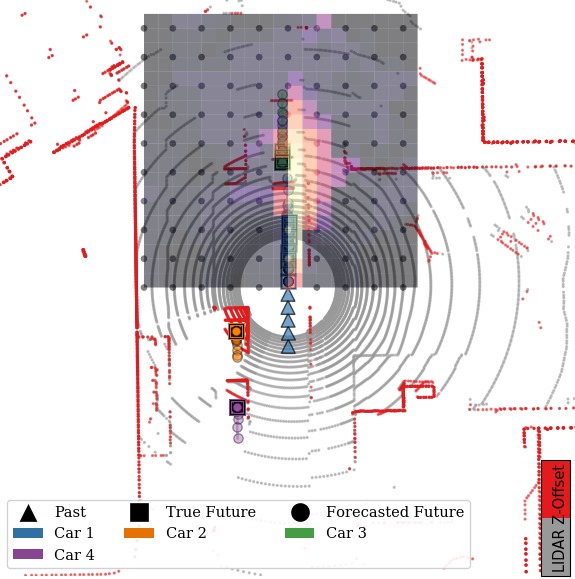}
       \end{subfigure}
    }
    \hfill
        \FBox{
        \begin{subfigure}[t]{\postscanwidth}
           \includegraphics[width=\textwidth,clip]{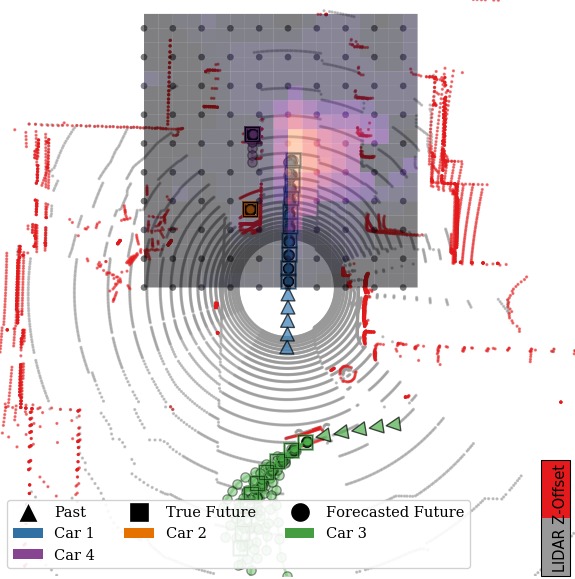}
       \end{subfigure}
    }
       \begin{flushleft}
    \begin{subfigure}[t]{\postscanwidth}
           \includegraphics[width=\textwidth,clip]{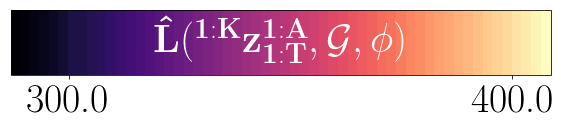}
       \end{subfigure}
       \end{flushleft}
     \vspace{-15pt}
    \caption{
    Plotting the planning criterion, $\hat L$, after planning to various positions (small circular dots in each plot) input to Alg.~\ref{alg:stateprecog}, with values interpolated between each position, in CARLA. The planning criterion input corresponds to a spatio-temporal goal at $T=20$ in the future ($4$ seconds). The planning criterion prefers locations within its lane, unless it is uncertain about the possibility of turning. When the vehicle was stationary in the past, the planning criterion is highest at positions at or close in front of the vehicle.      
    } 
    \label{fig:posterior_scan_carla}
\end{figure*}

\newcommand{\nuscpostscanwidth}[0]{.29\textwidth}
\begin{figure*}[htp]
    \centering
    \FBox{ 
    \begin{subfigure}[t]{\nuscpostscanwidth}
           \includegraphics[width=\textwidth,clip]{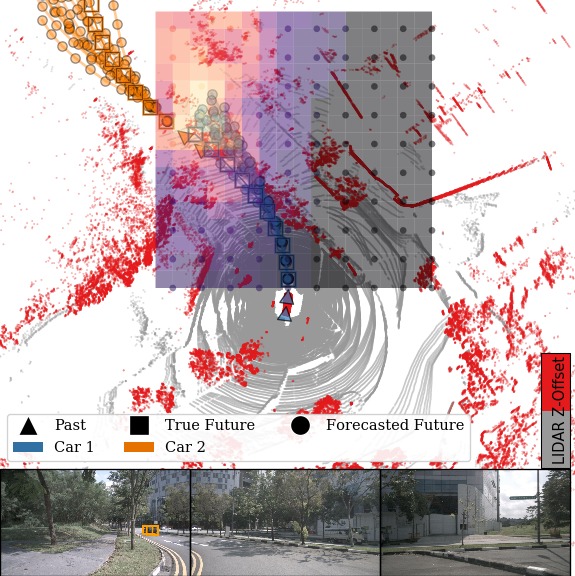}
       \end{subfigure}
    }
    \hfill
        \FBox{ 
    \begin{subfigure}[t]{\nuscpostscanwidth}
           \includegraphics[width=\textwidth,clip]{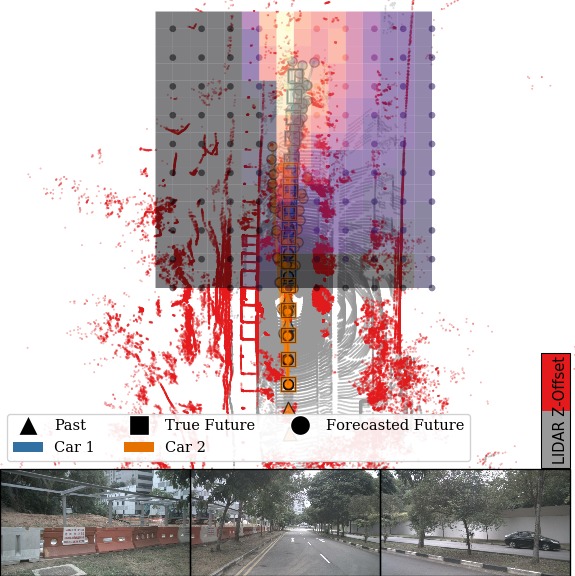}
       \end{subfigure}
    }
        \hfill
           \FBox{ 
    \begin{subfigure}[t]{\nuscpostscanwidth}
           \includegraphics[width=\textwidth,clip]{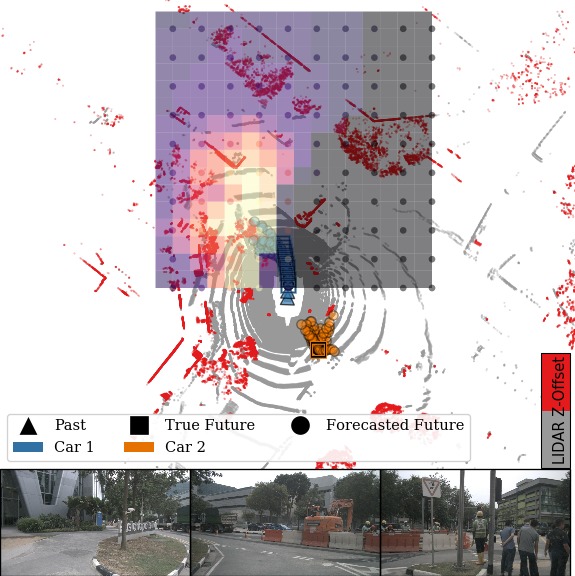}
       \end{subfigure}
    }
           \FBox{ 
    \begin{subfigure}[t]{\nuscpostscanwidth}
           \includegraphics[width=\textwidth,clip]{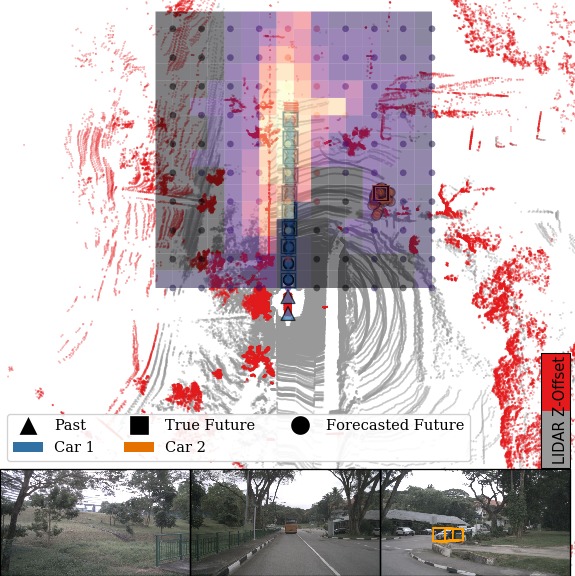}
       \end{subfigure}
    }
        \hfill
           \FBox{ 
    \begin{subfigure}[t]{\nuscpostscanwidth}
           \includegraphics[width=\textwidth,clip]{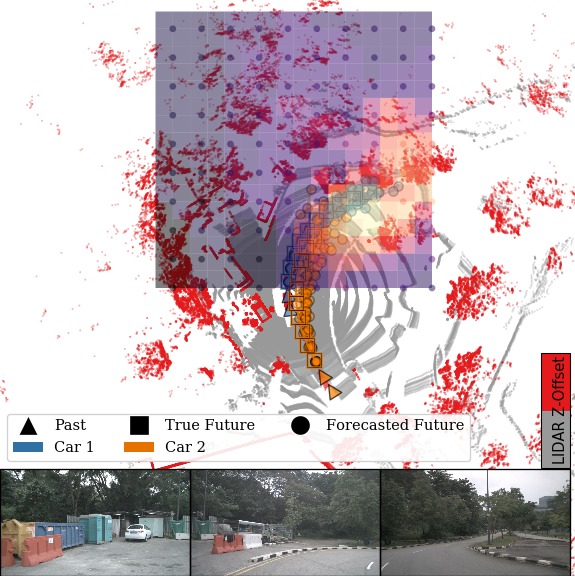}
       \end{subfigure}
    }
        \hfill
           \FBox{ 
    \begin{subfigure}[t]{\nuscpostscanwidth}
           \includegraphics[width=\textwidth,clip]{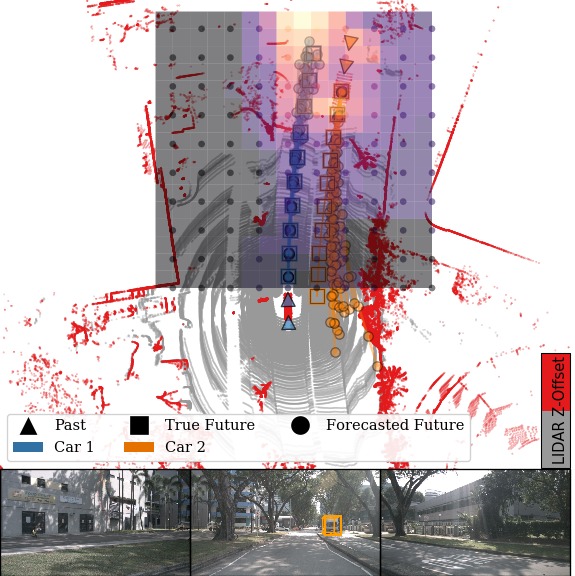}
       \end{subfigure}
    }
    
    \FBox{ 
    \begin{subfigure}[t]{\nuscpostscanwidth}
           \includegraphics[width=\textwidth,clip]{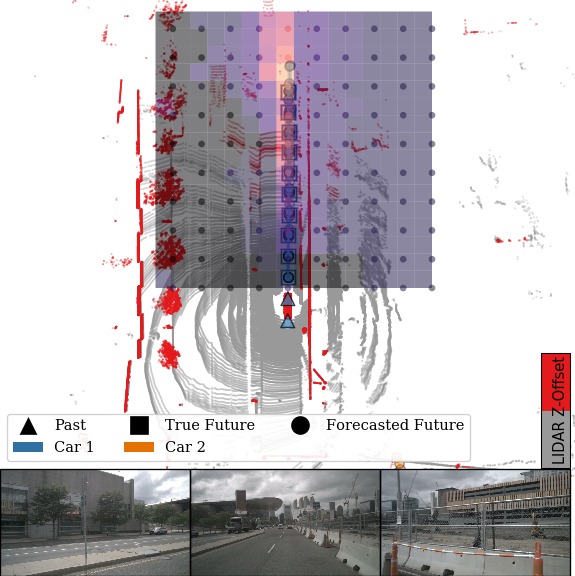}
       \end{subfigure}
    }
    \hfill
    \FBox{ 
    \begin{subfigure}[t]{\nuscpostscanwidth}
           \includegraphics[width=\textwidth,clip]{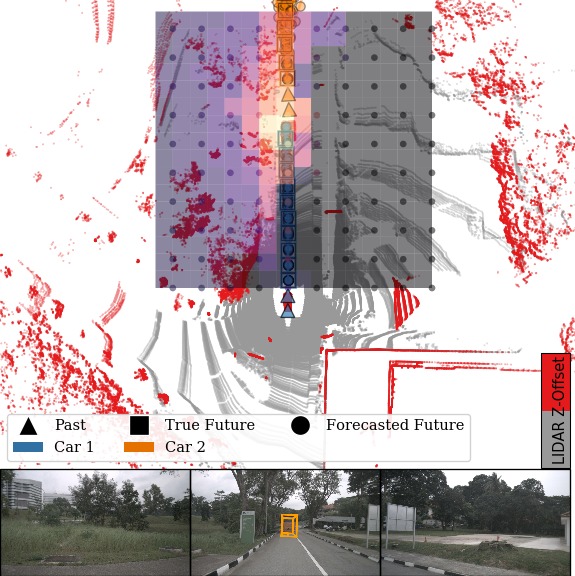}
       \end{subfigure}
    }
    \hfill
    \FBox{ 
    \begin{subfigure}[t]{\nuscpostscanwidth}
           \includegraphics[width=\textwidth,clip]{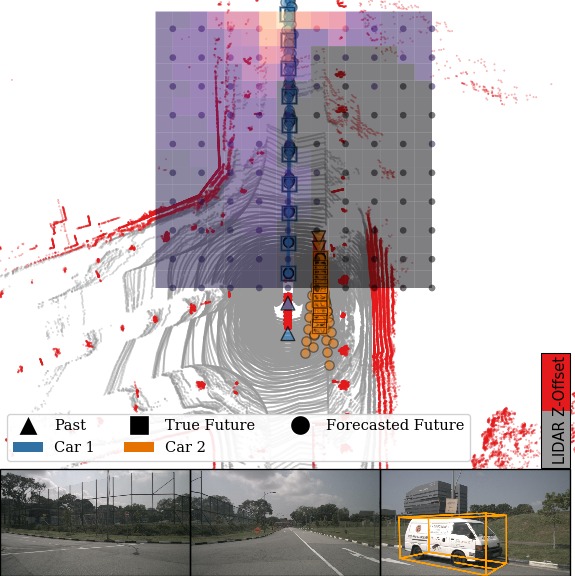}
       \end{subfigure}
    }
    
    \FBox{ 
    \begin{subfigure}[t]{\nuscpostscanwidth}
           \includegraphics[width=\textwidth,clip]{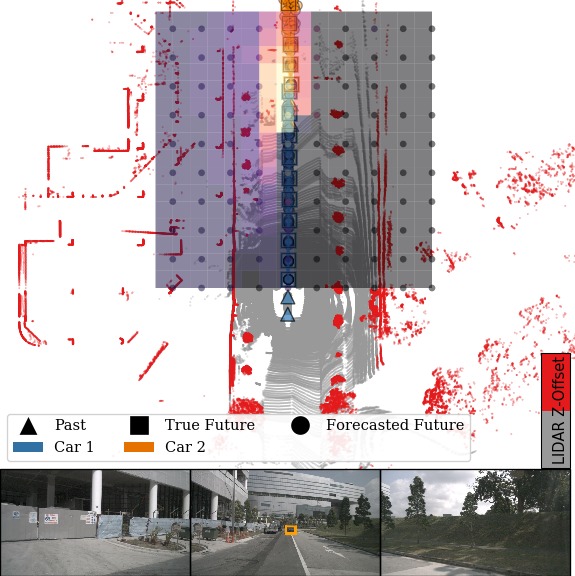}
       \end{subfigure}
    }
    \hfill
    \FBox{ 
    \begin{subfigure}[t]{\nuscpostscanwidth}
           \includegraphics[width=\textwidth,clip]{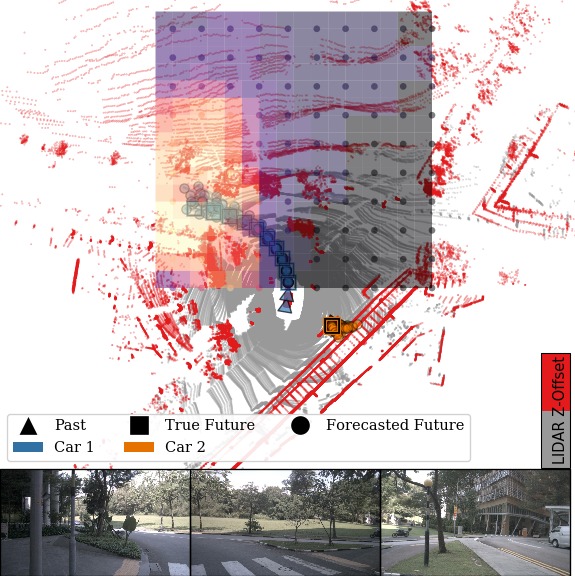}
       \end{subfigure}
    }
    \hfill
    \FBox{ 
    \begin{subfigure}[t]{\nuscpostscanwidth}
           \includegraphics[width=\textwidth,clip]{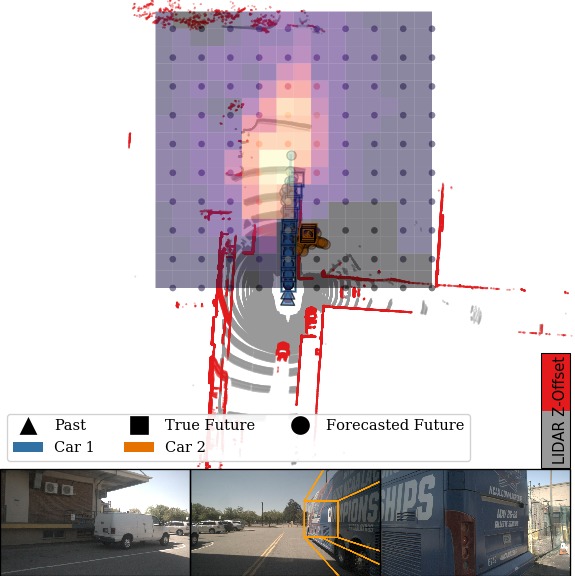}
       \end{subfigure}
    }
    \begin{flushleft}
    \begin{subfigure}[t]{\nuscpostscanwidth}
           \includegraphics[width=\textwidth,clip]{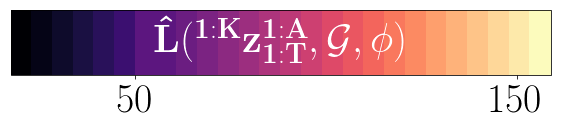}
       \end{subfigure}
       \end{flushleft}
     \vspace{-15pt}
    \caption{
    Plotting the planning criterion, $\hat L$, after planning to various positions (small circular dots in each plot) input to Alg.~\ref{alg:stateprecog}, with values interpolated between each position, in nuScenes. The planning criterion input corresponds to a spatio-temporal goal at $T=20$ in the future ($4$ seconds). The planning criterion prefers locations within its lane, unless it is uncertain about the possibility of turning. When the vehicle was stationary in the past, the planning criterion is highest at positions at or close in front of the vehicle.   
    } 
    \label{fig:posterior_scan_nuscenes}
\end{figure*}

\end{document}